\RequirePackage{fix-cm}
\documentclass[twocolumn]{svjour3}          
\smartqed  
\usepackage{graphicx}
\usepackage{natbib}
\usepackage{url}
\usepackage[colorlinks=true,citecolor=blue,linkcolor=blue,urlcolor=blue]{hyperref}
\usepackage{amsmath,amssymb}
\usepackage{color}
\usepackage{subfigure}
\usepackage{verbatim}
\usepackage{afterpage}
\usepackage{microtype}
\usepackage{theorem}
\usepackage{forloop}
\usepackage{pstool}

\theoremstyle{plain}

\newcommand{\cL}{{\cal{L}}}

\newcommand{\cC}{{\mathcal{C}}}
\newcommand{\cF}{{\mathcal{F}}}
\newcommand{\cG}{{\mathcal{G}}}
\newcommand{\cI}{{\mathcal{I}}}
\newcommand{\cM}{{\mathcal{M}}}
\newcommand{\cN}{{\mathcal{N}}}

\renewcommand{\Re}{{\mathbb{R}}}

\newcommand{\f}{{\mathbf{f}}}

\newcommand{\x}{{\mathbf{x}}}
\newcommand{\z}{{\mathbf{z}}}
\newcommand{\y}{{\mathbf{y}}}
\newcommand{\vv}{{\mathbf{v}}}

\newcommand{\w}{{\mathbf{w}}}
\renewcommand{\ll}{{\mathbf{l}}}
\newcommand{\q}{\mathbf{q}}
\newcommand{\cc}{\mathbf{c}}
\renewcommand{\u}{{\mathbf{u}}}
\newcommand{\Rho}{{\mathbf{P}}}

\newcommand{\argmin}{\mathop{\rm argmin}}
\newcommand{\argmax}{\mathop{\rm argmax}}

\newcommand{\dsum}{\displaystyle \sum}
\newcommand{\iid}{\stackrel{\mathrm{iid}}{\sim}}

\newcommand{\etc}{etc.}
\newcommand{\eg}{e.g.,\,}
\newcommand{\etal}{et al.\ }
\newcommand{\ie}{i.e.,\,}
\newcommand{\wrt}{w.r.t.\,}
\newcommand{\myparagraph}[1]{\smallskip\noindent\textbf{#1.} }
\newcommand{\figref}[1]{Fig. \ref{#1}}
\newcommand{\figrefs}[2]{Figs. \ref{#1}-\ref{#2}}
\renewcommand{\eqref}[1]{Eq. (\ref{#1})}
\newcommand{\eqsref}[2]{Eqs. (\ref{#1}-\ref{#2})}
\begin{document}

\title{Dynamic Template Tracking and Recognition
}


\author{Rizwan Chaudhry       \and
               Gregory Hager 	\and 
	    Ren\'{e} Vidal		
}


\institute{R. Chaudhry \at
              Center for Imaging Science \\
	   Johns Hopkins University \\
	   Baltimore, MD \\ 
              Tel.: +1-410-516-4095\\
              Fax: +1-410-516-4594\\
              \email{\href{mailto:rizwanch@cis.jhu.edu}{rizwanch@cis.jhu.edu}}           
           \and
           G. Hager \at
	Johns Hopkins University \\
	Baltimore, MD \\ 
           \email{\href{mailto:hager@cs.jhu.edu}{hager@cs.jhu.edu}}           
	\and
	R. Vidal \at
	Center for Imaging Science \\
	Johns Hopkins University \\
	Baltimore, MD \\ 
           \email{\href{mailto:rvidal@cis.jhu.edu}{rvidal@cis.jhu.edu}}           
}


\maketitle

\begin{abstract}
In this paper we address the problem of tracking non-rigid objects whose local appearance and motion changes as a function of time. This class of objects includes dynamic textures such as steam, fire, smoke, water, etc., as well as articulated objects such as humans performing various actions. We model the temporal evolution of the object's appearance/motion using a Linear Dynamical System (LDS). We learn such models from sample videos and use them as dynamic templates for tracking objects in novel videos. We pose the problem of tracking a dynamic non-rigid object in the current frame as a maximum a-posteriori estimate of the location of the object and the latent state of the dynamical system, given the current image features and the best estimate of the state in the previous frame. The advantage of our approach is that we can specify a-priori the type of texture to be tracked in the scene by using previously trained models for the dynamics of these textures. Our framework naturally generalizes common tracking methods such as SSD and kernel-based tracking from static templates to dynamic templates. We test our algorithm on synthetic as well as real examples of dynamic textures and show that our simple dynamics-based trackers perform at par if not better than the state-of-the-art. Since our approach is general and applicable to any image feature, we also apply it to the problem of human action tracking and build action-specific optical flow trackers that perform better than the state-of-the-art when tracking a human performing a particular action. Finally, since our approach is generative, we can use a-priori trained trackers for different texture or action classes to simultaneously track and recognize the texture or action in the video. 
\end{abstract}

\section{Introduction}\label{introduction}
Object tracking is arguably one of the most important and actively researched areas in computer vision. Accurate object tracking is generally a pre-requisite for vision-based control, surveillance and object recognition in videos. Some of the challenges to accurate object tracking are moving cameras, changing pose, scale and velocity of the object, occlusions, non-rigidity of the object shape and changes in appearance due to ambient conditions. A very large number of techniques have been proposed over the last few decades, each trying to address one or more of these challenges under different assumptions. The comprehensive survey by \citet{Yilmaz:CSUR06} provides an analysis of over 200 publications in the general area of object tracking.

In this paper, we focus on tracking objects that undergo non-rigid transformations in shape and appearance as they move around in a scene. Examples of such objects include fire, smoke, water, and fluttering flags, as well as humans performing different actions. Collectively called dynamic templates, these objects are fairly common in natural videos. Due to their constantly evolving appearance, they pose a challenge to state-of-the-art tracking techniques that assume consistency of appearance distributions or consistency of shape and contours.
 However, the change in appearance and motion profiles of dynamic templates is not entirely arbitrary and can be explicitly modeled using Linear Dynamical Systems (LDS). Standard tracking methods either use subspace models or simple Gaussian models to describe appearance changes of a mean template. Other methods use higher-level features such as skeletal structures or contour deformations for tracking to reduce dependence on appearance features. Yet others make use of foreground-background classifiers and learn discriminative features for the purpose of tracking. However, all these methods ignore the temporal dynamics of the appearance changes that are characteristic to the dynamic template.

Over the years, several methods have been developed for segmentation and recognition of dynamic templates, in particular dynamic textures. However, to the best of our knowledge the only work that explicitly addresses tracking of dynamic textures was done by \citet{Peteri:MVA10}. As we will describe in detail later, this work also does not consider the temporal dynamics of the appearance changes and does not perform well in experiments.

\myparagraph{Paper Contributions and Outline}
In the proposed approach, we model the temporal evolution of the appearance of dynamic templates using Linear Dynamical Systems (LDSs) whose parameters are learned from sample videos. These LDSs will be incorporated in a kernel based tracking framework that will allow us to track non-rigid objects in novel video sequences. In the remaining part of this section, we will review some of the related works in tracking and motivate the need for dynamic template tracking method. We will then review static template tracking in \S\ref{sec:generalFramework}. In \S\ref{sec:jointFramework}, we pose the tracking problem as the maximum a-posteriori estimate of the location of the template as well as the internal state of the LDS, given a kernel-weighted histogram observed at a test location in the image and the internal state of the LDS at the previous frame. This results in a novel joint optimization approach that allows us to simultaneously compute the best location as well as the internal state of the moving dynamic texture at the current time instant in the video. We then show how our proposed approach can be used to perform simultaneous tracking and recognition in \S\ref{sec:jointTrackingRecognition}. In \S\ref{sec:evaluation}, we first evaluate the convergence properties of our algorithm on synthetic data before validating it with experimental results on real datasets of Dynamic Textures and Human Activities in \S\ref{sec:DTExperiments}, \S\ref{sec:ActivityExperiments} and \S\ref{sec:jointTrackingRecognitionExperiments}. Finally, we will mention future research directions and conclude in \S\ref{sec:conclusions}.



\myparagraph{Prior Work on Tracking Non-Rigid and Articulated Objects} In the general area of tracking, \citet{Isard:IJCV98} and \citet{North:PAMI00} hand craft models for object contours using splines and learn their dynamics using Expectation Maximization (EM). They then use particle filtering and Markov Chain Monte-Carlo methods to track and classify the object motion. However for most of the cases, the object contours do not vary significantly during the tracking task. In the case of dynamic textures, generally there is no well-defined contour and hence this approach is not directly applicable. \citet{Black:IJCV98} propose using a robust appearance subspace model for a known object to track it later in a novel video. However there are no dynamics associated to the appearance changes and in each frame, the projection coefficients are computed independently from previous frames. \citet{Jepson:CVPR01} propose an EM-based method to estimate parameters of a mixture model that combines stable object appearance, frame-to-frame variations, and an outlier model for robustly tracking objects that undergo appearance changes. Although, the motivation behind such a model is compelling, its actual application requires heuristics and a large number of parameters. Moreover, dynamic textures do not have a stable object appearance model, instead the appearance changes according to a distinct Gauss-Markov process characteristic to the class of the dynamic texture.

Tracking of non-rigid objects is often motivated by the application of human tracking in videos. In \citet{Pavlovic:ICCV99}, a Dynamic Bayesian Network is used to learn the dynamics of human motion in a scene. Joint angle dynamics are modeled using switched linear dynamical systems and used for classification, tracking and synthesis of human motion. Although, the tracking results for human skeletons are impressive, extreme care is required to learn the joint dynamic models from manually extracted skeletons or motion capture data. Moreover a separate dynamical system is learnt for each joint angle instead of a global model for the entire object.
%
Approaches such as \citet{Leibe:PAMI08} maintain multiple hypotheses for object tracks and continuously refine them as the video progresses using a Minimum Description Length (MDL) framework. 
The work by \citet{Lim:CVPR06} models dynamic appearance by using non-linear dimensionality reduction techniques and learns the temporal dynamics of these low-dimensional representation to predict future motion trajectories. \citet{Nejhum:CVPR08} propose an online approach that deals with appearance changes due to articulation by updating the foreground shape using rectangular blocks that adapt to find the best match in every frame. However foreground appearance is assumed to be stationary throughout the video. 

Recently, classification based approaches have been proposed in \citet{Grabner:BMVC06} and \citet{Babenko:CVPR09} where classifiers such as boosting or Multiple Instance Learning are used to adapt foreground vs background appearance models with time. This makes the tracker invariant to gradual appearance changes due to object rotation, illumination changes \etc~This discriminative approach, however, does not incorporate an inherent temporal model of appearance variations, which is characteristic of, and potentially very useful, for dynamic textures.

In summary, all the above works lack a unified framework that simultaneously models the temporal dynamics of the object appearance and shape as well as the motion through the scene. Moreover, most of the works concentrate on handling the appearance changes due to articulation and are not directly relevant to dynamic textures where there is no articulation. 

\myparagraph{Prior Work on Tracking Dynamic Templates}
Recently, \citet{Peteri:MVA10} propose a first method for tracking dynamic textures using a particle filtering approach similar to the one presented in \citet{Isard:IJCV98}. However their approach can best be described as a static template tracker that uses optical flow features. The method extracts histograms for the magnitude, direction, divergence and curl of the optical flow field of the dynamic texture in the first two frames. It then assumes that the change in these flow characteristics with time can simply be modeled using a Gaussian distribution with the initially computed histograms as the mean. The variance of this distribution is selected as a parameter. Furthermore, they do not model the characteristic temporal dynamics of the intensity variations specific to each class of dynamic textures, most commonly modeled using LDSs. As we will also show in our experiments, their approach performs poorly on several real dynamic texture examples.

LDS-based techniques have been shown to be extremely valuable for dynamic texture recognition \citep{Saisan:CVPR01,Doretto:IJCV03,Chan:CVPR07,Ravichandran:CVPR09}, synthesis \citep{Doretto:IJCV03}, and registration \citep{Ravichandran:ECCV08}. They have also been successfully used to model the temporal evolution of human actions for the purpose of activity recogntion \citep{Bissacco:CVPR01,Bissacco:PAMI07,Chaudhry:CVPR09}. Therefore, it is only natural to assume that such a representation should also be useful for tracking.

Finally, \citet{Vidal:CVPR05-dyntextures} propose a method to jointly compute the dynamics as well as the optical flow of a scene for the purpose of segmenting moving dynamic textures. Using the Dynamic Texture Constancy Constraint (DTCC), the authors show that if the motion of the texture is slow, 
the optical flow corresponding to 2-D rigid motion of the texture (or equivalently the motion of the camera) can be computed using a method similar to the Lucas-Kanade optical flow algorithm. 
In principle, this method can be extended to track a dynamic texture in a framework similar to the KLT tracker. However, the requirement of having a slow-moving dynamic texture is particularly strict, especially for high-order systems and would not work in most cases. Moreover, the authors do not enforce any spatial coherence of the moving textures, which causes the segmentation results to have holes.

In light of the above discussion, we posit that there is a need to develop a principled approach for tracking dynamic templates that explicitly models the characteristic temporal dynamics of the appearance and motion. As we will show, by incorporating these dynamics, our proposed method achieves superior tracking results as well as allows us to perform simultaneous tracking and recognition of dynamic templates.

\section{Review of Static Template Tracking}\label{sec:generalFramework}

In this section, we will formulate the tracking problem as a maximum a-posteriori estimation problem and show how standard static template tracking methods such as Sum-of-Squared-Differences (SSD) and kernel-based tracking are special cases of this general problem.

Assume that we are given a static template $\mathcal{I}:\Omega \to \Re$, centered at the origin on the discrete pixel grid, $\Omega \subset \Re^2$. At each time instant, $t$, we observe an image frame, $\y_t:\cF \to \Re$, where $\cF \subset \Re^2$ is the discrete pixel domain of the image. As the template moves in the scene, it undergoes a translation, $\ll_t \in \Re^2$, from the origin of the template reference frame $\Omega$. Moreover, assume that due to noise in the observed image the intensity at each pixel in $\cF$ is corrupted by i.i.d. Gaussian noise with mean $0$, and standard deviation $\sigma_Y$. Hence, for each pixel location $\z \in \Omega + \ll_t = \{ \z' + \ll_t : \z' \in \Omega \}$, we have,
\begin{align} \label{eq:staticGenerativeModel}
\y_t(\z) & = \cI(\z\!-\!\ll_t)\!+\!\w_t(\z),~\text{where}~\w_t(\z) \iid \cN (0, \sigma_Y^2).\!\!\!\!\!
\end{align}
Therefore, the likelihood of the image intensity at pixel $\z \in \Omega + \ll_t $ is $p(\y_t(\z)|\ll_t) = \cG_{\y_t(\z)}(\cI(\z - \ll_t), \sigma_Y^2)$, where
\begin{align}
\cG_\x(\mu,\Sigma) = \frac{1}{(2\pi)^{\frac{n}{2}}|\Sigma|^{\frac{1}{2}}}\exp\left\{-\frac{1}{2}(\x-\mu)^\top\Sigma^{-1}(\x-\mu)\right\} \nonumber
\end{align}
is the $n$-dimensional Gaussian pdf with mean $\mu$ and covariance $\Sigma$. Given $\y_t=[\y_t(\z)]_{\z\in\cF}$, \ie the stack of all the pixel intensities in the frame at time $t$, we would like to maximize the posterior, $p(\ll_t|\y_t)$. Assuming a uniform background, \ie $p(\y_t(\z) | \ll_t) = 1/K$ if $\z-\ll_t \not\in \Omega$ and a uniform prior for $\ll_t$, \ie $p(\ll_t) = |\cF|^{-1}$, we have,
\begin{align}
p(\ll_t|\y_t) & = \dfrac{p(\y_t|\ll_t) p(\ll_t)}{p(\y_t)} \nonumber \\
& \propto p(\y_t|\ll_t) = \prod_{\z\in\Omega + \ll_t}p(\y_t(\z)|\ll_t) \label{eq:posteriorUniformLocation} \\
& \propto \exp\left\{ -\frac{1}{2\sigma_Y^2}\dsum_{\z \in \Omega + \ll_t } (\y_t(\z)-\cI(\z-\ll_t))^2\right\}. \nonumber
\end{align}
The optimum value, $\hat{\ll}_t$ will maximize the log posterior and after some algebraic manipulations, we get
\begin{align}
\hat{\ll}_t = \argmin_{\ll_t}\dsum_{\z \in \Omega + \ll_t} (\y_t(\z)-\cI(\z-\ll_t))^2.
\end{align}
Notice that with the change of variable, $\z^\prime = \z-\ll_t$, we can shift the domain from the image pixel space $\cF$ to the template pixel space $\Omega$, to get $\hat{\ll}_t = \argmin_{\ll_t} O(\ll_t)$, where 
\begin{align} \label{eq:SSD}
O(\ll_t) = \dsum_{\z^\prime \in \Omega} (\y_t(\z^\prime+\ll_t)-\cI(\z^\prime))^2.
\end{align}
\eqref{eq:SSD} is the well known optimization function used in Sum of Squared Differences (SSD)-based tracking of static templates. The optimal solution is found either by searching brute force over the entire image frame or, given an initial estimate of the location, by performing gradient descent iterations: $\ll_t^{i+1} = \ll_t^i - \gamma\nabla_{\ll_t}O(\ll_t^i)$. Since the image intensity function, $\y_t$ is non-convex, a good initial guess, $\ll_t^0$, is very important for the gradient descent algorithm to converge to the correct location. Generally, $\ll_t^0$ is initialized using the optimal location found at the previous time instant, \ie $\hat{\ll}_{t-1}$, and $\hat{\ll}_0$ is hand set or found using an object detector.

In the above description, we have assumed that we observe the intensity values, $\y_t$, directly. However, to develop an algorithm that is invariant to nuisance factors such as changes in contrast and brightness, object orientations, \etc, we can choose to compute the value of a more robust function that also considers the intensity values over a neighborhood $\Gamma(\z) \subset \Re^2$ of the pixel location $\z$,
\begin{equation}
\f_t(\z) = f([\y_t(\z^\prime)]_{\z^\prime \in \Gamma(\z)}), \quad f: \Re^{|\Gamma|} \to \Re^d,
\end{equation}
where $[\y_t(\z^\prime)]_{\z^\prime \in \Gamma(\z)}$ represents the stack of intensity values of all the pixel locations in $\Gamma(\z)$. We can therefore treat the value of $\f_t(\z)$ as the observed random variable at the location $\z$ instead of the actual intensities, $\y_t(\z)$.

Notice that even though the conditional probability of the intensity of individual pixels is Gaussian, as in \eqref{eq:staticGenerativeModel}, under the (possibly) non-linear transformation, $f$, the conditional probability of $\f_t(\z)$ will no longer be Gaussian. However, from an empirical point of view, using a Gaussian assumption in general provides very good results. Therefore, due to changes in the location of the template, we observe $\f_t(\z) = f([\cI(\z^\prime)]_{\z^\prime \in \Gamma(\z-\ll_t)}) + \w_t^f(\z)$, where $\w_t^f(\z) \iid \cN(0,\sigma_f^2)$ is isotropic Gaussian noise.

Following the same derivation as before, the new cost function to be optimized becomes,
\begin{align}
O(\ll_t) & = \dsum_{\z \in \Omega}\|f([\y_t(\z^\prime)]_{\z^\prime \in \Gamma(\z+\ll_t)}) - f([\cI(\z^\prime)]_{\z^\prime \in \Gamma(\z)})\|^2 \nonumber \\
& \doteq \|F(\y_t(\ll_t)) - F(\cI)\|^2, \label{eq:generalSSDFunction}
\end{align}
where,
\begin{align}
F(\y_t(\ll_t)) & \doteq [f([\y_t(\z^\prime)]_{\z^\prime \in \Gamma(\z+\ll_t)})]_{\z \in \Omega} \nonumber \\
F(\cI) & \doteq [f([\cI(\z^\prime)]_{\z^\prime \in \Gamma(\z)})]_{\z \in \Omega} \nonumber
\end{align}
By the same argument as in \eqref{eq:SSD}, $\hat{\ll}_t=\argmin_{\ll_t}O(\ll_t)$ also maximizes the posterior, $p(\ll_t|F(\y_t))$, where
\begin{align}
F(\y_t) \doteq [f([\y_t(\z^\prime)]_{\z^\prime \in \Gamma(\z)}]_{\z \in \cF},
\end{align}
is the stack of all the function evaluations with neighborhood size $\Gamma$ over all the pixels in the frame.

For the sake of simplicity, from now on as in \eqref{eq:generalSSDFunction}, we will abuse the notation and use $\y_t(\ll_t)$ to denote the stacked vector $[\y_t(\z^\prime)]_{\z^\prime \in \Gamma(\z+\ll_t)}$, and $\y_t$ to denote the full frame, $[\y_t(\z)]_{\z\in\cF}$. Moreover, assume that the ordering of pixels in $\Omega$ is in column-wise format, denoted by the set $\{1,\ldots,N\}$. Finally, if the size of the neighborhood, $\Gamma$, is equal to the size of the template, $\cI$, \ie $|\Gamma| = |\Omega|$, $f$ will only need to be computed at the central pixel of $\Omega$, shifted by $\ll_t$, \ie
\begin{align} \label{eq:fullPatchEQSSD}
O(\ll_t) & =\|f(\y_t(\ll_t)) - f(\cI)\|^2
\end{align}

\myparagraph{Kernel based Tracking} One special class of functions that has commonly been used in \emph{kernel-based tracking} methods \citep{Comaniciu:PAMI02,Comaniciu:PAMI03} is that of kernel-weighted histograms of intensities. These functions have very useful properties in that, with the right choice of the kernel, they can be made either robust to variations in the pose of the object, or sensitive to certain discriminatory characteristics of the object. This property is extremely useful in common tracking problems and is the reason for the wide use of kernel-based tracking methods. In particular, a kernel-weighted histogram, $\rho=[\rho_1,\ldots,\rho_B]^\top$ with $B$ bins, $ u = 1, \ldots, B $, computed at pixel location $\ll_t$, is defined as,
\begin{align}
\rho_u(\y_t(\ll_t)) = \frac{1}{\kappa}\dsum_{\z\in\Omega}K(\z)\delta(b(\y_t(\z+\ll_t))-u),\label{eq:histFunction}
\end{align}
where $b$ is a binning function for the intensity $\y_t(\z)$ at the pixel $\z$, $\delta$ is the discrete Kronecker delta function, and $\kappa = \sum_{\z\in\Omega} K(\z)$ is a normalization constant such that the sum of the histogram equals 1. One of the more commonly used kernels is the Epanechnikov kernel,
\begin{equation} \label{eq:epanKernel}
K(\z) = \left\{ \begin{array}{l l}
					1 - \|H\z\|^2, & \|H\z\| < 1 \\
					0, 	 &	\text{otherwise}
					\end{array} \right.
\end{equation}
where $H = \text{diag}([r^{-1}, c^{-1}])$ is the bandwidth scaling matrix of the kernel corresponding to the size of the template, \ie $|\Omega| = r \times c$.

Using the fact that we observe $F = \sqrt{\rho}$, the entry-wise square root of the histogram, we get the Matusita metric between the kernel weighted histogram computed at the current location in the image frame and that of the template:
\begin{align}\label{eq:singleSSDKernel}
O(\ll_t) = \| \sqrt{\mathbf{\rho}(\y_t(\ll_t))} - \sqrt{\rho(\cI)}\|^2.
\end{align}
\citet{Hager:CVPR04} showed that the minimizer of the objective function in \eqref{eq:singleSSDKernel} is precisely the solution of the meanshift tracker as originally proposed by \citet{Comaniciu:PAMI02} and \citet{Comaniciu:PAMI03}. The algorithm in \citet{Hager:CVPR04} then proceeds by computing the optimal $\ll_t$ that minimizes \eqref{eq:singleSSDKernel} using a Newton-like approach. We refer the reader to \citet{Hager:CVPR04} for more details. \citet{Hager:CVPR04} then propose using multiple kernels and \citet{Fan:PAMI07} propose structural constraints  to get unique solutions in difficult cases. All these formulations eventually boil down to the solution of a problem of the form in  \eqref{eq:fullPatchEQSSD}.

\myparagraph{Incorporating Location Dynamics} The generative model in \eqref{eq:posteriorUniformLocation} assumes a uniform prior on the probability of the location of the template at time $t$ and that the location at time $t$ is independent of the location at time $t-1$. If applicable, we can improve the performance of the tracker by imposing a known motion model, $\ll_t = g(\ll_{t-1})+\w_t^g$, such as constant velocity or constant acceleration. In this case, the likelihood model is commonly appended by,
\begin{align}
p(\ll_t|\ll_{t-1}) = \cG_{\ll_t}(g(\ll_{t-1}), \Sigma_g) .\label{eq:locationLikelihood}
\end{align}
From here, it is a simple exercise to see that the maximum a-posteriori estimate of $\ll_t$ given all the frames, $\y_0, \ldots, \y_t$ can be computed by the extended Kalman filter or particle filters since $f$, in \eqref{eq:fullPatchEQSSD}, is a function of the image intensities and therefore a non-linear function on the pixel domain.


\section{Tracking Dynamic Templates} \label{sec:jointFramework}

In the previous section we reviewed kernel-based methods for tracking a static template $\cI : \Omega\to\Re$. In this section we propose a novel kernel-based framework for tracking a dynamic template $\cI_t: \Omega\to\Re$. For ease of exposition, we derive the framework under the assumption that the location of the template $\ll_t$ is equally likely on the image domain. For the case of a dynamic prior on the location, the formulation will result in an extended Kalman or particle filter as briefly mentioned at the end of \S\ref{sec:generalFramework}.

We model the temporal evolution of the dynamic template $\cI_t$ using Linear Dynamical Systems (LDSs). LDSs are represented by the tuple $(\mu, A,C,B)$ and satisfy the following equations for all time $t$:
\begin{eqnarray}
\x_t & = & A\x_{t-1} + B\vv_t, \label{eq:LDS-1} \\
\cI_t & = & \mathbf{\mu} + C\x_t  . \label{eq:LDS-2}
\end{eqnarray}
Here $\cI_t\in\Re^{|\Omega|}$ is the stacked vector, $[\cI_t(\z)]_{\z \in \Omega}$, of image intensities of the dynamic template at time $t$, and $\x_t$ is the (hidden) state of the system at time $t$. The current state is linearly related to the previous state by the state transition matrix $A$ and the current output is linearly related to the current state by the observation matrix $C$. $\vv_t$ is the process noise, which is assumed to be Gaussian and independent from $\x_t$. Specifically, $B\vv_t \sim \cN(0, Q)$, where $Q = BB^\top$.


Tracking dynamic templates requires knowledge of the system parameters, $(\mu, A, C, B)$, for dynamic templates of interest. Naturally, these parameters have to be learnt from training data. Once these parameters have been learnt, they can be used to track the template in a new video. However the size, orientation, and direction of motion of the template in the test video might be very different from that of the training videos and therefore our procedure will need to be invariant to these changes. In the following, we will propose our dynamic template tracking framework by describing in detail each of these steps,
\begin{enumerate}
\item Learning the system parameters of a dynamic template from training data,
\item Tracking dynamic templates of the same size, orientation, and direction of motion as training data, 
\item Discussing the convergence properties and parameter tuning, and
\item Incorporating invariance to size, orientation, and direction of motion.
\end{enumerate}

\subsection{LDS Parameter Estimation} \label{subsec:sysID}

We will first describe the procedure to learn the system parameters, $(\mu, A, C, B)$, of a dynamic template from a training video of that template. Assuming that we can manually, or semi-automatically, mark a bounding box of size $|\Omega| = r \times c$ rows and columns, which best covers the spatial extent of the dynamic template at each frame of the training video. The center of each box gives us the location of the template at each frame, which we use as the ground truth template location. Next, we extract the sequence of column-wise stacked pixel intensities, $\cI = [\cI_1, \ldots, \cI_N]$ corresponding to the appearance of the template at each time instant in the marked bounding box. We can then compute the system parameters using the system identification approach proposed in \cite{Doretto:IJCV03}. Briefly, given the sequence, $\cI$, we compute a compact, rank $n$, singular value decompostion (SVD) of the matrix, $\tilde{\cI} = [\cI_1 - \mu, \ldots, \cI_N - \mu] = U \Sigma V^\top$. Here $\mu = \frac{1}{N} \sum_{t=1}^N \cI_i$, and $n$ is the order of the system and is a parameter. For all our experiments, we have chosen $n = 5$. We then compute $C = U$, and the state sequence $X_{1}^{N} = \Sigma V^\top$, where $X_{t_1}^{t_2} = [\x_{t_1}, \ldots, \x_{t_2}]$. Given $X_1^N$, the matrix $A$ can be computed using least-squares as $A = X_2^N (X_1^{N-1})^\dagger$, where $X^\dagger$ represents the pseudo-inverse of $X$. Also, $Q = \frac{1}{N-1} \sum_{t=1}^{N-1} \vv_t^\prime (\vv_t^\prime)^\top$ where $\vv_t^\prime = B\vv_t = \x_{t+1} - \x_t$. $B$ is computed using the Cholesky factorization of $Q = BB^\top$.

\subsection{Tracking Dynamic Templates} \label{subsec:trackingDT}

\begin{figure}
\centering
\includegraphics[width=0.7\linewidth]{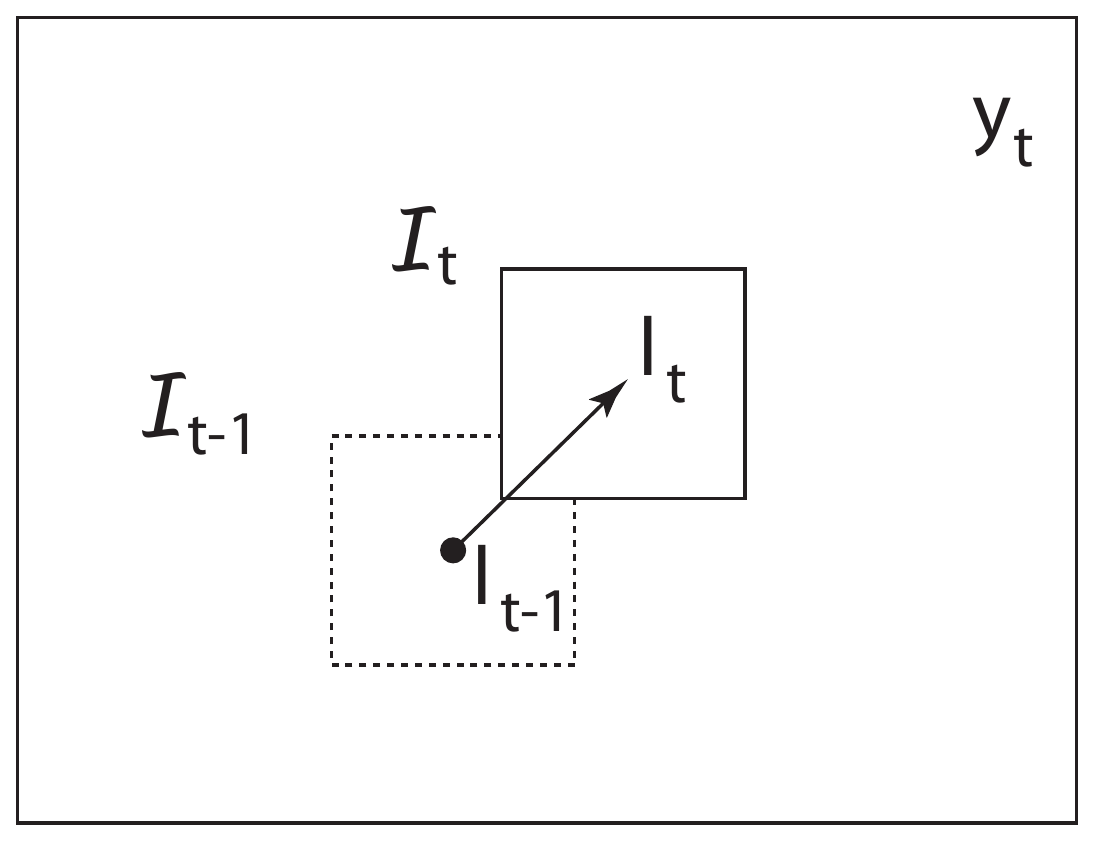}
\caption{Illustration of the dynamic template tracking problem.}
\label{fig:tracking-problem}
\end{figure}

\begin{figure}
\centering
\includegraphics[width=0.7\linewidth]{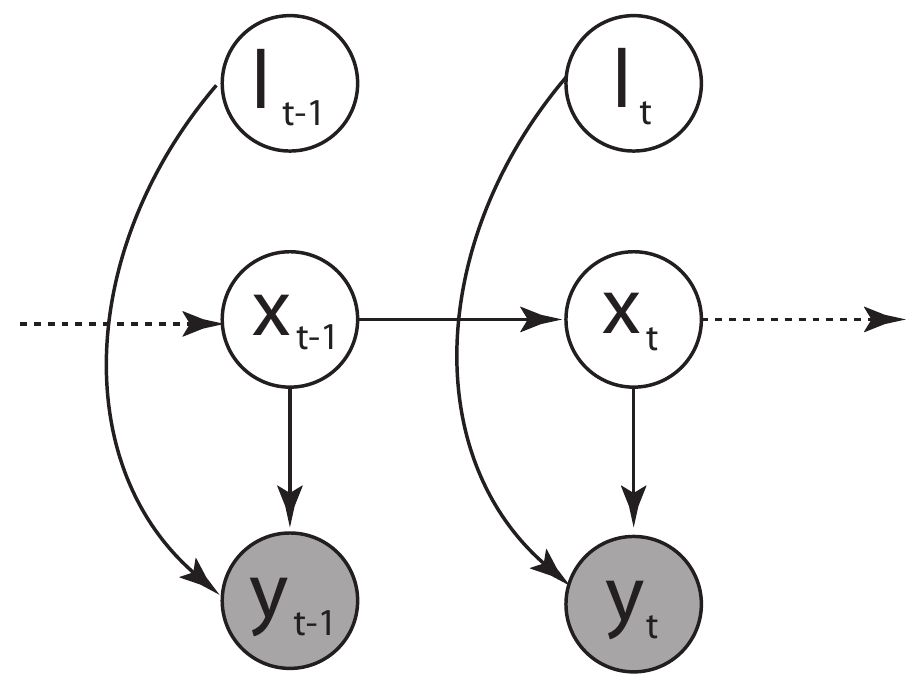}
\caption{Graphical representation for the generative model of the observed template.}
\label{fig:graphical-model}
\end{figure}

\myparagraph{Problem Formulation}
We will now formulate the problem of tracking a dynamic template of size $|\Omega| = r \times c$, with known system parameters $(\mu, A, C, B)$. Given a test video, at each time instant, $t$, we observe an image frame, $\y_t:\cF \to \Re$, obtained by translating the template $\cI_t$ by an amount $\ll_t \in \Re^2$ from the origin of the template reference frame $\Omega$. Previously, at time instant $t-1$, the template was observed at location $\ll_{t-1}$ in frame $\y_{t-1}$. In addition to the change in location, the intensity of the dynamic template changes according to \eqsref{eq:LDS-1}{eq:LDS-2}. Moreover, assume that due to noise in the observed image the intensity at each pixel in $\cF$ is corrupted by i.i.d. Gaussian noise with mean $0$, and standard deviation $\sigma_Y$. Therefore, the intensity at pixel $\z\in\cF$ given the location of the template and the current state of the dynamic texture is
\begin{align}
\y_t(\z) & = \cI_t(\z-\ll_t)+\w_t(\z)\\
& = \mu(\z-\ll_t)+C(\z-\ll_t)^\top\x_t + \w_t(\z),
\end{align}
where the pixel $\z$ is used to index $\mu$ in \eqref{eq:LDS-1} according to the ordering in $\Omega$, \eg in a column-wise fashion. Similarly, $C(\z)^\top$ is the row of the $C$ matrix in \eqref{eq:LDS-1} indexed by the pixel $\z$. \figref{fig:tracking-problem} illustrates the tracking scenario and \figref{fig:graphical-model} shows the corresponding graphical model representation\footnote{Notice that since we have assumed that the location of the template at time $t$ is independent of its location at time $t-1$, there is no link from $\ll_{t-1}$ to $\ll_t$ in the graphical model.}. We only observe the frame, $\y_t$ and the appearance of the frame is conditional on the location of the dynamic template, $\ll_t$ and its state, $\x_t$. 

As described in \S\ref{sec:generalFramework}, rather than using the image intensities $\y_t$ as our measurements, we compute a kernel-weighted histogram centered at each test location $\ll_t$,
\begin{align}
\rho_u(\y_t(\ll_t)) & = \frac{1}{\kappa}\dsum_{\z\in\Omega} K(\z)\delta(b(\y_t(\z+\ll_t))-u) .
\label{eq:trueKWHist}
\end{align}
In an entirely analogous fashion, we compute a kernel-weighted histogram of the template
\begin{align}
\rho_u(\cI_t(\x_t)) = 
\frac{1}{\kappa} \dsum_{\z \in \Omega} K(\z)\delta(b(\mu(\z) + C(\z)^\top \x_t) - u), \!\!\!
\label{eq:meanHistProcess}
\end{align}
where we write $\cI_t(\x_t)$ to emphasize the dependence of the template $\cI_t$ on the latent state $\x_t$, which needs to be estimated together with the template location $\ll_t$.


Since the space of histograms is a non-Euclidean space, we need a metric on the space of histograms to be able to correctly compare the observed kernel-weighted histogram with that generated by the template. One convenient metric on the space of histograms is the Matusita metric, 
\begin{align}
d(\rho^1, \rho^2) = \dsum_{i=1}^B \left(\sqrt{\rho_i^1} - \sqrt{\rho_i^2}\right)^2 \label{eq:MatusitaMetric}
\end{align}
which was also previously used in \eqref{eq:singleSSDKernel} for the static template case by \cite{Hager:CVPR04}. Therefore, we approximate the probability of the square root of each entry of the histogram as,
\begin{align}
p(\sqrt{\rho_u(\y_t)}|\ll_t,\x_t) \approx \cG_{\sqrt{\rho_u(\y_t(\ll_t))}}(\sqrt{\rho_u(\cI_t(\x_t))},\sigma_H^2), \!\!
\end{align}
where $\sigma_H^2$ is the variance of the entries of the histogram bins. The tracking problem therefore results in the maximum a-posteriori estimation,
\begin{align} \label{eq:trackingEstimateProblem}
(\hat{\ll}_t,\hat{\x}_t) = \argmax_{\ll_t,\x_t}p(\ll_t,\x_t|\sqrt{\rho(\y_1)},\ldots,\sqrt{\rho(\y_t)})
\end{align}
where $\{\rho(\y_i)\}_{i=1}^t$ are the kernel-weighted histograms computed at each image location in all the frames with the square-root taken element-wise. Deriving an optimal non-linear filter for the above problem might not be computationally feasible and we might have to resort to particle filtering based approaches. However, as we will explain, we can simplify the above problem drastically by proposing a greedy solution that, although not provably optimal, is computationally efficient. Moreover, as we will show in the experimental section, this solution results in an algorithm that performs at par or even better than state-of-the-art tracking methods.

\begin{figure}
\centering
\includegraphics[width=0.6\linewidth]{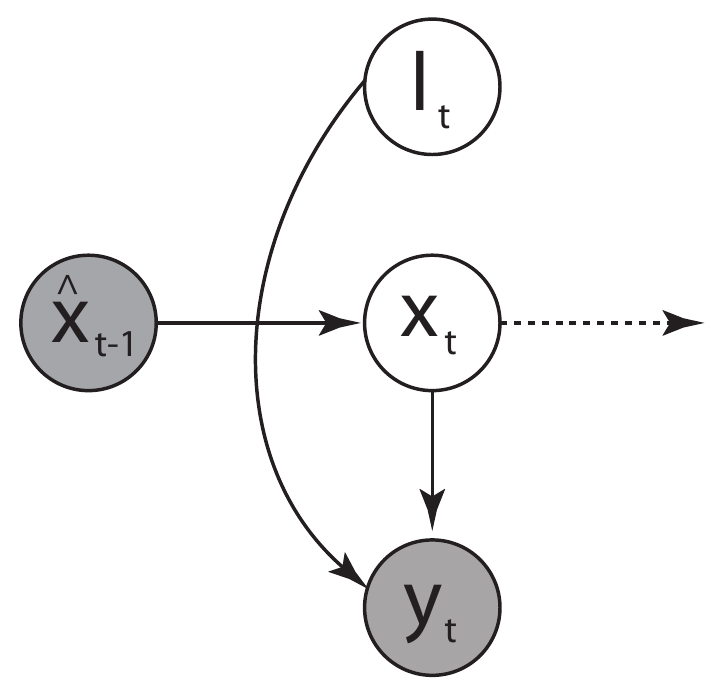}
\caption{Graphical Model for the approximate generative model of the observed template}
\label{fig:approx-graphical-model}
\end{figure}

\myparagraph{Bayesian Filtering}
Define $\Rho_t = \{\sqrt{\rho(\y_1)} \ldots \sqrt{\rho(\y_t)}\}$, and consider the Bayesian filter derivation for \eqref{eq:trackingEstimateProblem}:
\begin{align}
\begin{split}
& p(\ll_t, \x_t|\Rho_t)  = p(\ll_t, \x_t  | \Rho_{t-1}, \sqrt{\rho(\y_t)})\nonumber \\
& \quad = \dfrac{p(\sqrt{\rho(\y_t)}| \ll_t, \x_t) p(\ll_t, \x_t|\Rho_{t-1})}{p(\sqrt{\rho(\y_t)}|\Rho_{t-1})} \nonumber \\
& \quad \propto p(\sqrt{\rho(\y_t)}| \ll_t, \x_t) p(\ll_t, \x_t|\Rho_{t-1}) \nonumber \\
& \quad = p(\sqrt{\rho(\y_t)}| \ll_t, \x_t) \int_{\mathcal{X}_{t-1}} p(\ll_t, \x_t, \x_{t-1}|\Rho_{t-1}) d\x_{t-1} \nonumber \\
& \quad = p(\sqrt{\rho(\y_t)}| \ll_t, \x_t). \nonumber \\
& \quad \qquad \int_{\mathcal{X}_{t-1}} p(\ll_t, \x_t | \x_{t-1}) p(\x_{t-1}|\Rho_{t-1}) d\x_{t-1} \nonumber \\
& \quad = p(\sqrt{\rho(\y_t)}| \ll_t, \x_t)  p(\ll_t) . \nonumber \\
& \quad \qquad \int_{\mathcal{X}_{t-1}} p(\x_t | \x_{t-1}) p(\x_{t-1}|\Rho_{t-1}) d\x_{t-1}, \label{eq:truePosteriorRecursion}
\end{split}
\end{align}
where we have assumed a uniform prior $p(\ll_t) = |\cF|^{-1}$. Assuming that we have full confidence in the estimate $\hat{\x}_{t-1}$ of the state at time $t-1$, we can use the \emph{greedy} posterior, $p(\x_{t-1}|\Rho_{t-1}) = \delta(\x_{t-1} = \hat{\x}_{t-1})$, to greatly simplify \eqref{eq:truePosteriorRecursion} as,
\begin{align}
p(\ll_t, \x_t|\Rho_t) & \propto p(\sqrt{\rho(\y_t)}|\ll_t,\x_t) p(\x_t|\x_{t-1} = \hat{\x}_{t-1}) p(\ll_t) \nonumber \\
			 & \propto p(\sqrt{\rho(\y_t)}|\ll_t,\x_t) p(\x_t|\x_{t-1} = \hat{\x}_{t-1}).
\end{align}
 \figref{fig:approx-graphical-model} shows the graphical model corresponding to this approximation.

After some algebraic manipulations, we arrive at,
\begin{align}
(\hat{\ll}_t, \hat{\x}_t) = \argmin_{\ll_t,\x_t} O(\ll_t,\x_t)
\end{align}
where,
\begin{align}
O(\ll_t,\x_t) = & \dfrac{1}{2\sigma_H^2}\|\sqrt{\rho(\y_t(\ll_t))}-\sqrt{\rho(\mu+C\x_t)}\|^2+ \nonumber \\
& \dfrac{1}{2}(\x_t-A\hat{\x}_{t-1})^\top Q^{-1}(\x_t-A\hat{\x}_{t-1}). \label{eq:jointOptimization}
\end{align}

\begin{figure}
\centering
\subfigure[Binning function, $b(.)$, for a histogram with 10 bins, $B=10$.]{
\includegraphics[width=0.7\linewidth]{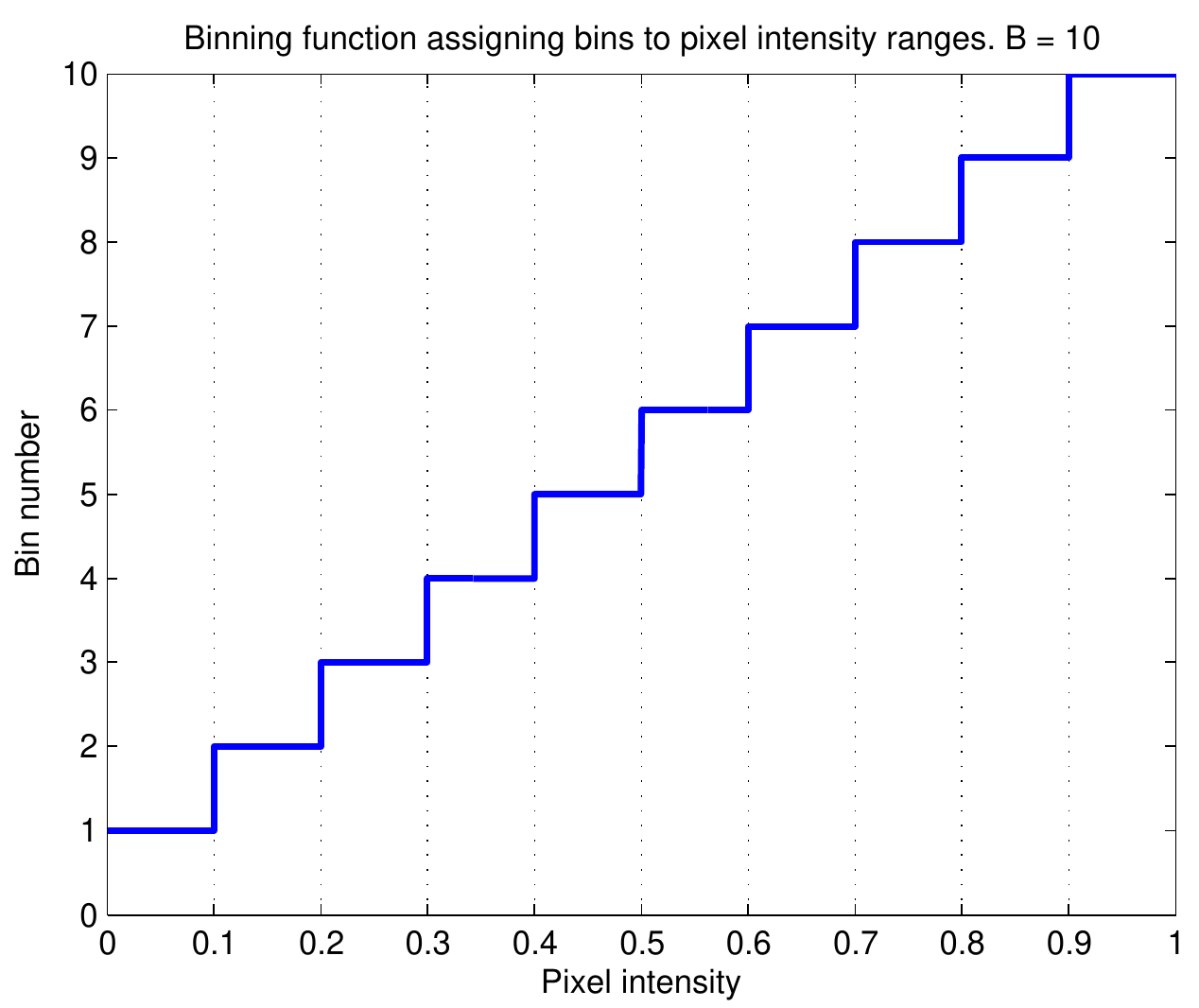}
\label{fig:binningFunction}
}
\\
\subfigure[Exact binning]{
\includegraphics[width=0.45\linewidth]{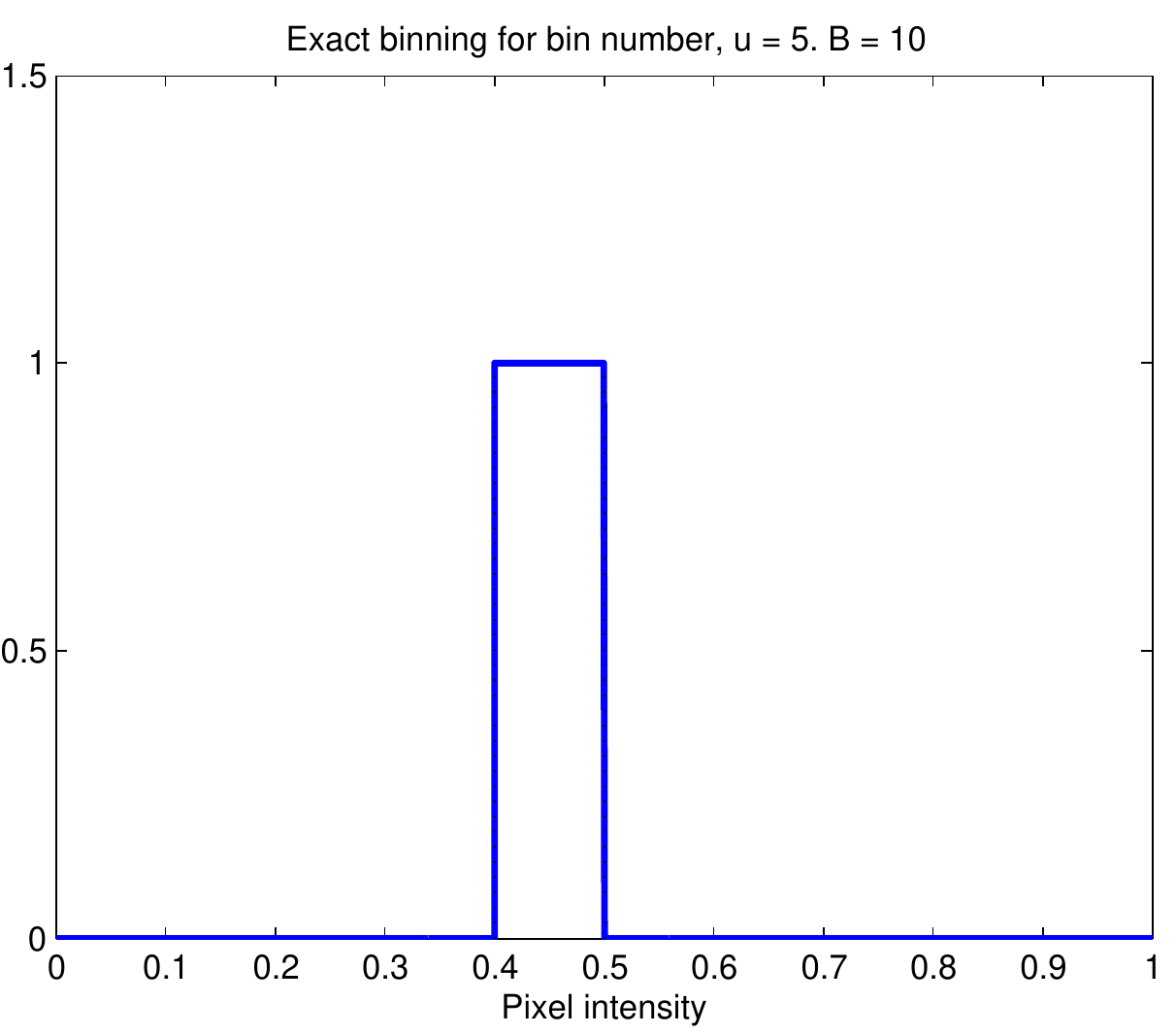}
\label{fig:exactBinning}
}
\subfigure[Approximate binning]{
\includegraphics[width=0.45\linewidth]{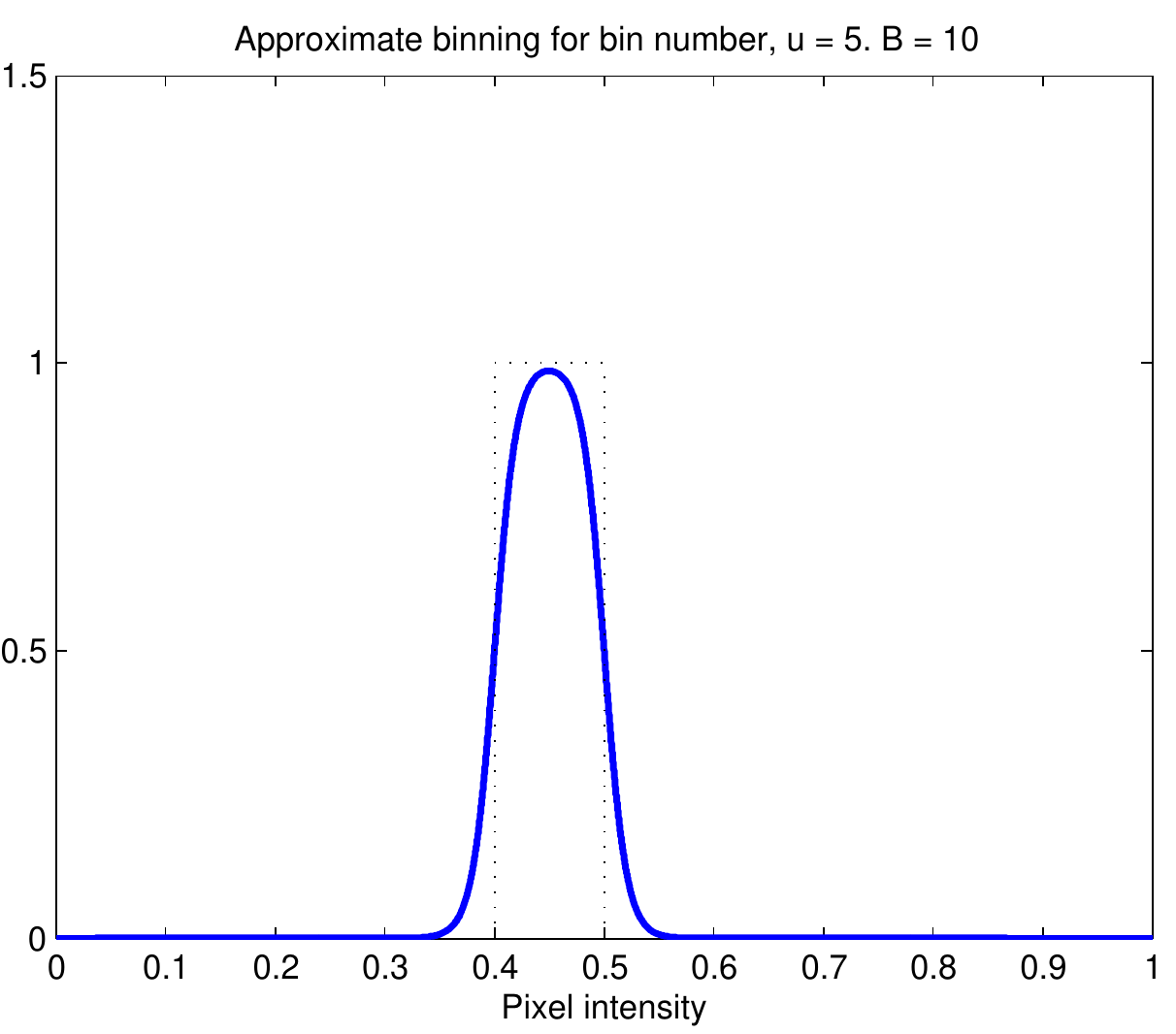}
\label{fig:approximateBinning}
}
\caption{Binning function for $B = 10$ bins and the corresponding non-differentiable exact binning strategy vs our proposed differentiable but approximate binning strategy, for bin number $u = 5$.}
\end{figure}

\myparagraph{Simultaneous State Estimation and Tracking}
To derive a gradient descent scheme, we need to take derivatives of $O$ \wrt $\x_t$. However, the histogram function, $\rho$ is not differentiable \wrt~$\x_t$ because of the $\delta$ function and hence we cannot compute the required derivative of \eqref{eq:jointOptimization}. Instead of $\rho$, we propose to use $\zeta$, where $\zeta$ is a \emph{continuous} histogram function defined as,
\begin{align} \label{eq:contHistFunction}
\zeta_u(\y_t(\ll_t)) = & \dfrac{1}{\kappa}\dsum_{\z \in \Omega} K(\z) \nonumber \\
&\;\;\quad \left(\phi_{u-1}(\y_t(\z\!+\!\ll_t))\!-\!\phi_{u}(\y_t(\z\!+\!\ll_t))\right),
\end{align}
where $\phi_u(s) = \left(1+\exp\{-\sigma(s-r(u))\}\right)^{-1}$ is the sigmoid function. 
With a suitably chosen $\sigma$ ($\sigma = 100$ in our case), the difference of the two sigmoids is a continuous and differentiable function that approximates a step function on the grayscale intensity range $[r(u-1),r(u)]$. For example, for a histogram with $B$ bins, to uniformly cover the grayscale intensity range from 0 to 1, we have $r(u) = \frac{u}{B}$. The difference, $\phi_{u-1}(y)-\phi_{u}(y)$, will therefore give a value close to 1 when the pixel intensity is in the range $[\frac{u-1}{B},\frac{u}{B}]$, and close to 0 otherwise, thus contributing to the $u$-th bin. \figref{fig:binningFunction} illustrates the binning function $b(.)$ in Eqs. (\ref{eq:histFunction}, \ref{eq:trueKWHist}, \ref{eq:meanHistProcess}) for a specific case where the pixel intensity range $[0,1]$ is equally divided into $B=10$ bins. \figref{fig:exactBinning} shows the non-differentiable but exact binning function, $\delta(b(.),u)$, for bin number $u=5$, whereas \figref{fig:approximateBinning} shows the corresponding proposed approximate but differentiable binning function, $(\phi_{u-1}-\phi_u)(.)$. As we can see, the proposed function responds with a value close to 1 for intensity values between $r(5-1) = 0.4$, and $r(5) = 0.5$. The spatial kernel weighting of the values is done in exactly the same way as for the non-continuous case in \eqref{eq:histFunction}. 

We can now find the optimal location and state at each time-step by performing gradient descent on \eqref{eq:jointOptimization} with $\rho$ replaced by $\zeta$. This gives us the following iterations (see Appendix \ref{sec:app-gradDescentDerivation} for a detailed derivation)
\begin{equation} \label{eq:gradDescent}
\begin{bmatrix}
\ll_t^{i+1} \\
\x_t^{i+1}
\end{bmatrix} = 
\begin{bmatrix}
\ll_t^i \\
\x_t^i
\end{bmatrix} - 2\gamma
\begin{bmatrix}
\mathbf{L}_i^\top \mathbf{a}_i \\
-\mathbf{M}_i^\top \mathbf{a}_i + \mathbf{d}_i
\end{bmatrix},
\end{equation}
where,
\begin{eqnarray*}
\mathbf{a} & = & \sqrt{\mathbf{\zeta}(\y_t(\ll_t))} - \sqrt{\mathbf{\zeta}(\mu+C\x_t)}\\
\mathbf{d} & = & Q^{-1}(\x_t-A\hat{\x}_{t-1}) \\
\mathbf{L} & = & \dfrac{1}{2\sigma_H^2} \text{diag}(\mathbf{\zeta}(\y_t(\ll_t)))^{-\frac{1}{2}}\mathbf{\tilde{U}}^\top\mathbf{J}_K \\
\mathbf{M} & = & \dfrac{1}{2\sigma_H^2} \text{diag}(\mathbf{\zeta}(\mu+C\x_t))^{-\frac{1}{2}}(\mathbf{\Phi}^\prime)^\top \frac{1}{\kappa}\text{diag}(\mathbf{K})C.
\end{eqnarray*}
The index $i$ in \eqref{eq:gradDescent} represents evaluation of the above quantities using the estimates $(\ll_t^i, \x_t^i)$ at iteration $i$. Here, $\mathbf{J}_K$ is the Jacobian of the kernel $K$, 
\begin{equation}
\mathbf{J}_K = [\nabla K(\z_1) \ldots \nabla K(\z_N) ],  \nonumber
\end{equation}
and $\mathbf{\tilde{U}} = [\tilde{\u}_1,\tilde{\u}_2,\ldots,\tilde{\u}_B]$ is a real-valued \emph{sifting} matrix (analogous to that in \citet{Hager:CVPR04}) with,
\begin{equation}
\tilde{\u}_j = \begin{bmatrix}
\phi_{j-1}(\y_t(\z_1))-\phi_{j}(\y_t(\z_1)) \\
\phi_{j-1}(\y_t(\z_2))-\phi_{j}(\y_t(\z_2)) \\
\vdots \\
\phi_{j-1}(\y_t(\z_N))-\phi_{j}(\y_t(\z_N))
\end{bmatrix}, \nonumber
\end{equation}
where the numbers $1,\ldots,N$ provide an index in the pixel domain of $\Omega$, as previously mentioned. $\mathbf{\Phi}^\prime = [\Phi_1^\prime, \Phi_2^\prime, \ldots, \Phi_B^\prime] \in \Re^{N \times B}$ is a matrix composed of derivatives of the difference of successive sigmoid functions with,
\begin{equation} \label{eq:phiPrimeColumn}
\Phi_j^\prime = 
\begin{bmatrix}
(\phi_{j-1}^\prime - \phi_j^\prime)(\mu(\z_1)+C(\z_1)^\top \x_t) \\
(\phi_{j-1}^\prime - \phi_j^\prime)(\mu(\z_2)+C(\z_2)^\top \x_t) \\
\vdots \\
(\phi_{j-1}^\prime - \phi_j^\prime)(\mu(\z_N)+C(\z_N)^\top \x_t)
\end{bmatrix}.
\end{equation}

\myparagraph{Initialization}
Solving \eqref{eq:gradDescent} iteratively will simultaneously provide the location of the dynamic texture in the scene as well as the internal state of the dynamical system. However, notice that the function $O$ in \eqref{eq:jointOptimization} is not convex in the variables $\x_t$ and $\ll_t$, and hence the above iterative solution can potentially converge to local minima. To alleviate this to some extent, it is possible to choose a good initialization of the state and location as $\x_t^0 = A\hat{\x}_{t-1}$, and $\ll_t^0 = \hat{\ll}_{t-1}$. To initialize the tracker in the first frame, we use $\ll_0$ as the initial location marked by the user, or determined by a detector. To initialize $\x_0$, we use the pseudo-inverse computation,
\begin{equation} \label{eq:PinvInit}
\hat{\x}_0 = C^\dagger (\y_0(\ll_0) - \mu),
\end{equation}
which coincides with the maximum a-posteriori estimate of the initial state given the correct initial location and the corresponding texture at that time. A good value of the step-size $\gamma$ can be chosen using any standard step-size selection procedure \citet{Gill:OpitimizationBook} such as the Armijo step-size strategy.

We call the above method in its entirety, Dynamic Kernel SSD Tracking (\textbf{DK-SSD-T}). For ease of exposition, we have considered the case of a single kernel, however the derivation for multiple stacked and additive kernels \citet{Hager:CVPR04}, and multiple collaborative kernels with structural~constraints~\citet{Fan:PAMI07}~follows~similarly.

\subsection{Convergence analysis and parameter selection} \label{subsec:convergence}


\myparagraph{Convergence of Location} We would first like to discuss the case when the template is static, and can be represented completely using the mean, $\mu$. This is similar to the case when we can accurately synthesize the expected dynamic texture at a particular time instant before starting the tracker. In this case, our proposed approach is analogous to the original meanshift algorithm \citep{Comaniciu:PAMI03} and follows all the (local) convergence guarantees for that method. 

\myparagraph{Convergence of State} The second case, concerning the convergence of the state estimate is more interesting. In traditional filtering approaches such as the Kalman filter, the variance in the state estimator is minimized at each time instant given new observations. However, in the case of non-linear dynamical systems, the Extended Kalman Filter (EKF) only minimizes the variance of the linearized state estimate and not the actual state. Particle filters such as condensation \citep{Isard:IJCV98} usually have asymptotic convergence guarantees with respect to the number of particles and the number of time instants. Moreover efficient resampling is needed to deal with cases where all but one particle have non-zero probabilities. Our greedy cost function on the other hand aims to maximize the posterior probability of the state estimate at each time instant by assuming that the previous state is estimated correctly. This might seem like a strong assumption but as our experiments in \S\ref{sec:evaluation} will show that with the initialization techniques described earlier, we always converge to the correct state.

\myparagraph{Parameter Tuning}
The variance of the values of individual histogram bins, $\sigma_H^2$, could be empirically computed by using the EM algorithm, given kernel-weighted histograms extracted from training data. However, we fixed the value at $\sigma_H^2 = 0.01$ for all our experiments and this choice consistently gives good tracking performance. The noise parameters, $\sigma_H^2$ and $Q$, can also be analyzed as determining the relative weights of the two terms in the cost function in \eqref{eq:jointOptimization}. The first term in the cost function can be interpreted as a \emph{reconstruction} term that computes the difference between the observed kernel-weighted histogram and the \emph{predicted} kernel weighted histogram given the state of the system. The second term can similarly be interpreted as a \emph{dynamics} term that computes the difference between the current state and the predicted state given the previous state of the system, regularized by the state-noise covariance. Therefore, the values of $\sigma_H^2$ and $Q$ implictly affect the relative importance of the reconstruction term and the dynamics term in the tracking formulation. As $Q$ is computed during the system identification stage, we do not control the value of this parameter. In fact, if the noise covariance of a particular training system is large, thereby implying less robust dynamic parameters, the tracker will automatically give a low-weight to the dynamics term and a higher one to the reconstruction term.

\subsection{Invariance to Scale, Orientation and Direction of Motion} \label{subsec:scaleDirInv}

As described at the start of this section, the spatial size, $|\Omega| = r \times c$, of the dynamic template in the training video need not be the same as the size, $|\Omega^\prime| = r^\prime \times c^\prime$, of the template in the test video. Moreover, while tracking, the size of the tracked patch could change from one time instant to the next.  For simplicity, we have only considered the case where the size of the patch in the test video stays constant throughout the video. However, to account for a changing patch size, a dynamic model (\eg a random walk) for $|\Omega_t^\prime| = r_t^\prime \times c_t^\prime$, can easily be incorporated in the derivation of the optimization function in \eqref{eq:jointOptimization}. Furthermore, certain objects such as flags or human actions have a specific direction of motion, and the direction of motion in the training video need not be the same as that in the test video.

To make the tracking procedure of a learnt dynamic object invariant to the size of the selected patch, or the direction of motion, two strategies could be chosen. The first approach is to find a non-linear dynamical systems based representation for dynamic objects that is by design size and pose-invariant, \eg histograms. This would however pose additional challenges in the gradient descent scheme introduced above and would lead to increased computational complexity. The second approach is to use the proposed LDS-based representation for dynamic objects but transform the system parameters according to the observed size, orientation and direction of motion. We propose to use the second approach and transform the system parameters, $\mu$ and $C$, as required. 

Transforming the system parameters of a dynamic texture to model the transformation of the actual dynamic texture was first proposed by \cite{Ravichandran:PAMI11}, where it was noted that two videos of the same dynamic texture taken from different view points could be registered by computing the \emph{transformation between the system parameters} learnt individually from each video. To remove the basis ambiguity
\footnote{The time series, $\{\y_t\}_{t=1}^T$, can be generated by the system parameters, $(\mu, A, C, B)$, and the corresponding state sequence $\{\x_t\}_{t=1}^T$, or by system parameters $(\mu, PAP^{-1}, CP^{-1}, PB)$, and the state sequence, $\{P\x_t\}_{t=1}^T$. This inherent non-uniqueness of the system parameters given only the observed sequence is referred to as the basis ambiguity.}
, we first follow \cite{Ravichandran:PAMI11} and convert all system parameters to the Jordan Canonical Form (JCF). If the system parameters of the training video are $\mathcal{M} = (\mu, A, C, B)$ , $\mu \in \Re^{rc}$ is in fact the stacked mean template image, $\mu^{\text{im}} \in \Re^{r \times c}$. Similarly, we can imagine the matrix $C = [C_1, C_2, \ldots, C_n] \in \Re^{rc \times n}$ as a composition of basis images $C_i^{\text{im}} \in \Re^{r \times c}, i \in \{1,\ldots,n\}$.

Given an initialized bounding box around the test patch, we transform the observation parameters $\mu, C$ learnt during the training stage to the dimension of the test patch. This is achieved by computing $(\mu^\prime)^\text{im} = \mu^\text{im}(T(x))$ and $(C_i^\prime)^\text{im} = C_i^{\text{im}}(T(x)), i \in {1,\ldots,n}$, where $T(x)$ is the corresponding transformation on the image domain. For scaling, this transformation is simply an appropriate scaling of the mean image, $\mu^{\text{im}}$ and the basis images $C_i^{\text{im}}$ from $r \times c$ to $r^\prime \times c^\prime$ images using bilinear interpolation. Since the dynamics of the texture of the same type are assumed to be the same, we only need to transform $\mu$ and $C$. The remaining system parameters, $A,B$, and $\sigma_H^2$, stay the same. For other transformations, such as changes in direction of motion, the corresponding transformation $T(x)$ can be applied to the learnt $\mu,C$, system parameters before tracking. In particular, for human actions, if the change in the direction of motion is simply from left-to-right to right-to-left, $\mu^\text{im}$, and $C^\text{im}$ only need to be reflected across the vertical axis to get the transformed system parameters for the test video.

\myparagraph{A Note on Discriminative Methods} In the previous development, we have only considered foreground feature statistics. Some state-of-the-art methods also use background feature statistics and adapt the tracking framework according to changes in both foreground and background. For example, \citet{Collins:PAMI05} compute discriminative features such as foreground-to-background feature histogram ratios, variance ratios, and peak difference followed by Meanshift tracking for better performance. Methods based on tracking using classifiers \citet{Grabner:BMVC06}, \citet{Babenko:CVPR09} also build features that best discriminate between foreground and background. Our framework can be easily adapted to such a setting to provide even better performance. We will leave this as future work as our proposed method, based only on foreground statistics, already provides results similar to or better than the state-of-the-art.

\section{Tracking and Recognition of Dynamic Templates} \label{sec:jointTrackingRecognition}

The proposed generative approach presented in \S\ref{sec:jointFramework} has another advantage. As we will describe in detail in this section, we can use the value of the objective function in \eqref{eq:jointOptimization} to perform simultaneous tracking and recognition of dynamic templates. Moreover, we can learn the LDS parameters of the tracked dynamic template from the corresponding bounding boxes and compute a system-theoretic distance to all the LDS parameters from the training set. This distance can then be used as a discriminative cost to simultaneously provide the best tracks and class label in a test video.
%
%
In the following we propose three different approaches for tracking and recognition of dynamic templates using the tracking approach presented in the previous section at their core.

\myparagraph{Recognition using tracking objective function} The dynamic template tracker computes the optimal location and state estimate at each time instant by minimizing the objective function in \eqref{eq:jointOptimization} given the system parameters, $\mathcal{M}=(\mu,A,C,B)$, of the dynamic template, and the kernel histogram variance, $\sigma_H^2$. From here on, we will use the more general expression, $\mathcal{M} = (\mu, A, C, Q, R)$, to describe all the tracker parameters, where $Q=BB^\top$ is the covariance of the state noise process and $R$ is the covariance of the observed image function, \eg in our proposed kernel-based framework, $R = \sigma_H^2 I$. Given system parameters for multiple dynamic templates, for example, multiple sequences of the same dynamic texture, or sequences belonging to different classes, we can track a dynamic texture in a test video using each of the system parameters. For each tracking experiment, we will obtain location and state estimates for each time instant as well as the value of the objective function at the computed minima. At each time instant, the objective function value computes the value of the negative logarithm of the posterior probability of the location and state given the system parameters. We can therefore compute the average value of the objective function across all frames and use this \emph{dynamic template reconstruction cost} as a measure of how close the observed dynamic template tracks are to the model of the dynamic template used to track it.

More formally, given a set of training system parameters, $\{\mathcal{M}_i\}_{i=1}^N$ corresponding to a set of training videos with dynamic templates with class labels, $\{\mathcal{L}_i\}_{i=1}^N$, and test sequence $j \ne i$, we compute the optimal tracks and states, $\{(\hat{\ll}_t^{(j,i)}, \hat{\x}_t^{(j,i)})\}_{t=1}^{T_j}$ for all $i \in {1,\ldots,N}, j \ne i$ and the corresponding objective function values,
\begin{align}
\bar{O}_j^i = \dfrac{1}{T_j} \dsum_{t=1}^{T_j} O_j^i(\hat{\ll}_t^{(j,i)}, \hat{\x}_t^{(j,i)}), \label{eq:jointOptimizationAllFramesGivenTrain}
\end{align}
where $O_j^i(\hat{\ll}_t^{(j,i))}, \hat{\x}_t^{(j,i)})$ represents the value of the objective function in \eqref{eq:jointOptimization} computed at the optimal $\hat{\ll}_t^{(j,i)}$ and $\hat{\x}_t^{(j,i)}$, when using the system parameters $\mathcal{M}_i = (\mu_i, A_i, C_i, Q_i, R_i)$ corresponding to training sequence $i$ and tracking the template in sequence $j \ne i$. The value of the objective function represents the \emph{dynamic template reconstruction cost} for the test sequence, $j$, at the computed locations $\{\ll_t^{(j,i)}\}_{t=1}^{T_j}$ as modeled by the dynamical system $\mathcal{M}_i$. System parameters that correspond to the dynamics of the same class as the observed template should therefore give the smallest objective function value whereas those that correspond to a different class should give a greater objective function value. Therefore, we can also use the value of the objective function as a classifier to simultaneously determine the class of the dynamic template as well as its tracks as it moves in a scene. The dynamic template class label is hence found as $\mathcal{L}_j = \mathcal{L}_k$, where $k = \argmin_i \bar{O}_j^i$, \ie the label of the training sequence with the minimum objective function value. The corresponding tracks $\{\hat{\ll}_t^{(j,k)}\}_{t=1}^{T_j}$ are used as the final tracking result. Our tracking framework therefore allows us to perform simultaneous tracking and recognition of dynamic objects. We call this method for simultaneously tracking and recognition using the objective function value, Dynamic Kernel SSD - Tracking and Recognition using Reconstruction (\textbf{DK-SSD-TR-R}). 

\myparagraph{Tracking then recognizing} In a more traditional dynamic template recognition framework, it is assumed that the optimal tracks, $\{\hat{\ll}_t^j\}_{t=1}^{T_j}$ for the dynamic template have already been extracted from the test video. Corresponding to these tracks, we can extract the sequence of bounding boxes, $Y_j = \{\y_t(\hat{\ll}_t^j)\}_{t=1}^{T_j}$, and learn the system parameters, $\mathcal{M}_j$ for the tracked dynamic template using the approach described in \S\ref{subsec:sysID}. We can then compute a \emph{distance} between the test dynamical system, $\mathcal{M}_j$, and all the training dynamical systems, $\{\mathcal{M}_i\}, i = 1 \ldots N, j \ne i$.  A commonly used distance between two linear dynamical systems is the Martin distance, \cite{DeCockDeMoor:SCL02}, that is based on the subspace angles between the observability subspaces of the two systems. The Martin distance has been shown, \eg~in \citep{Chaudhry:TR09, Doretto:IJCV03, Bissacco:CVPR01}, to be discriminative between dynamical systems belonging to several different classes. We can therefore use the Martin distance with Nearest Neighbors as a classifier to recognize the test dynamic template by using the optimal tracks, $\{\hat{\ll}_t^{(j,k)}\}_{t=1}^{T_j}$, from the first method, DK-SSD-TR-R. We call this tracking \emph{then} recognizing method Dynamic Kernel SSD - Tracking then Recognizing (\textbf{DK-SSD-T+R}).

\myparagraph{Recognition using LDS distance classifier} As we will show in the experiments, the \emph{reconstruction cost} based tracking and recognition scheme, DK-SSD-TR-R, works very well when the number of classes is small. However, as the number of classes increases, the classification accuracy decreases. The objective function value itself is in fact not a very good classifier with many classes and high inter class similarity. Moreover, the tracking \emph{then} recognizing scheme, DK-SSD-T+R, disconnects the tracking component from the recognition part. It is possible that tracks that are slightly less optimal according to the objective function criterion may in fact be better for classification. To address this limitation, we propose to add a \emph{classification cost} to our original objective function and use a two-step procedure that computes a distance between the dynamical template as observed through the tracked locations in the test video and the actual training dynamic template. This is motivated by the fact that if the tracked locations in the test video are correct, and a dynamical-systems based distance between the observed template in the test video and the training template is small, then it is highly likely that the tracked dynamic texture in the test video belongs to the same class as the training video. Minimizing a \emph{reconstruction} and \emph{classification} cost will allow us to simultaneously find the best tracks and the corresponding label of the dynamic template.

Assume that with our proposed gradient descent scheme in \eqref{eq:gradDescent}, we have computed the optimal tracks and state estimates, $\{\hat{\ll}_t^{(j,i)}, \hat{\x}_t^{(j,i)}\}_{t=1}^{T_j}$, for all frames in test video $j$, using the system parameters corresponding to training dynamic template $\mathcal{M}_i$. As described above, we can then extract the corresponding tracked regions, $Y_j^i =\{\y_t(\hat{\ll}_t^{(j,i)})\}_{t=1}^{T_j}$, and learn the system parameters $\mathcal{M}_j^i = (\mu_j^i, A_j^i, C_j^i, Q_j^i)$ using, the system identification method in \S\ref{subsec:sysID}. If the dynamic template in the observed tracks, $\mathcal{M}_j^i$, is similar to the training dynamic template, $\mathcal{M}_i$, then the \emph{distance} between $\mathcal{M}_j^i$ and $\mathcal{M}_i$ should be small. Denoting the Martin distance between two dynamical systems, $\cM_1, \cM_2$ as $d_M(\cM_1,\cM_2)$, we propose to use the \emph{classification cost},
\begin{align}\label{eq:classificationCost}
\mathcal{C}_j^i = d_M(\cM_j^i, \cM_i).
\end{align}
Specifically, we classify the test video as $\cL_j = \cL_k$, where $k=\argmin_i \cC_j^i$, and use the extracted tracks, $\{\hat{\ll}_t^{(j,k)}\}_{t=1}^{T_j}$ as the final tracks. We call this method for simultaneous tracking and recognition using a classification cost as Dynamic-Kernel SSD - Tracking and Recognition using Classifier (\textbf{DK-SSD-TR-C}). As we will show in our experiments, DK-SSD-TR-C gives state-of-the-art results for tracking and recognition of dynamic templates.

\begin{figure*}
\begin{minipage}[t]{0.48\linewidth}
\centering
\includegraphics[width=\linewidth]{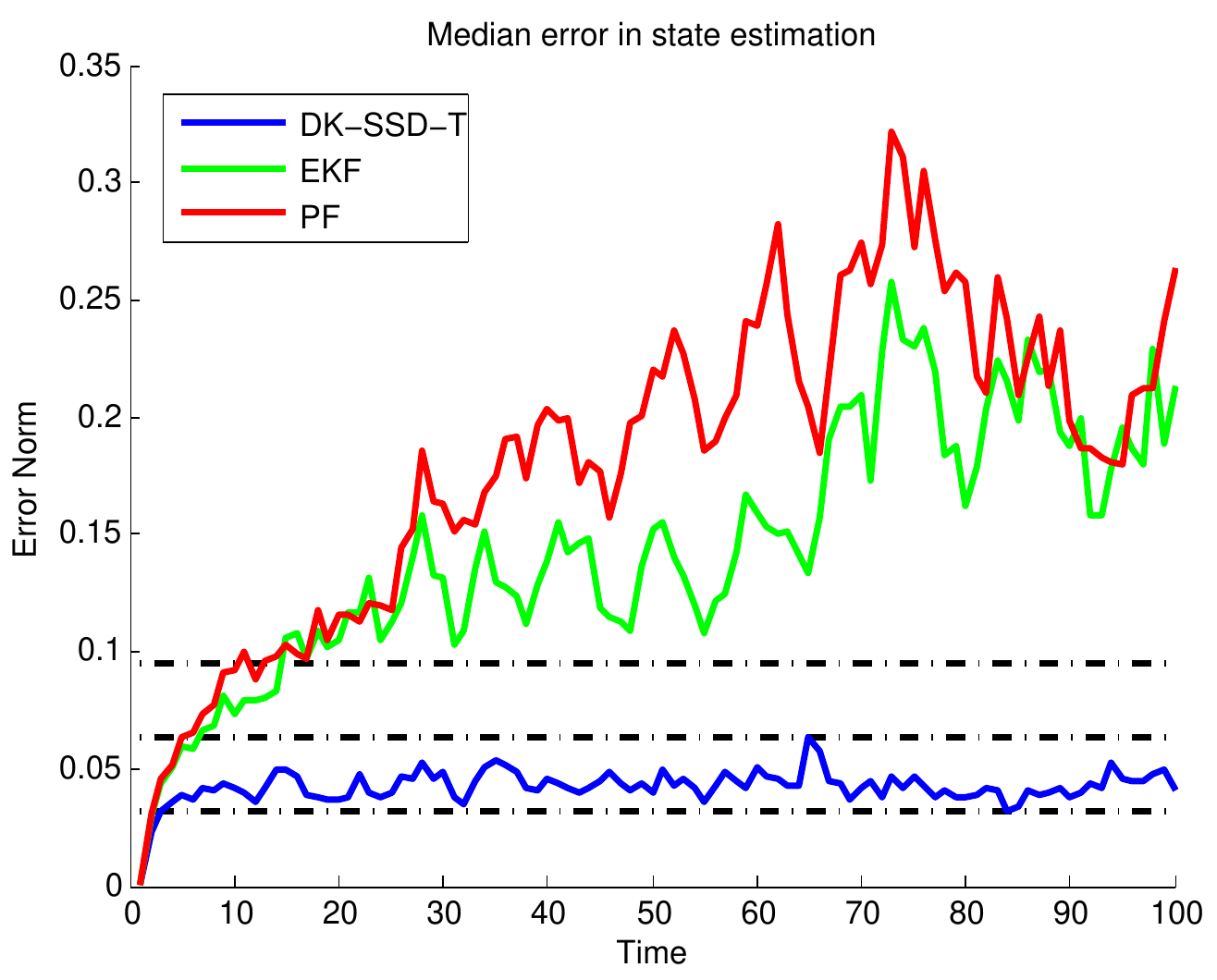}
\caption{Median state estimation error across 10 randomly generated dynamic textures with the initial state computed using the pseudo-inverse method in \eqref{eq:PinvInit} using our proposed, DK-SSD-T (blue), Extended Kalman Filter (EKF)  (green), and Condensation Particle Filter (PF) (red). The horizontal dotted lines represent 1-, 2-, and 3-standard deviations for the norm of the state noise.}
\label{fig:stateErrPinvInitRandSys}
\end{minipage}
\hfill
\begin{minipage}[t]{0.48\linewidth}
\centering
\includegraphics[width=\linewidth]{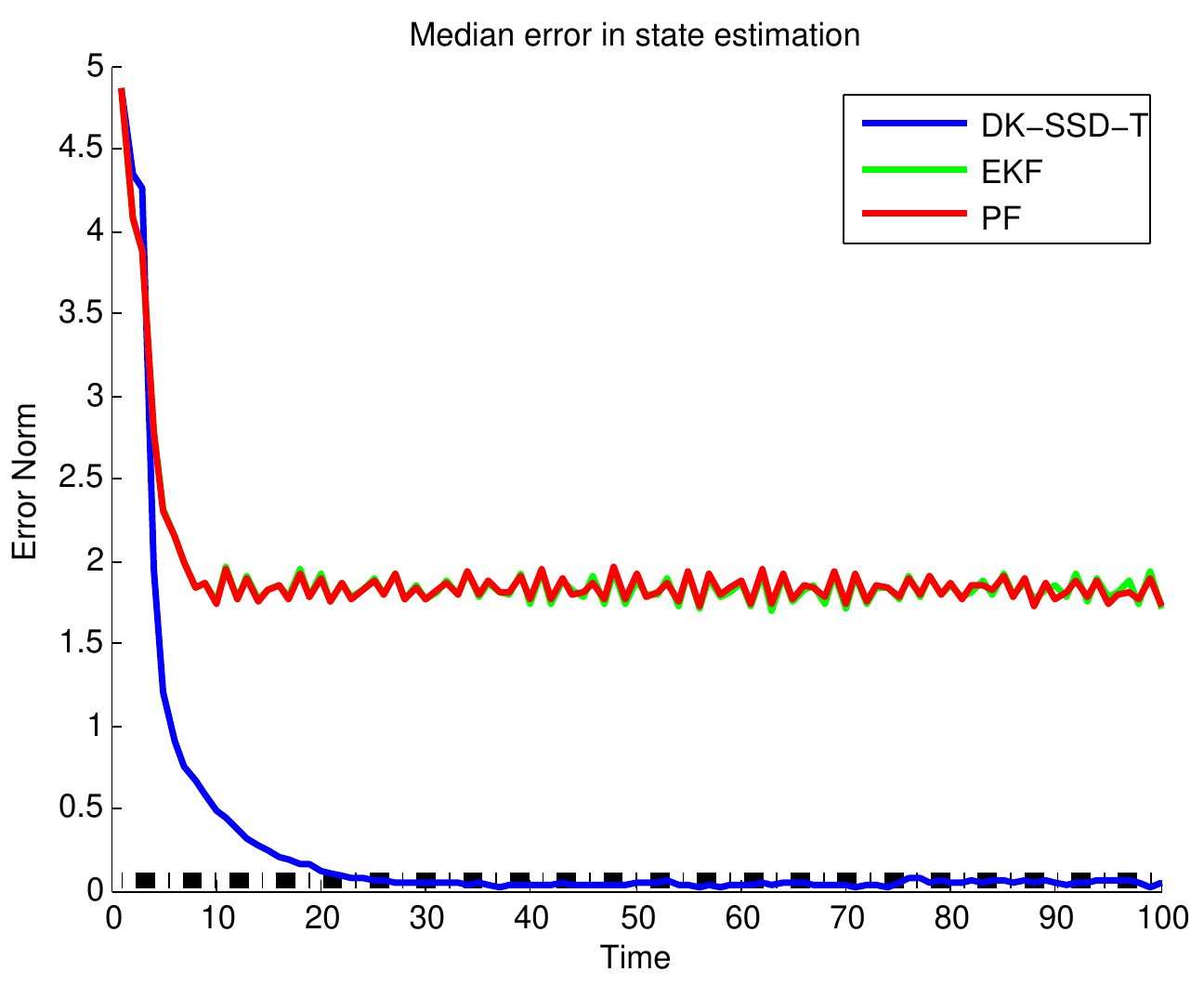}
\caption{Median state-error for 10 random initializations for the initial state when estimating the state of the same dynamic texture, using our proposed, DK-SSD-T (blue), Extended Kalman Filter (EKF)  (green), and Condensation Particle Filter (PF) (red). The horizontal dotted lines represent 1-, 2-, and 3-standard deviations for the norm of the state noise.}
\label{fig:stateErrRandInit}
\end{minipage}
\end{figure*}
\begin{figure*}
\centering
\subfigure[DK-SSD-T]{
\includegraphics[width=0.31\linewidth]{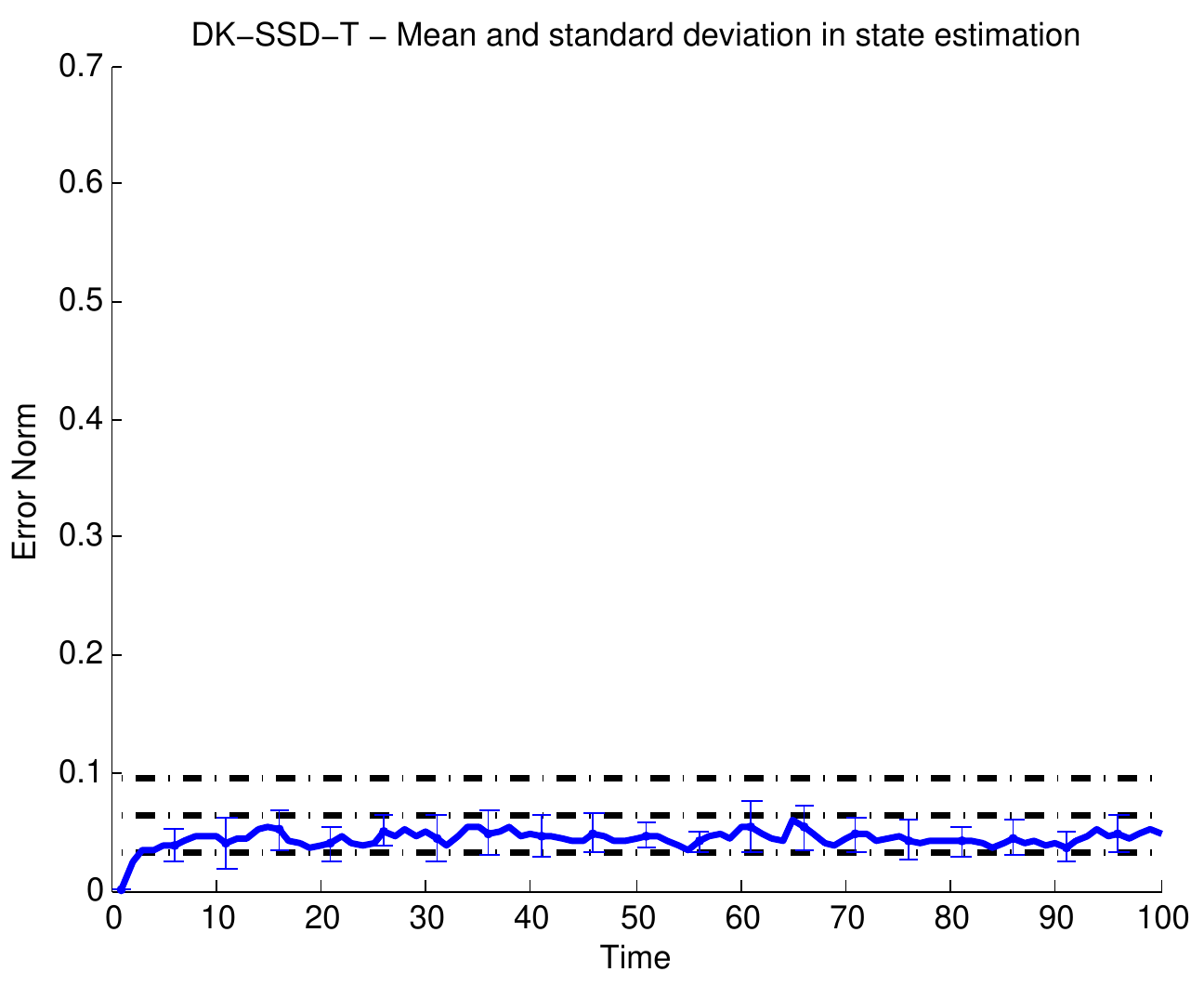}
\label{fig:stateErrPinvInitRandSysMeanSD-DKSSD}
}
\subfigure[Extended Kalman Filter]{
\includegraphics[width=0.31\linewidth]{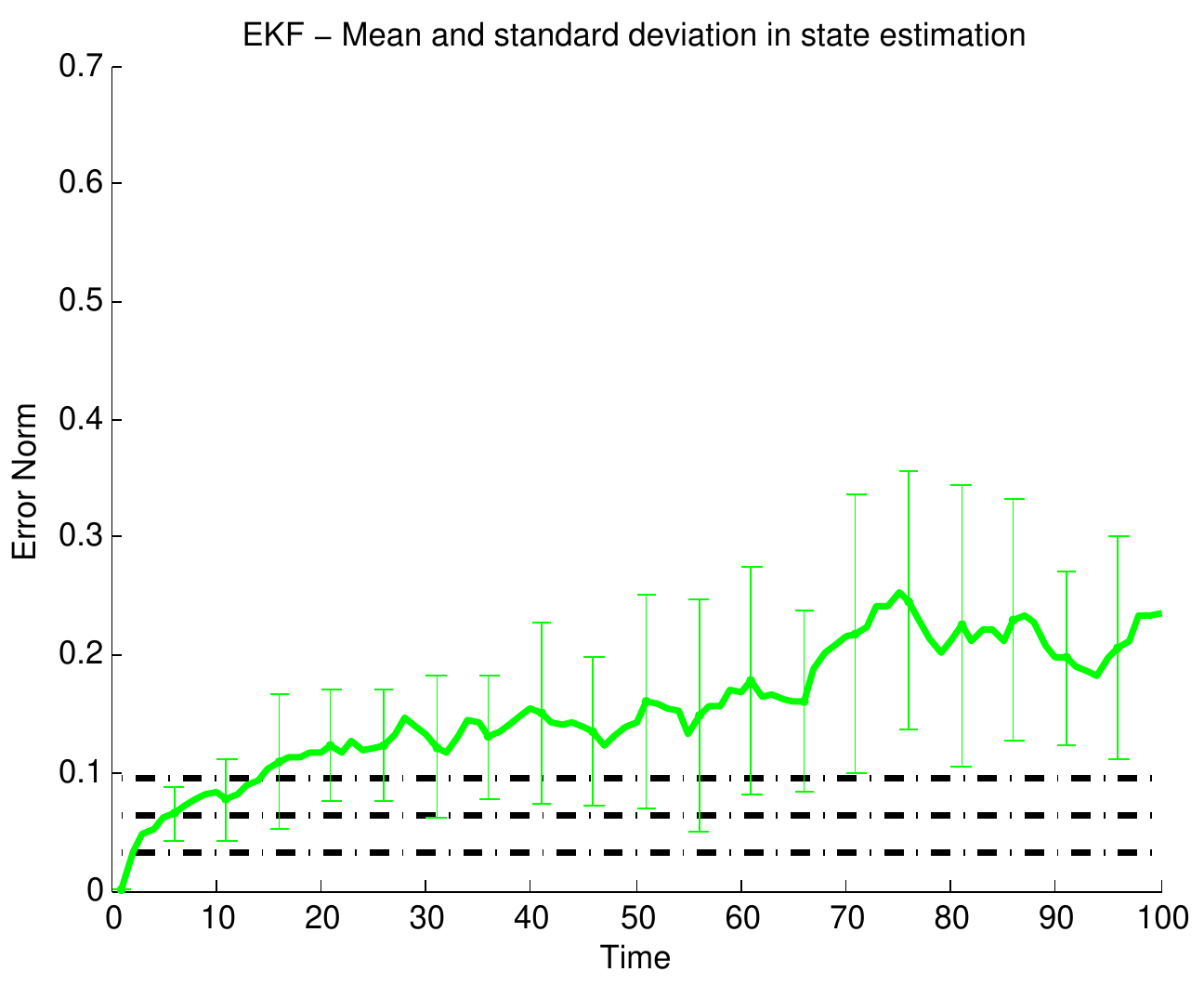}
\label{fig:stateErrPinvInitRandSysMeanSD-EKF}
}
\subfigure[Particle Filter]{
\includegraphics[width=0.31\linewidth]{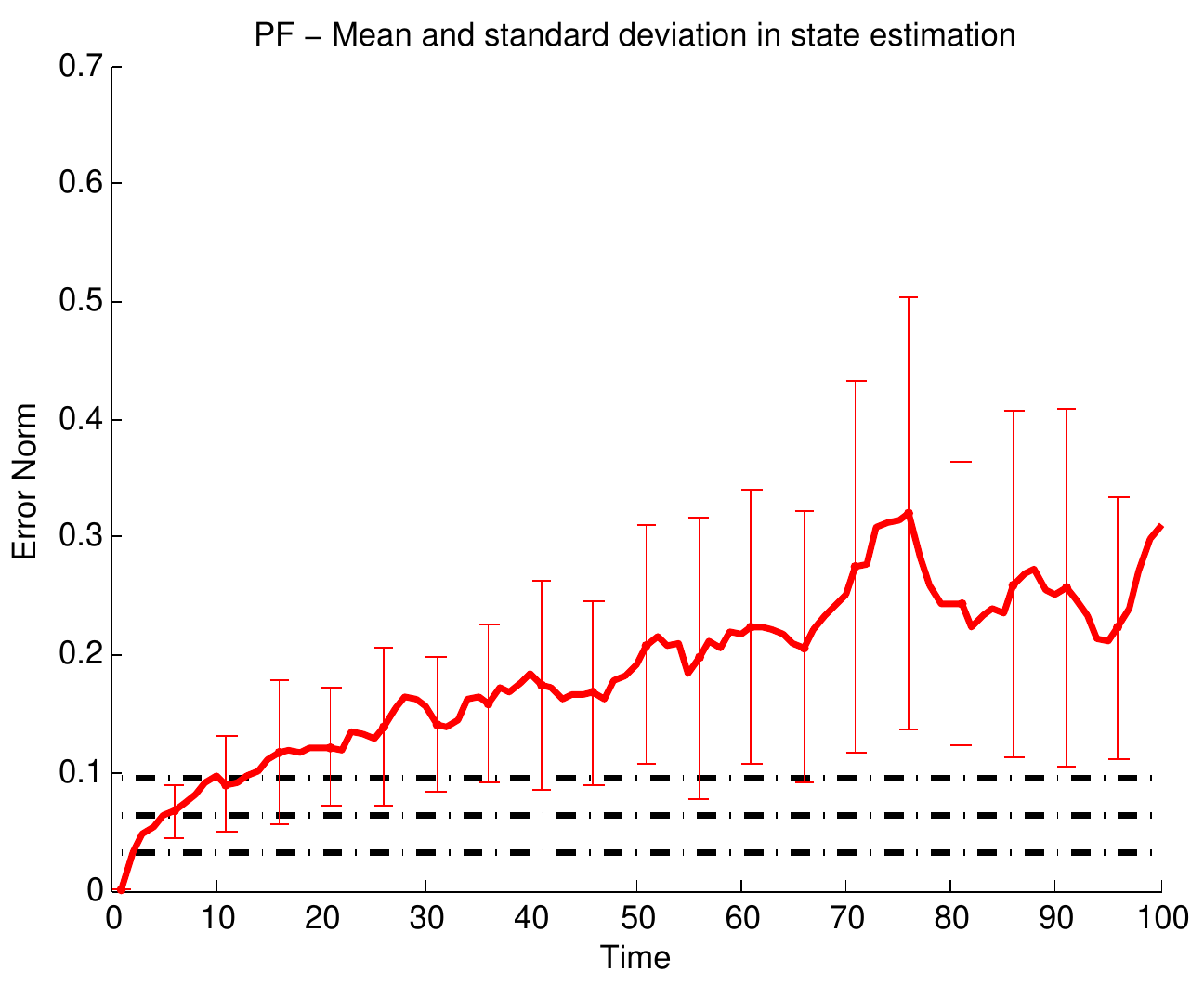}
\label{fig:stateErrPinvInitRandSysMeanSD-PF}
}
\caption{Mean and 1-standard deviation of state estimation errors for different algorithms across 10 randomly generated dynamic textures with the initial state computed using the pseudo-inverse method in \eqref{eq:PinvInit}. These figures correspond to the median results shown in \figref{fig:stateErrPinvInitRandSys}.}
\label{fig:convergenceResultsPinvInit}
\end{figure*}

\begin{figure*}
\centering
\subfigure[DK-SSD-T]{
\includegraphics[width=0.31\linewidth]{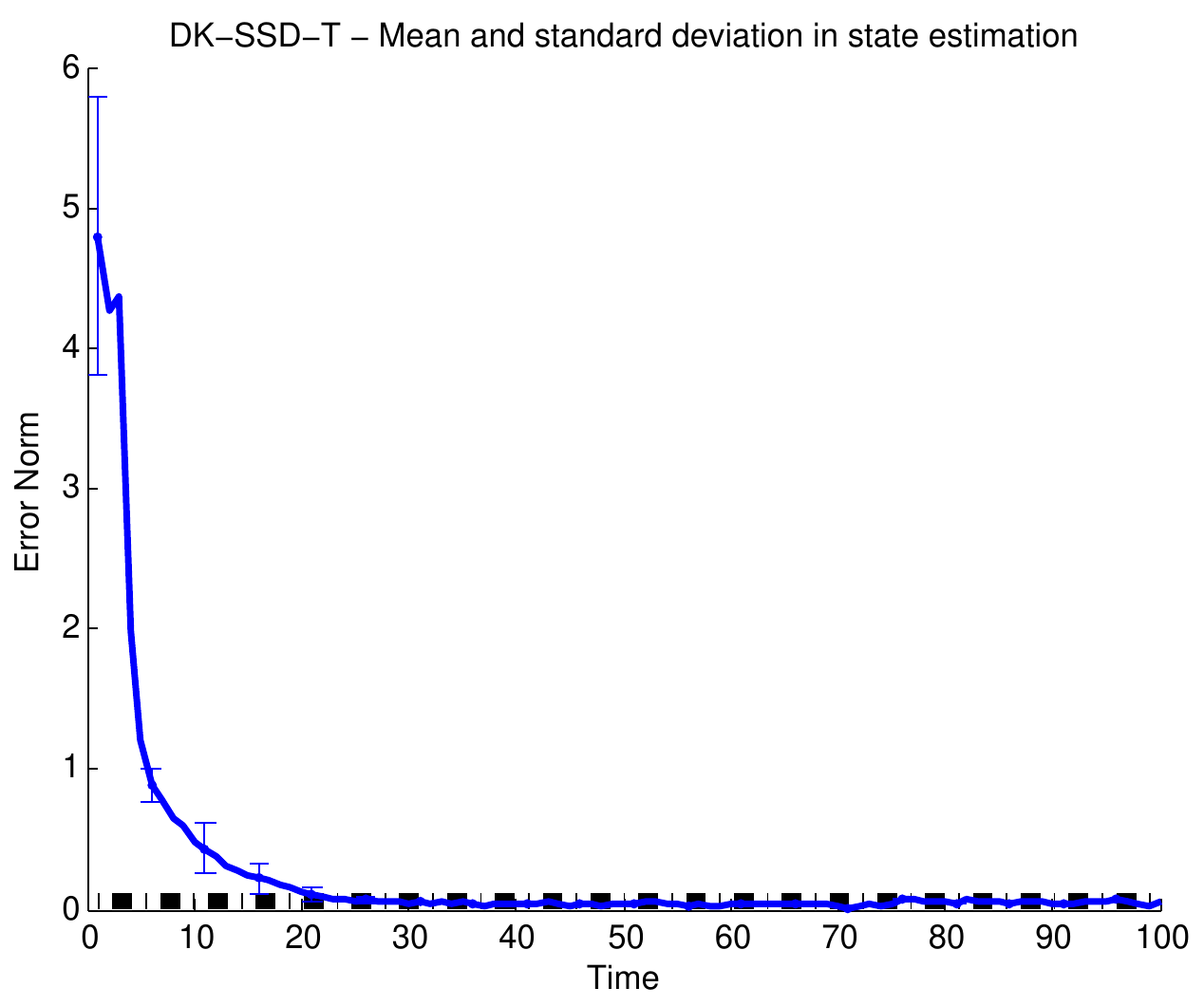}
\label{fig:stateErrRandInitMeanSD-DKSSD}
}
\subfigure[Extended Kalman Filter]{
\includegraphics[width=0.31\linewidth]{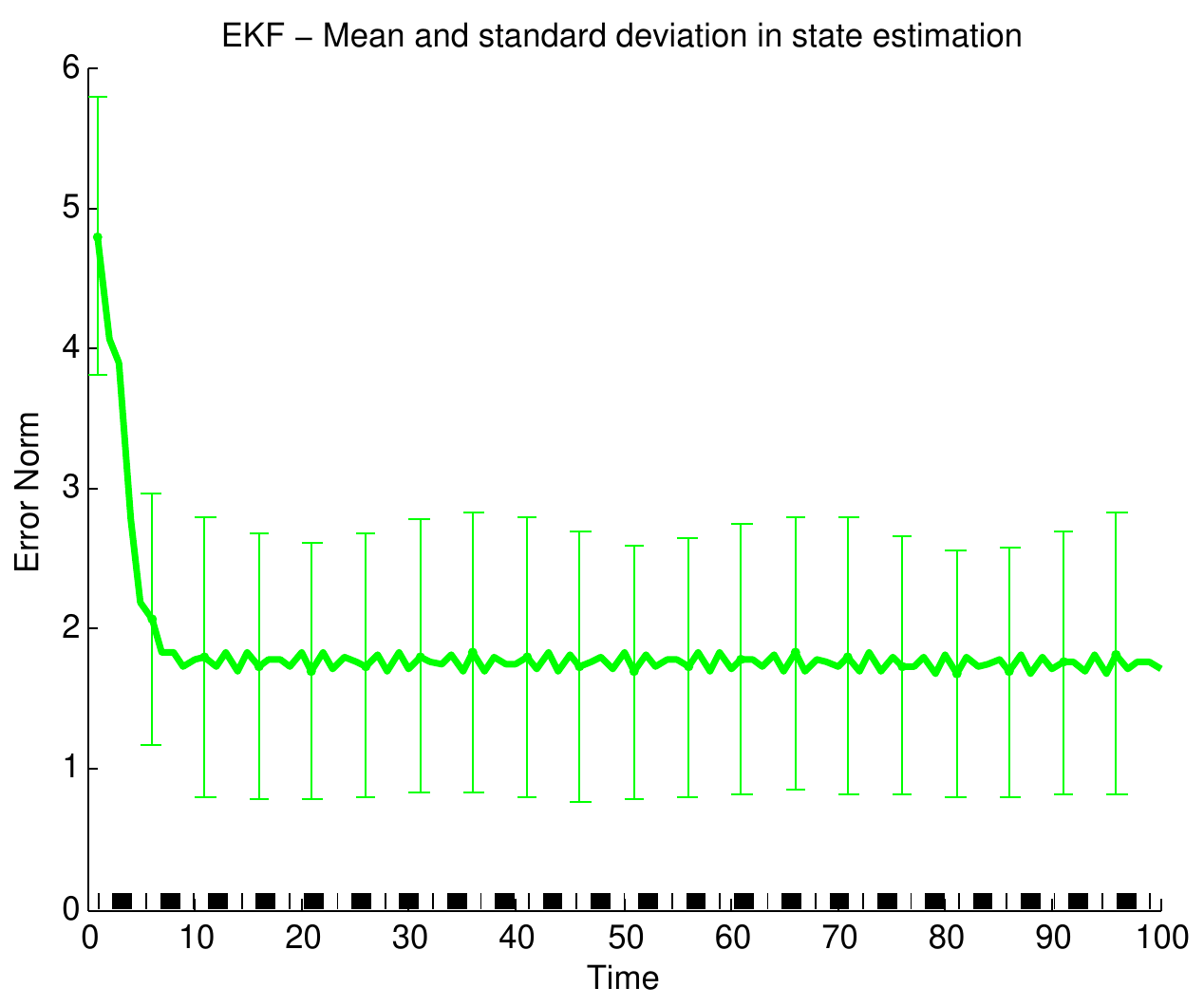}
\label{fig:stateErrRandInitRandSysMeanSD-EKF}
}
\subfigure[Particle Filter]{
\includegraphics[width=0.31\linewidth]{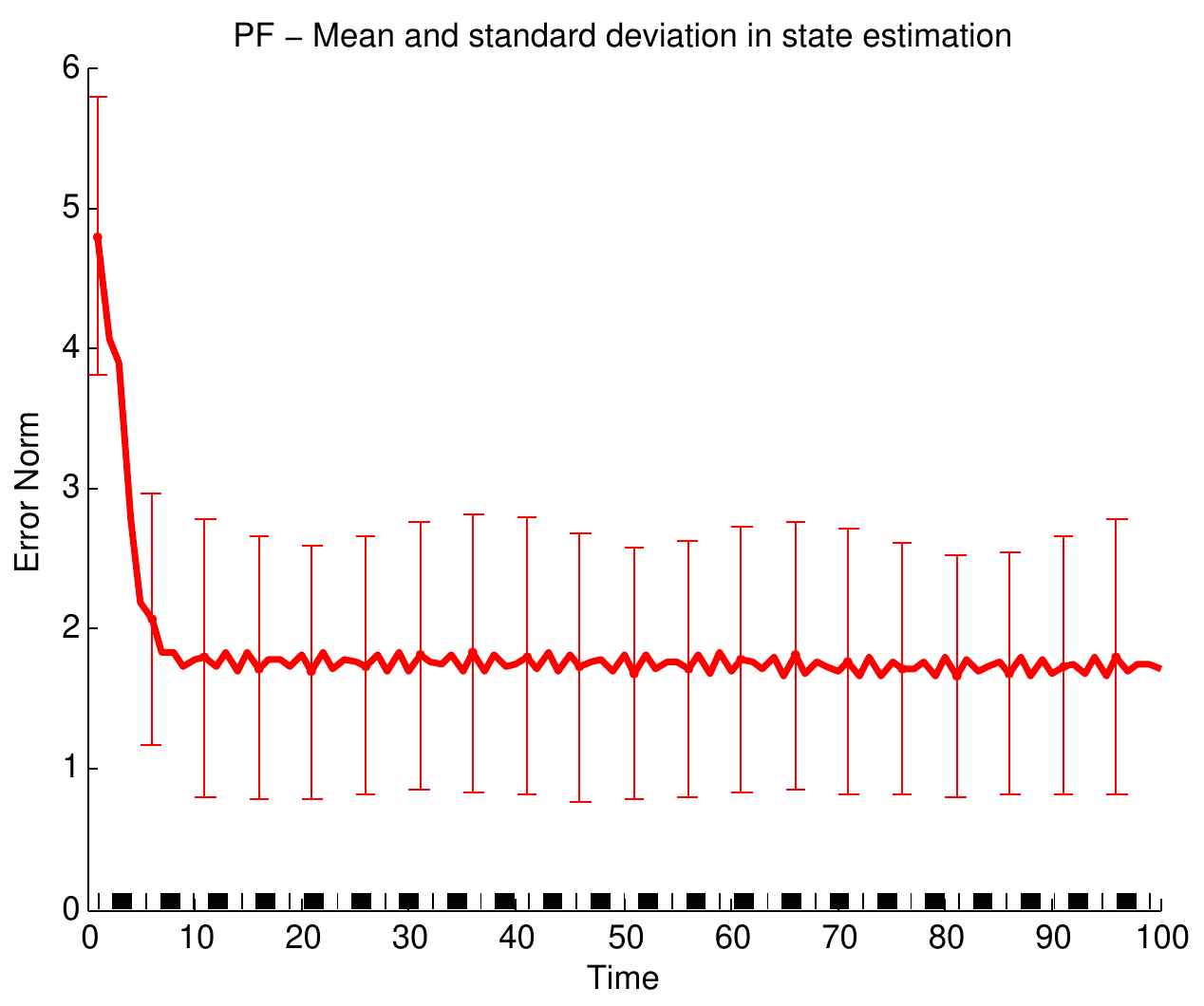}
\label{fig:stateErrRandInitMeanSD-PF}
}
\caption{Mean and 1-standard deviation of state estimation errors for different algorithms 10 random initializations for the initial state when estimating the state of the same dynamic texture. These figures correspond to the median results shown in \figref{fig:stateErrRandInit}.}
\label{fig:convergenceResultsRandInit}
\end{figure*}

\section{Empirical evaluation of state convergence}\label{sec:evaluation}

In \S\ref{subsec:convergence}, we discussed the convergence properties of our algorithm. Specifically, we noted that if the template was static or the true output of the dynamic template were known, our proposed algorithm is equivalent to the standard meanshift tracking algorithm. In this section, we will numerically evaluate the convergence of the state estimate of the dynamic template.

We generate a random synthetic dynamic texture with known system parameters and states at each time instant. We then fixed the location of the texture and assumed it was known a-priori thereby reducing the problem to only the estimation of the state given correct measurements. This is also the common scenario for state-estimation in controls theory. \figref{fig:stateErrPinvInitRandSys} shows the median error in state estimation for 10 randomly generated dynamic textures using the initial state computation method in \eqref{eq:PinvInit} in each case. For each of the systems, we estimated the state using our proposed method, Dynamic Kernel SSD Tracking (DK-SSD-T) shown in blue, Extended Kalman Filter (EKF) shown in green, and Condensation Particle Filter (PF), with 100 particles, shown in red, using the same initial state. Since the state, $\x_t$ is driven by stochastic inputs with covariance $Q$, we also display horizontal bars depicting 1-, 2-, and 3-standard deviations of the norm of the noise process to measure the accuracy of the estimate. As we can see, at all time-instants, the state estimation error using our method remains within 1- and 2-standard deviations of the state noise. The error for both EKF and PF, on the other hand, increases with time and becomes much larger than 3-standard deviations of the noise process. 

\figrefs{fig:stateErrPinvInitRandSysMeanSD-DKSSD}{fig:stateErrPinvInitRandSysMeanSD-PF} show the mean and standard deviation of the state estimates across all ten runs for DK-SSD-T, EKF and PF respectively. As we can see, our method has a very small standard deviation and thus all runs convege to within 1- and 2-standard deviations of the noise process norm. EKF and PF on the other hand, not only diverge from the true state but the variance in the state estimates also increases with time, thereby making the state estimates very unreliable. This is because our method uses a gradient descent scheme with several iterations to look for the (local) minimizer of the exact objective function, whereas the EKF only uses a linear approximation to the system equations at the current state and does not refine the state estimate any further at each time-step. With a finite number of samples, PF also fails to converge. This leads to a much larger error in the EKF and PF. The trade-off for our method is its computational complexity. Because of its iterative nature, our algorithm is computationally more expensive as compared to EKF and PF. On average it requires between 25 to 50 iterations for our algorithm to converge to a state estimate.

Similar to the above evaluation, \figref{fig:stateErrRandInit} shows the error in state estimation, for 10 different random initializations of the initial state, $\hat{\x}_0$, for one specific dynamic textures. As we can see, the norm of the state error is very large initially, but for our proposed method, as time proceeds the state error converges to below the state noise error. However the state error for EKF and PF remain very large. \figrefs{fig:stateErrRandInitMeanSD-DKSSD}{fig:stateErrRandInitMeanSD-PF} show the mean and standard deviation bars for the state estimation across all 10 runs. Again, our method converges for all 10 runs, whereas the variance of the state-error is very large for both EKF and PF. These two experiments show that choosing the initial state using the pseudo-inverse method results in very good state estimates. Moreover, our approach is robust to incorrect state initializations and will eventually converge to the correct state with under 2 standard deviations of the state noise.

In summary, the above evaluation shows that 
even though our method is only guaranteed to converge to a local minimum when estimating the internal state of the system, in practice, it performs very well and always converges to an error within two standard deviations of the state noise. Moreover, our method is robust to incorrect state initializations. Finally, since our method iteratively refines the state estimate at each time instant, it performs much better than traditional state estimation techniques such as the EKF and PF.

\section{Experiments on Tracking Dynamic Textures} \label{sec:DTExperiments}

We will now test our proposed algorithm on several synthetic and real videos with moving dynamic textures and demonstrate the tracking performance of our algorithm against the state-of-the-art. The full videos of the corresponding results can be found at \url{http://www.cis.jhu.edu/~rizwanch/dynamicTrackingIJCV11} using the figure numbers in this paper.

We will also compare our proposed dynamic template tracking framework against traditional kernel-based tracking methods such as Meanshift \citep{Comaniciu:PAMI03}, as well as the improvements suggested in \citet{Collins:PAMI05} that use features such as histogram ratio and variance ratio of the foreground versus the background before applying the standard Meanshift algorithm. We use the publicly available VIVID Tracking Testbed\footnote{\url{http://vision.cse.psu.edu/data/vividEval/}} \citep{Collins:PETS05} for implementations of these algorithms. We also compare our method against the publicly available\footnote{\url{http://www.vision.ee.ethz.ch/boostingTrackers/index.htm}} Online Boosting algorithm first proposed in \citet{Grabner:BMVC06}. As mentioned in the introduction, the approach presented in \citet{Peteri:MVA10} also addresses dynamic texture tracking using optical flow methods. Since the authors of \cite{Peteri:MVA10} were not able to provide their code, we implemented their method on our own to perform a comparison. We would like to point out that despite taking a lot of care while implementing, and getting in touch with the authors several times, we were not able to get the same results as those shown in \cite{Peteri:MVA10}. However, we are confident that our implementation is correct and besides specific parameter choices, accurately follows the approach proposed in \cite{Peteri:MVA10}. Finally, for a fair comparison between several algorithms, we did not use color features and we were able to get very good tracks without using any color information.  

For consistency, tracks for Online Boosting (\textbf{Boost}) are shown in magenta, Template Matching (\textbf{TM}) in yellow, Meanshift (\textbf{MS}) in black, Meanshift with Variance Ratio (\textbf{MS-VR}) and Histogram Ratio (\textbf{MS-HR}) in blue and red respectively. Tracks for Particle Filtering for Dynamic Textures (\textbf{DT-PF}) are shown in light brown, and the optimal tracks for our method, Dynamic Kernel SSD Tracking (\textbf{DK-SSD-T}), are shown in cyan whereas any ground truth is shown in green. To better illustrate the difference in the tracks, we have zoomed in to the active portion of the video.

\begin{figure*}[tb]
\centering
\subfigure{
\includegraphics*[width=0.23\linewidth, viewport=0 204 438 492]{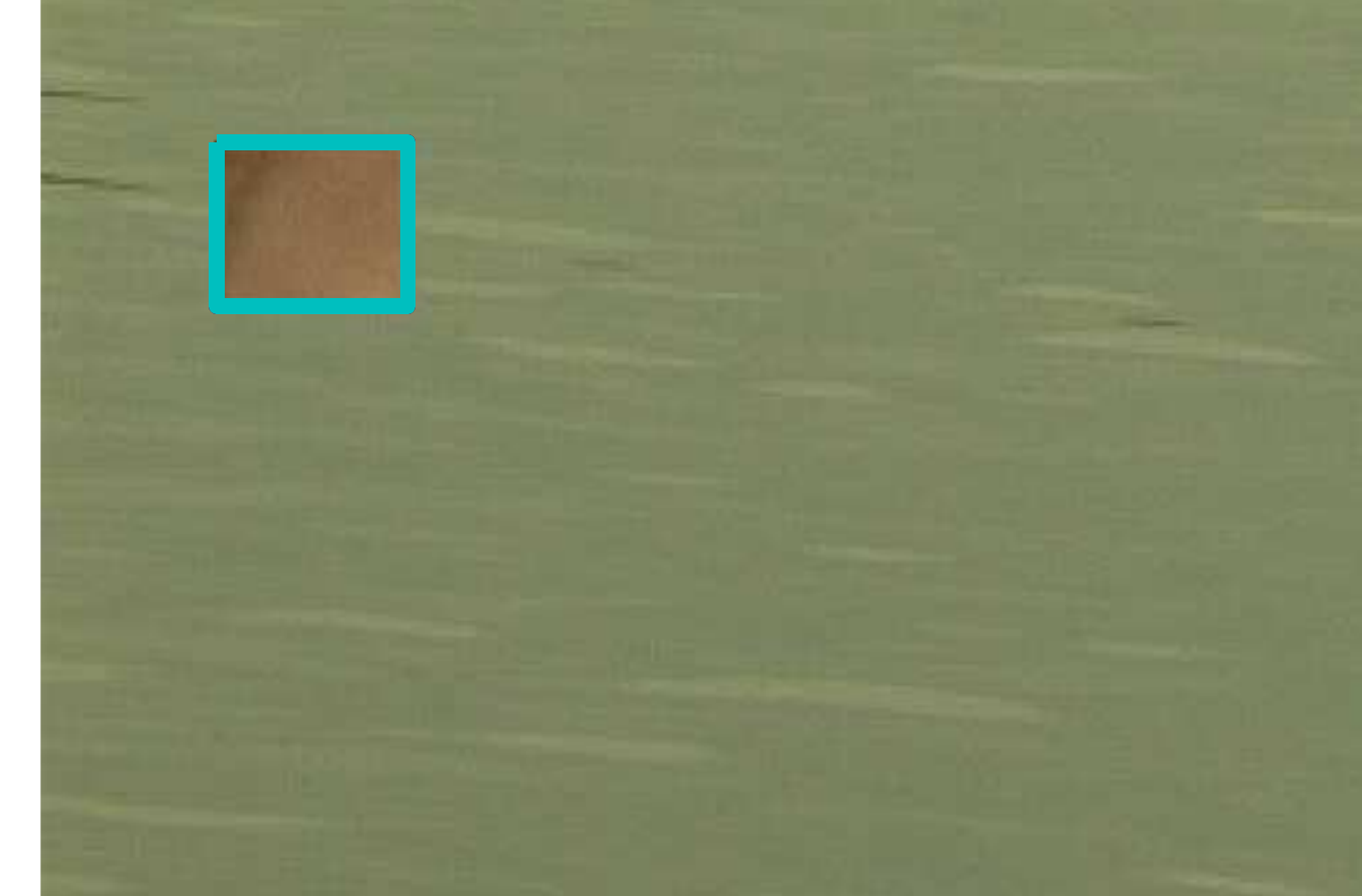}
\label{fig:dyntex1-frame1}
}
\subfigure{
\includegraphics*[width=0.23\linewidth, viewport=0 204 438 492]{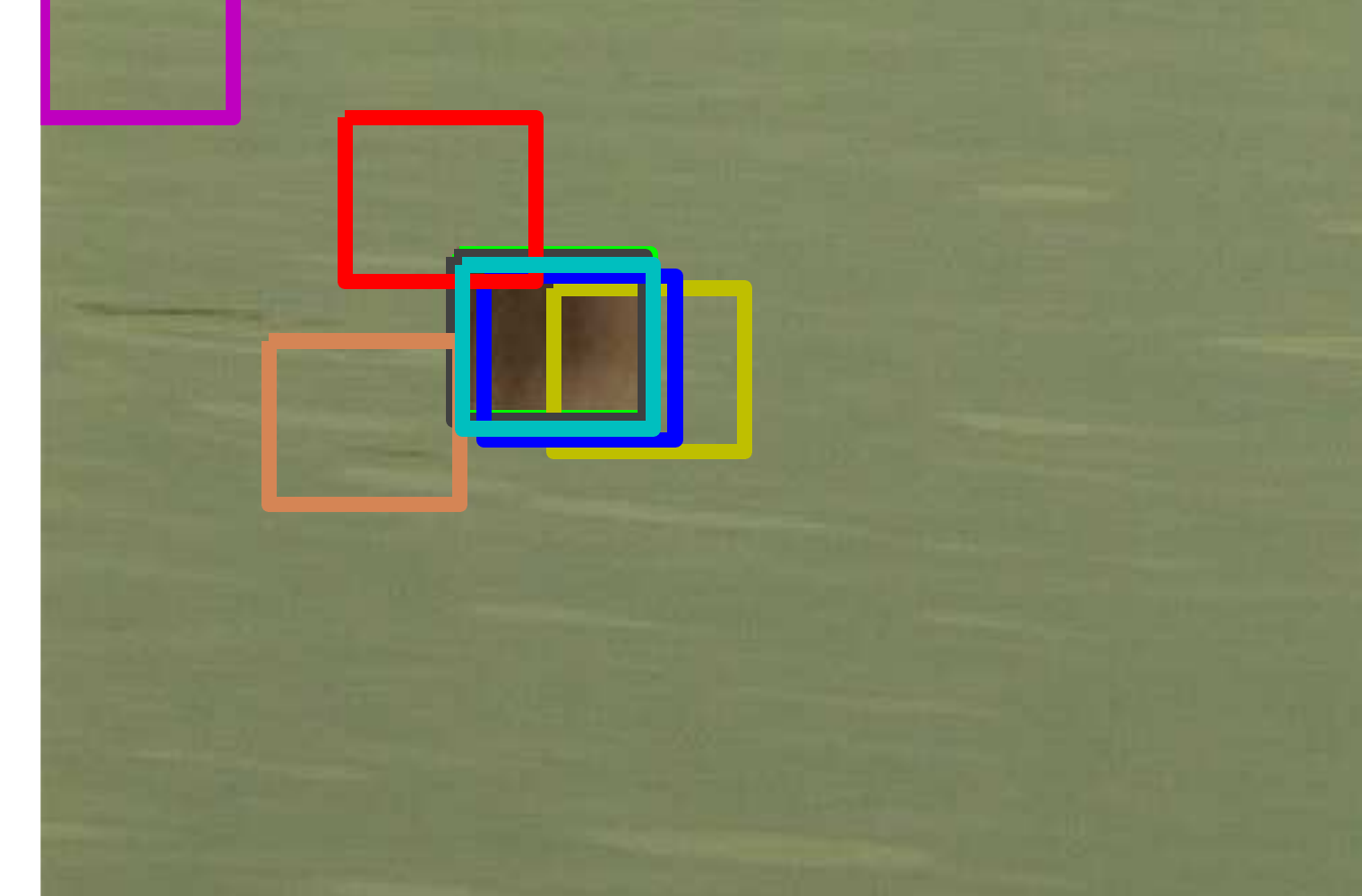}
\label{fig:dyntex1-frame2}
}
\subfigure{
\includegraphics*[width=0.23\linewidth, viewport=0 204 438 492]{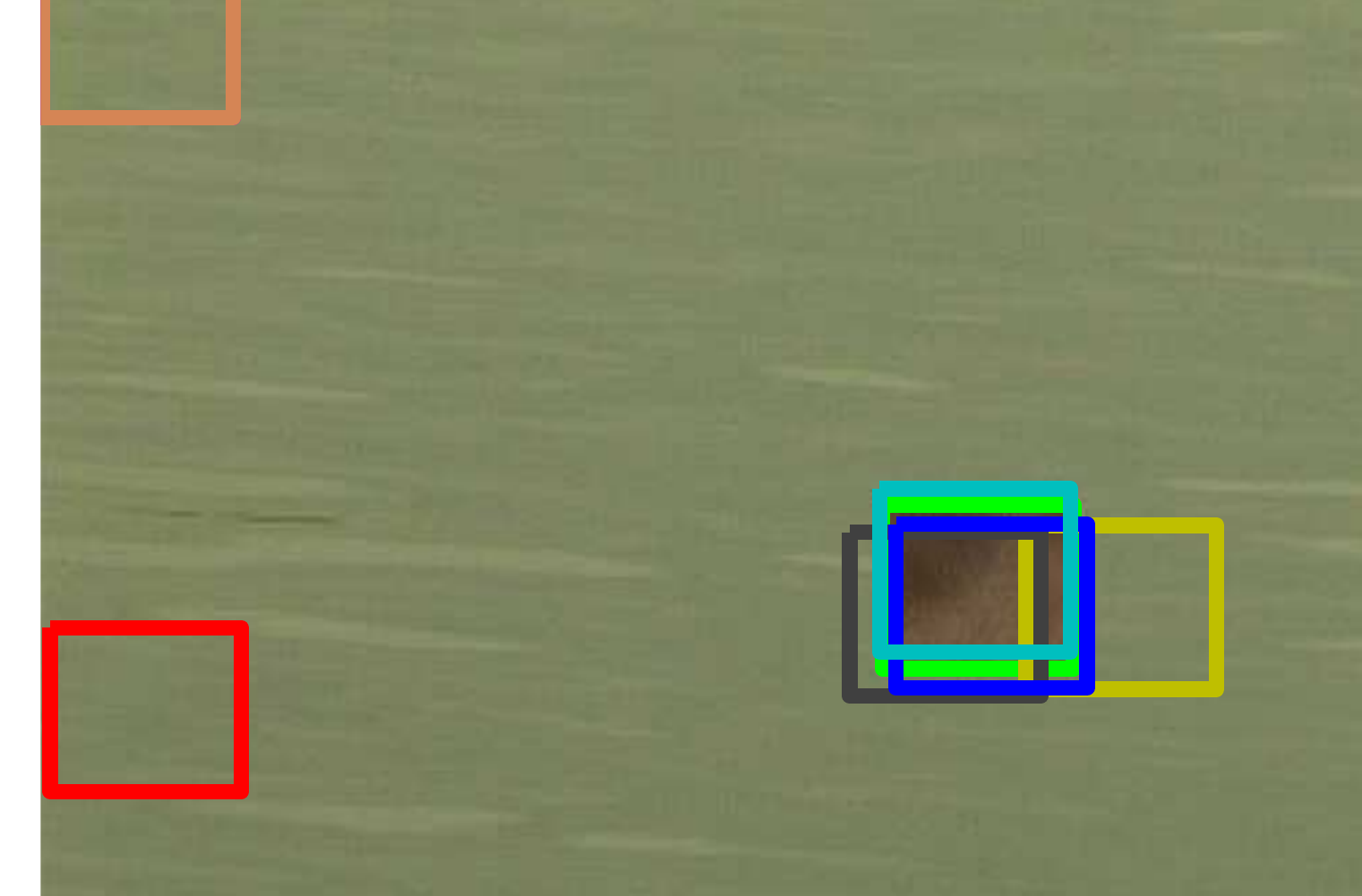}
\label{fig:dyntex1-frame3}
}
\subfigure{
\includegraphics*[width=0.23\linewidth, viewport=0 204 438 492]{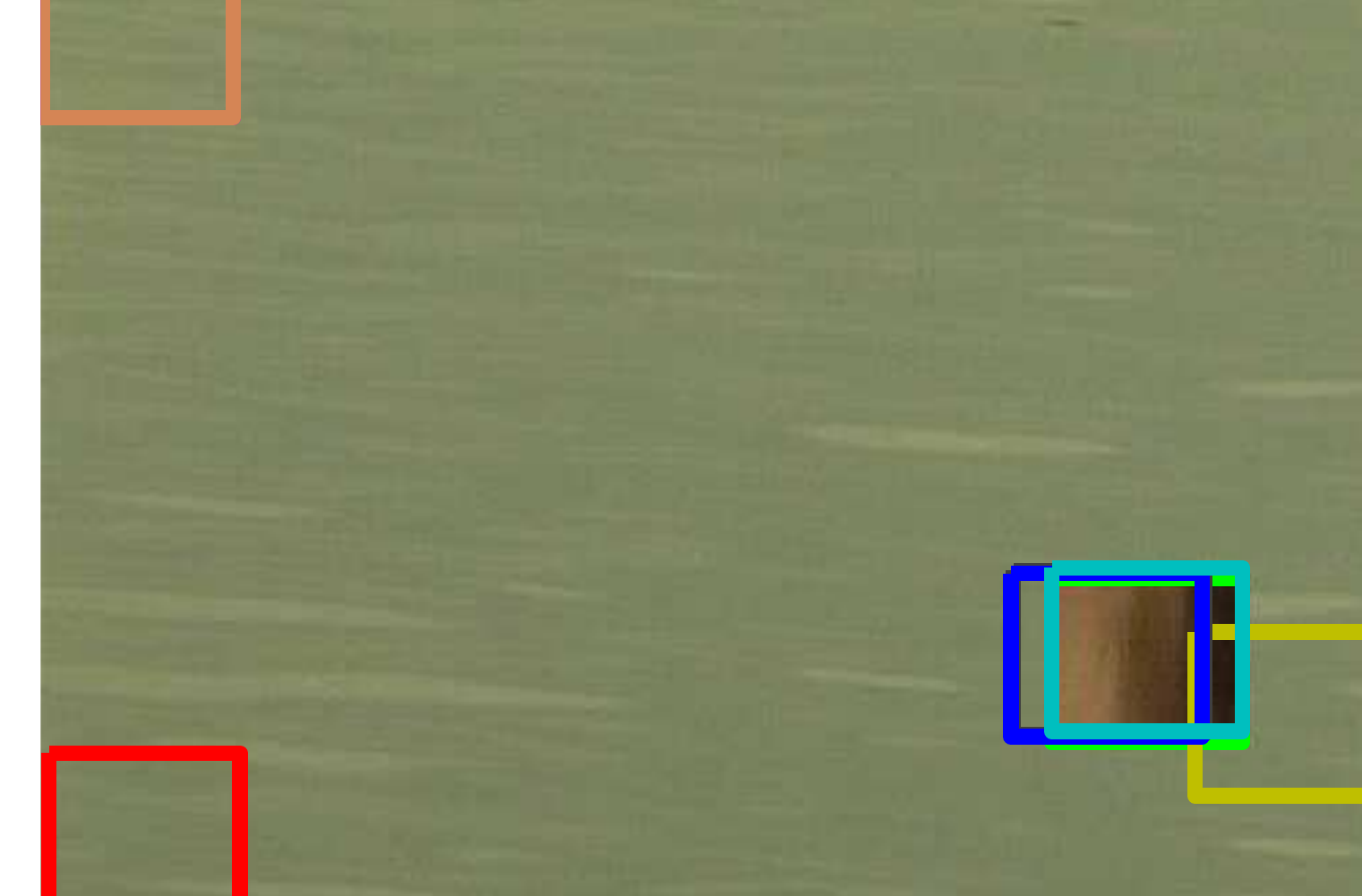}
\label{fig:dyntex1-frame4}
}
\caption{Tracking steam on water waves. [Boost (magenta), TM (yellow), MS (black), MS-VR (blue) and MS-HR (red), DT-PF (light brown), DK-SSD-T (cyan). Ground-truth (green)]}
\label{fig:results-dyntex1}
\end{figure*}
\begin{figure*}[tb]
\centering
\subfigure{
\includegraphics*[width=0.23\linewidth]{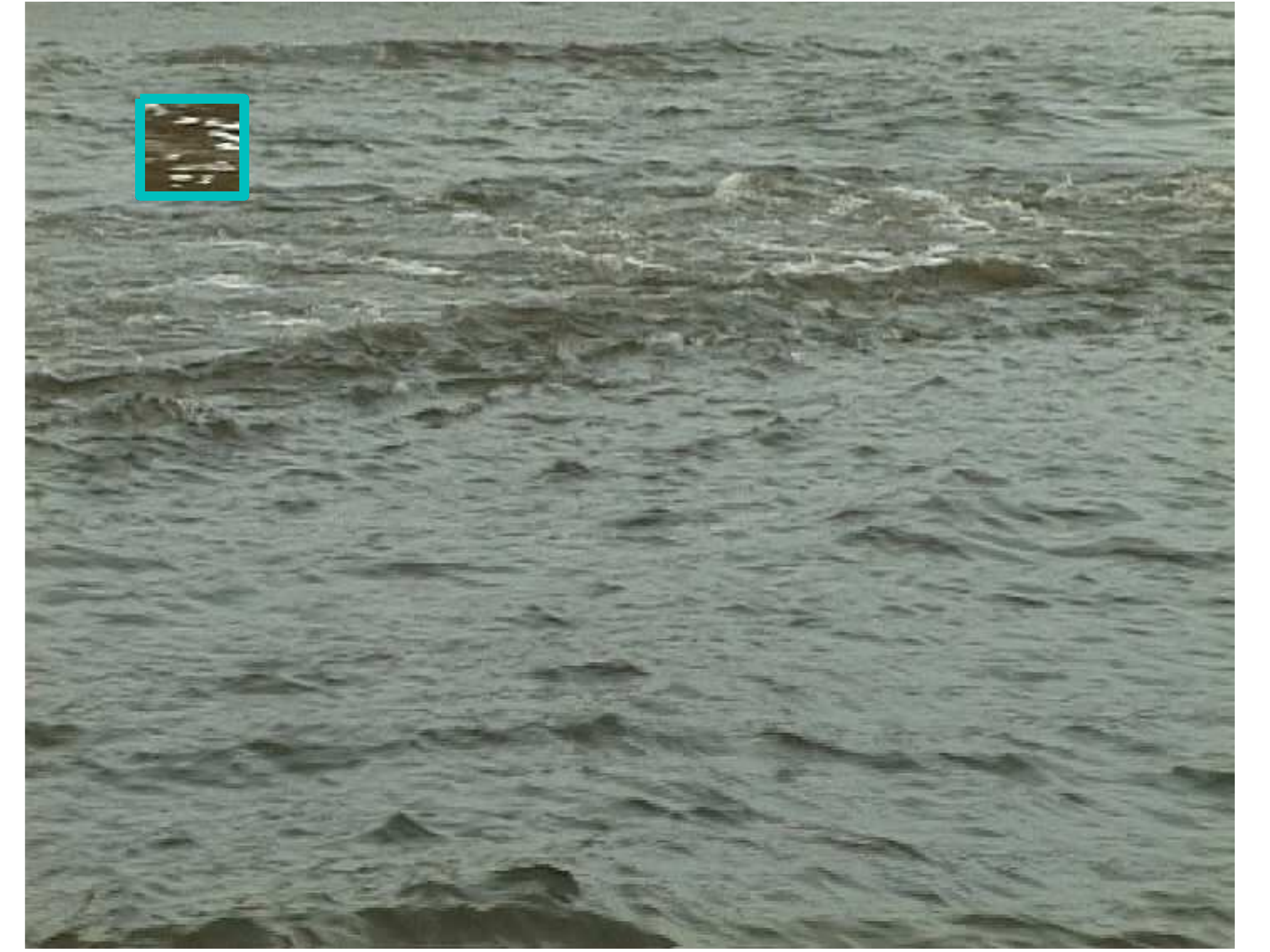}
\label{fig:dyntex2-frame1}
}
\subfigure{
\includegraphics*[width=0.23\linewidth]{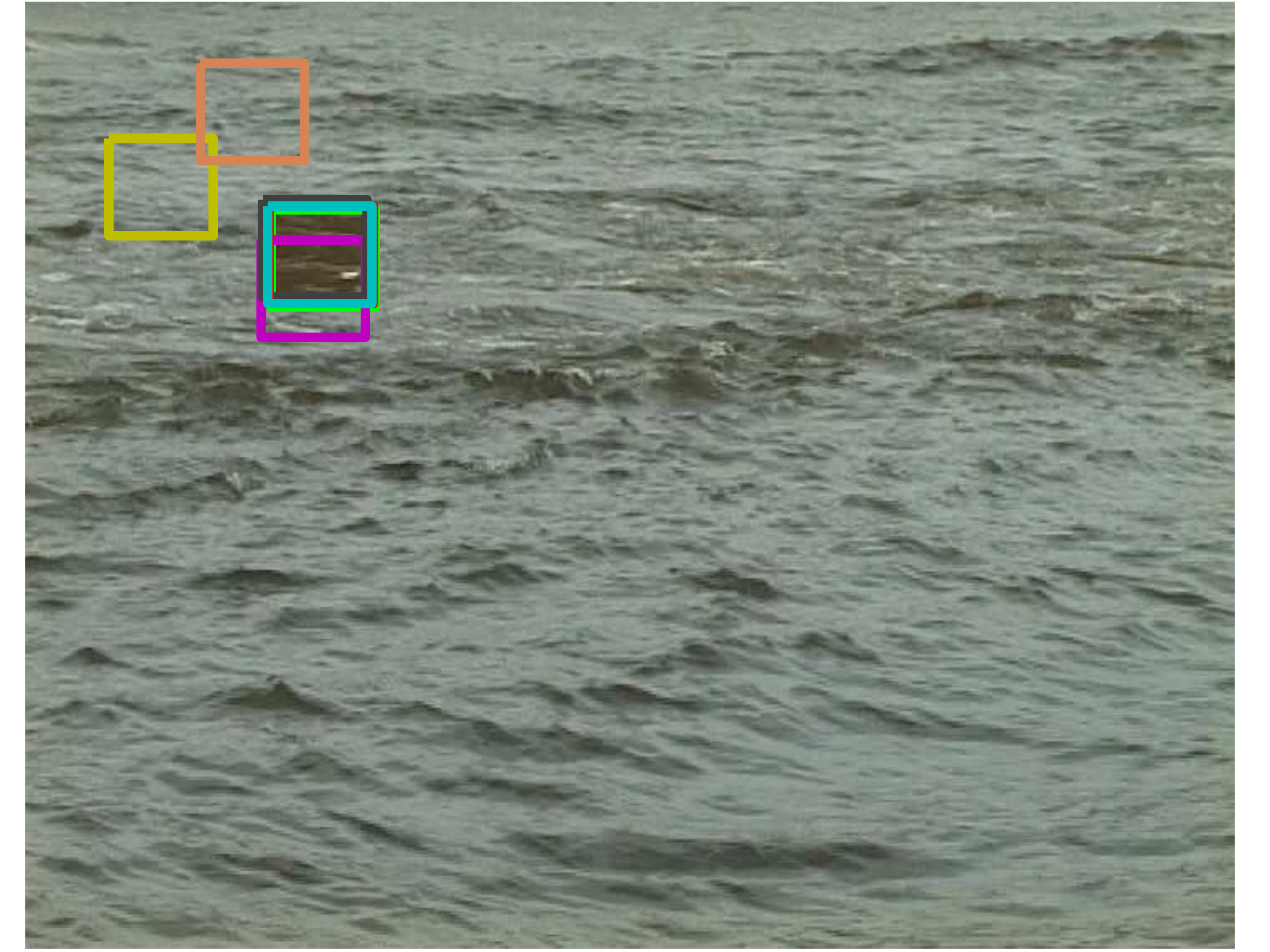}
\label{fig:dyntex2-frame2}
}
\subfigure{
\includegraphics*[width=0.23\linewidth]{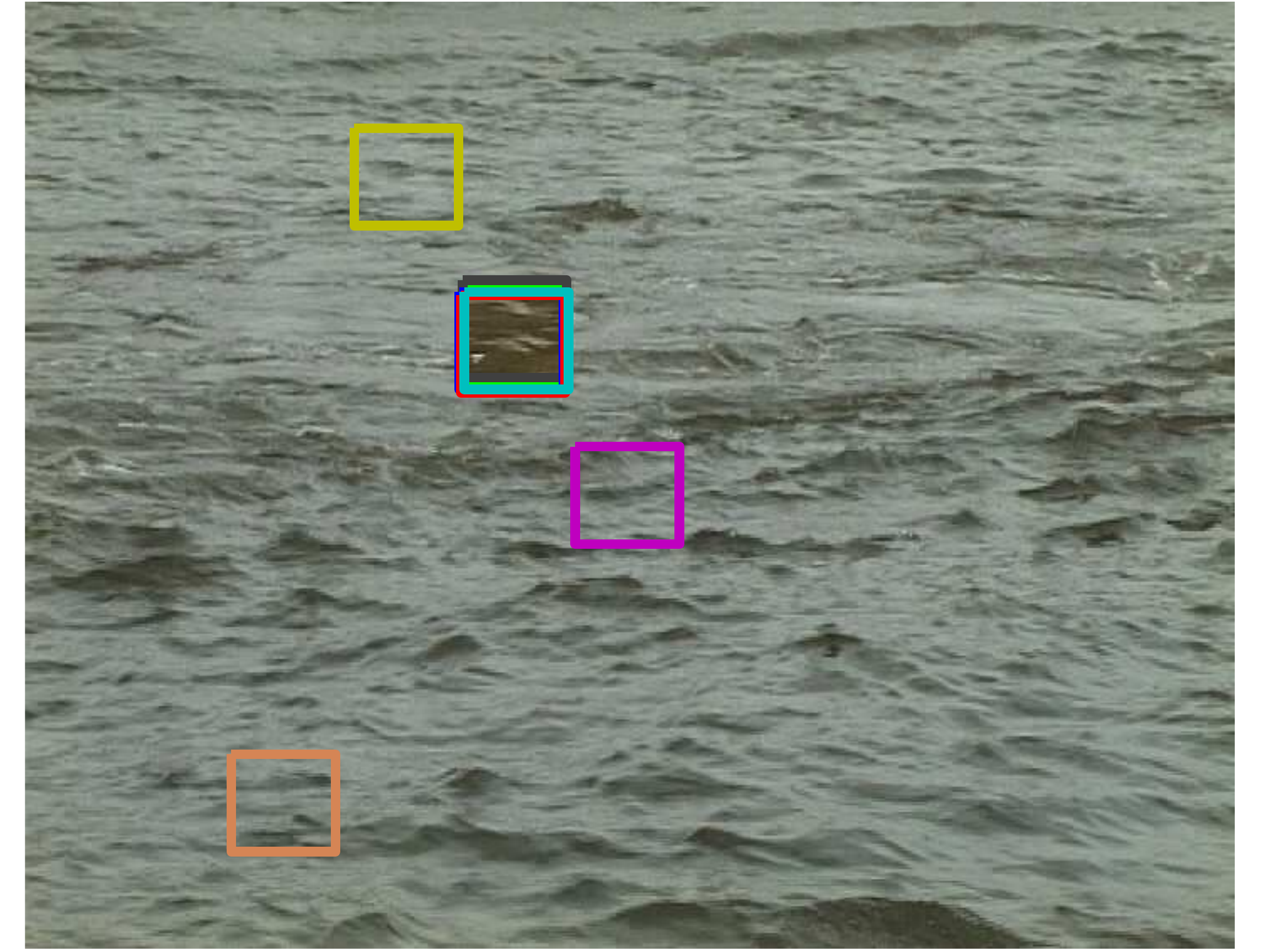}
\label{fig:dyntex2-frame3}
}
\subfigure{
\includegraphics*[width=0.23\linewidth]{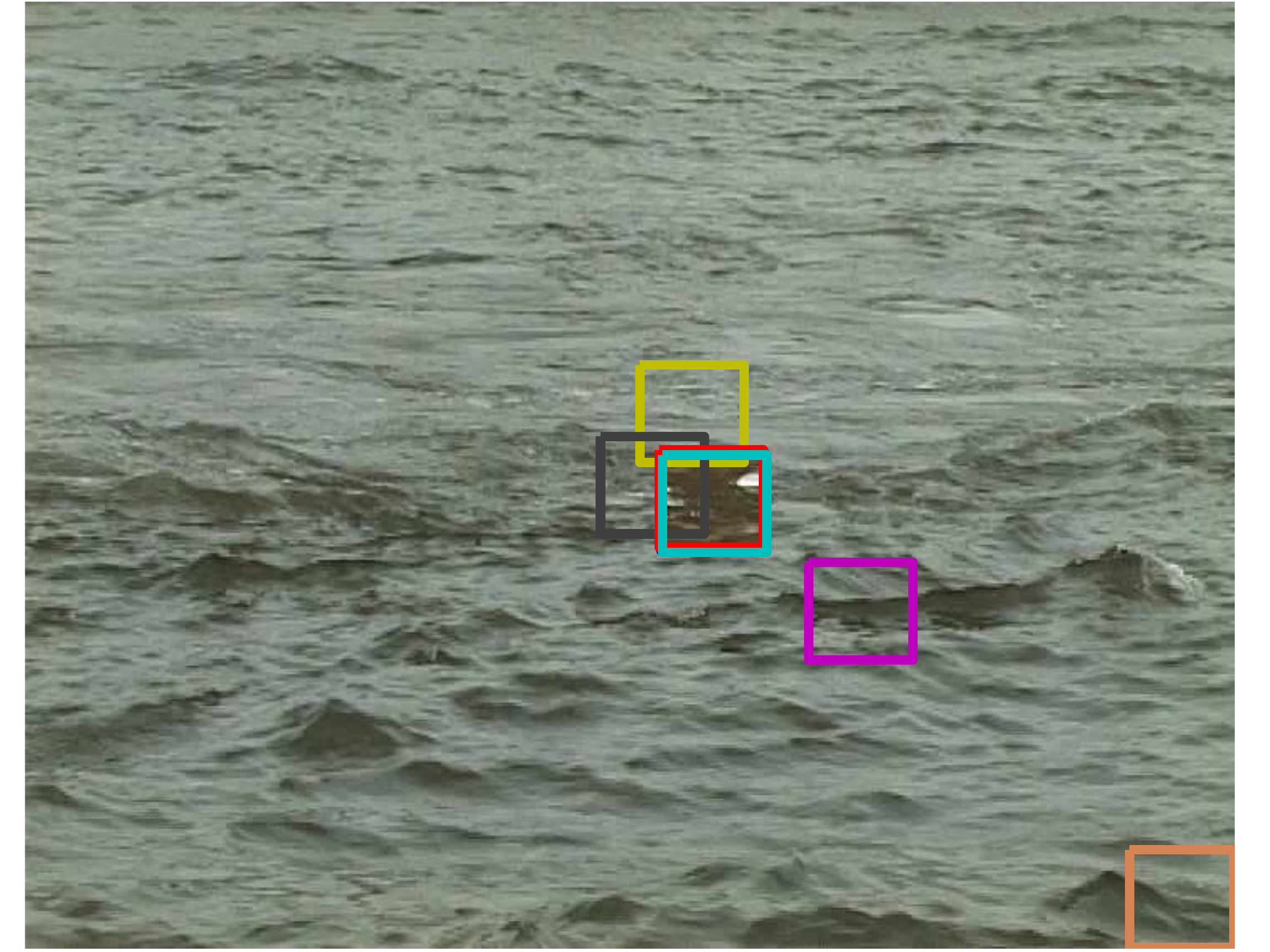}
\label{fig:dyntex2-frame4}
}
\caption{Tracking water with different appearance dynamics on water waves. [Boost (magenta), TM (yellow), MS (black), MS-VR (blue) and MS-HR (red), DT-PF (light brown), DK-SSD-T (cyan). Ground-truth (green)]}
\label{fig:results-dyntex2}
\end{figure*}

\begin{figure*}
\centering
\subfigure{
\includegraphics[width=0.45\linewidth]{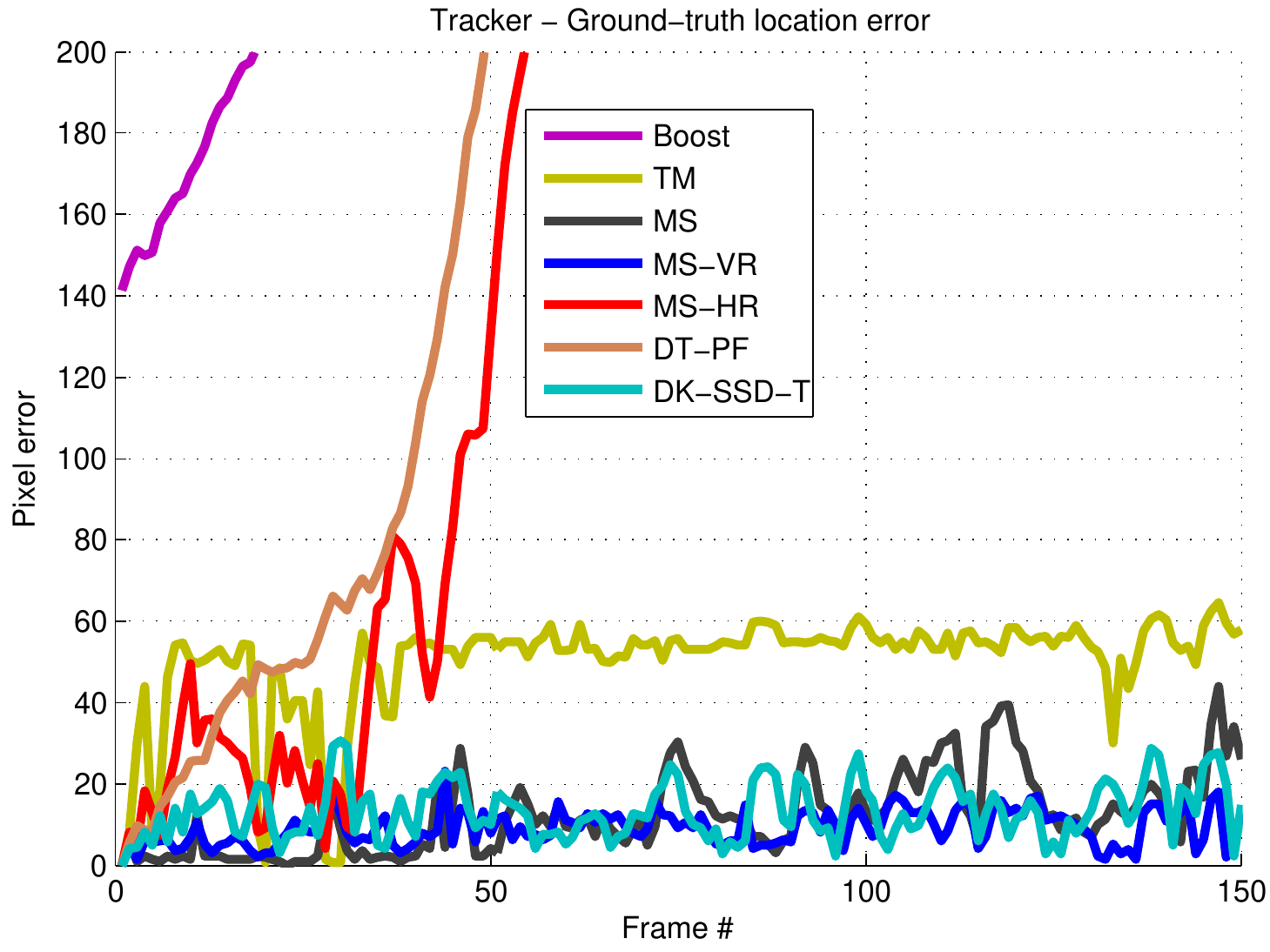}
\label{fig:locationError-dyntex1}
}
\subfigure{
\includegraphics[width=0.45\linewidth]{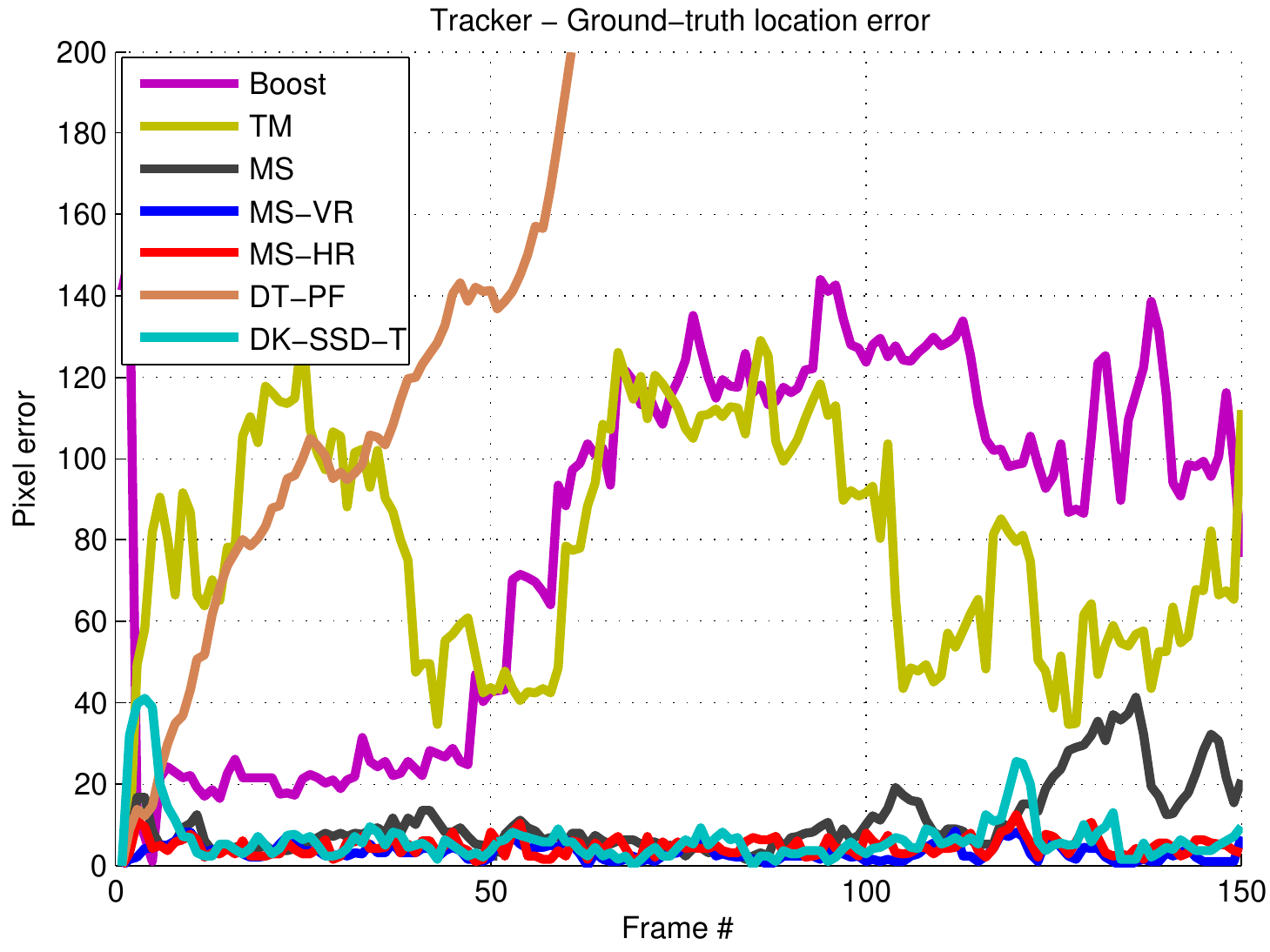}
\label{fig:locationError-dyntex2}
}
\caption{Pixel location error between tracker and ground truth location for videos in Fig. \ref{fig:results-dyntex1} (left) and Fig. \ref{fig:results-dyntex2} (right).}
\label{fig:locationErrorSynth}
\end{figure*}

\subsection{Tracking Synthetic Dynamic Textures}

To compare our algorithm against the state-of-the-art on dynamic data with ground-truth, we first create synthetic dynamic texture sequences by manually placing one dynamic texture patch on another dynamic texture. We use sequences from the DynTex database \citep{Peteri:DynTex} for this purpose. The dynamics of the foreground patch are learnt offline using the method for identifying the parameters, $(\mu,A,C,B,R)$, in \citet{Doretto:IJCV03}. These are then used in our tracking framework.

In Fig. \ref{fig:results-dyntex1}, the dynamic texture is a video of steam rendered over a video of water. We see that Boost, DT-PF, and MS-HR eventually lose track of the dynamic patch. The other methods stay close to the patch however, our proposed method stays closest to the ground truth till the very end. In Fig. \ref{fig:results-dyntex2}, the dynamic texture is a sequence of water rendered over a different sequence of water with different dynamics. Here again, Boost and TM lose tracks. DT-PF stays close to the patch initially but then diverges significantly. The other trackers manage to stay close to the dynamic patch, whereas our proposed tracker (cyan) still performs at par with the best. Fig. \ref{fig:locationErrorSynth} also shows the pixel location error at each frame for all the trackers and Table \ref{tab:meanLocationErrorSynth} provides the mean error and standard deviation for the whole video. Overall, MS-VR and our method seem to be the best, although MS-VR has a lower standard deviation in both cases. However note that our method gets similar performance without the use of background information, whereas MS-VR and MS-HR use background information to build more discriminative features. Due to the dynamic changes in background appearance, Boost fails to track in both cases. Even though DT-PF is designed for dynamic textures, upon inspection, all the particles generated turn out to have the same (low) probability. Therefore, the tracker diverges. Given the fact that our method is only based on the appearance statistics of the foreground patch, as opposed to the adaptively changing foreground/background model of MS-VR, we attribute the comparable performance (especially against other foreground-only methods) to the explicit inclusion of foreground dynamics.

\begin{table}
\centering
\begin{tabular}{|l||c|c|}
\hline
Algorithm & \figref{fig:results-dyntex1} & \figref{fig:results-dyntex2} \\
\hline
\hline
Boost & 389 $\pm$ 149 & 82 $\pm$ $\pm$ 44 \\
\hline
TM & 50 $\pm$ 13 & 78 $\pm$ 28 \\
\hline
MS & 12 $\pm$ 10 & 10 $\pm$ 8.6 \\
\hline
MS-VR & 9.2 $\pm$ 4.3 & 3.2 $\pm$ 2.0 \\
\hline
MS-HR & 258 $\pm$ 174 & 4.6 $\pm$ 2.3\\
\hline
DT-PF & 550 $\pm$ 474  & 635 $\pm$ 652\\
\hline
DK-SSD-T & 8.6 $\pm$ 6.8 & 6.5 $\pm$ 6.6 \\
\hline
\end{tabular}
\caption{Mean pixel error with standard deviation between tracked location and ground truth.}
\label{tab:meanLocationErrorSynth}
\end{table}

\begin{figure*}
\centering
\subfigure[Training video with labeled stationary patch for learning \emph{candle flame} dynamic texture system parameters.]{
\includegraphics*[width=0.25\linewidth, viewport=0 0 192 132]{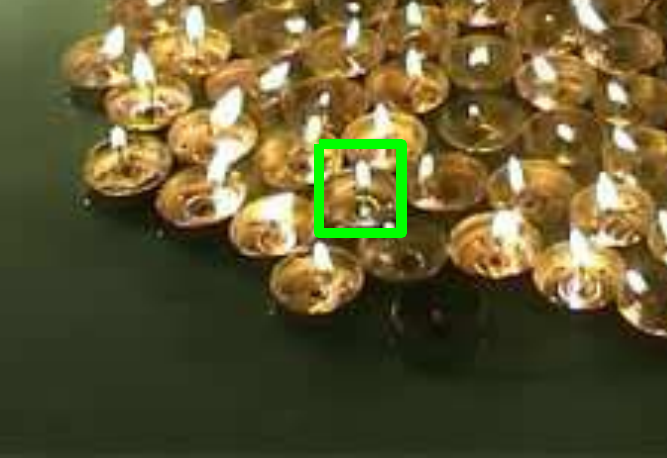}
\includegraphics*[width=0.25\linewidth, viewport=0 0 192 132]{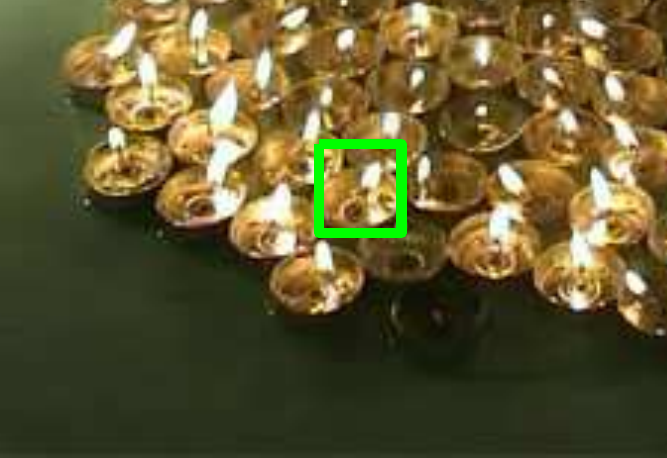}
\includegraphics*[width=0.25\linewidth, viewport=0 0 192 132]{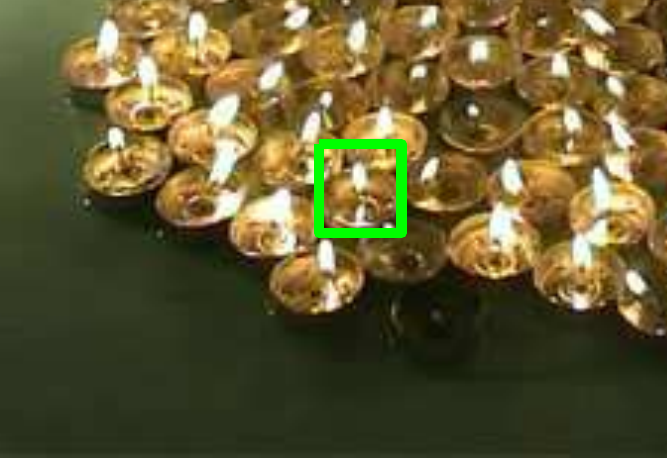}
\includegraphics*[width=0.25\linewidth, viewport=0 0 192 132]{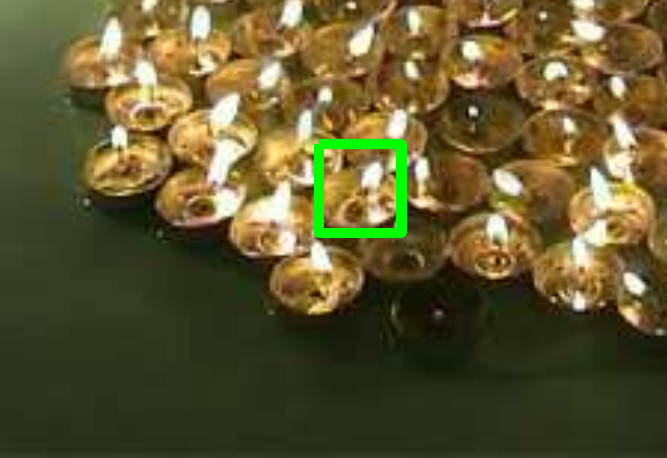}
}
\subfigure[Test video with tracked locations of \emph{candle flame}.]{
\includegraphics*[width=0.25\linewidth, viewport=0 0 192 132]{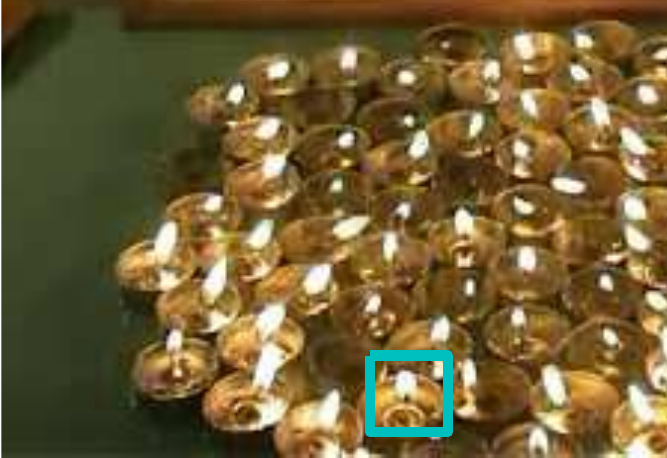}
\includegraphics*[width=0.25\linewidth, viewport=0 0 192 132]{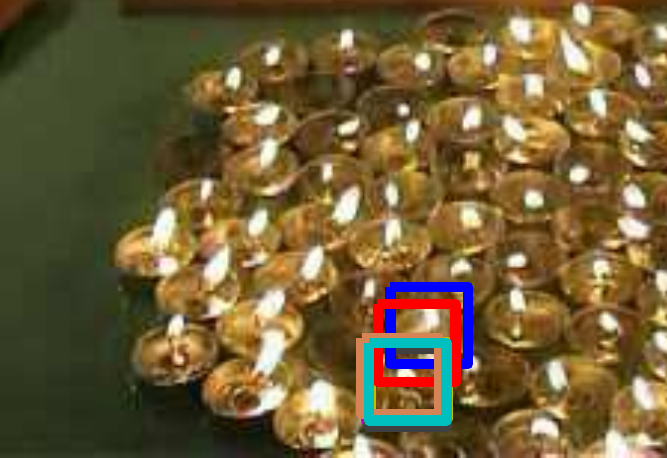}
\includegraphics*[width=0.25\linewidth, viewport=0 0 192 132]{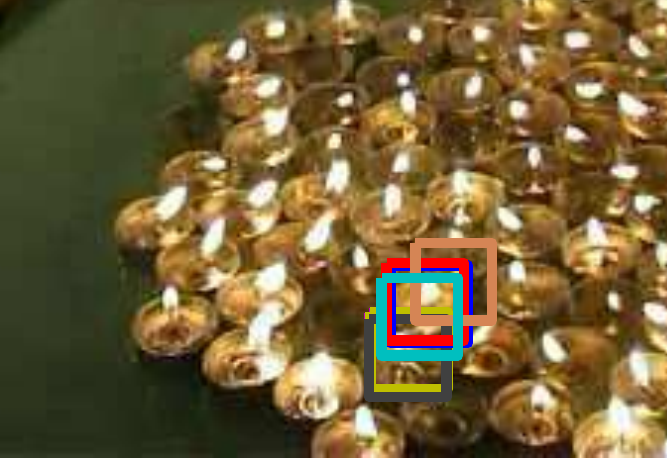}
\includegraphics*[width=0.25\linewidth, viewport=0 0 192 132]{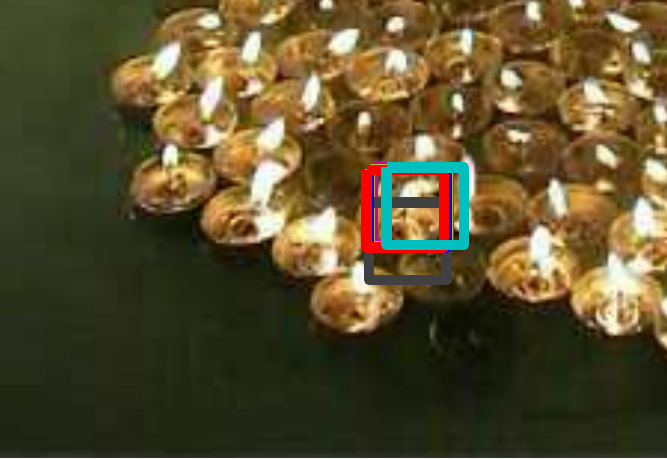}
}
%
\subfigure[Training video with labeled stationary patch for learning \emph{flag} dynamic texture system parameters.]{
\includegraphics[width=0.25\linewidth]{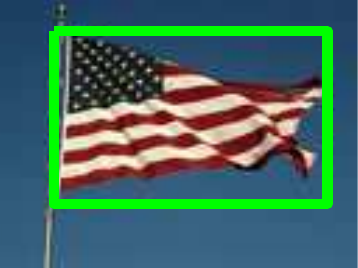}
\includegraphics[width=0.25\linewidth]{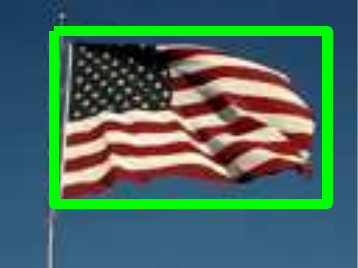}
\includegraphics[width=0.25\linewidth]{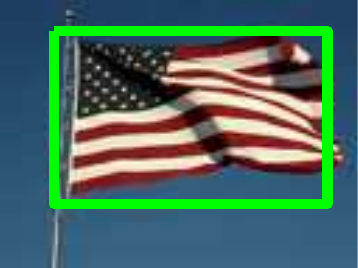}
\includegraphics[width=0.25\linewidth]{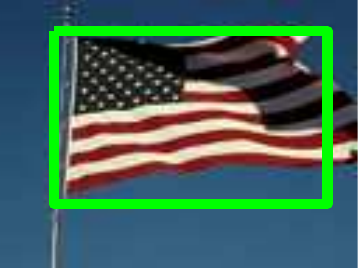}
}
\subfigure[Test video with tracked locations of \emph{flag}.]{
\includegraphics*[width=0.25\linewidth, viewport=60 0 276 144]{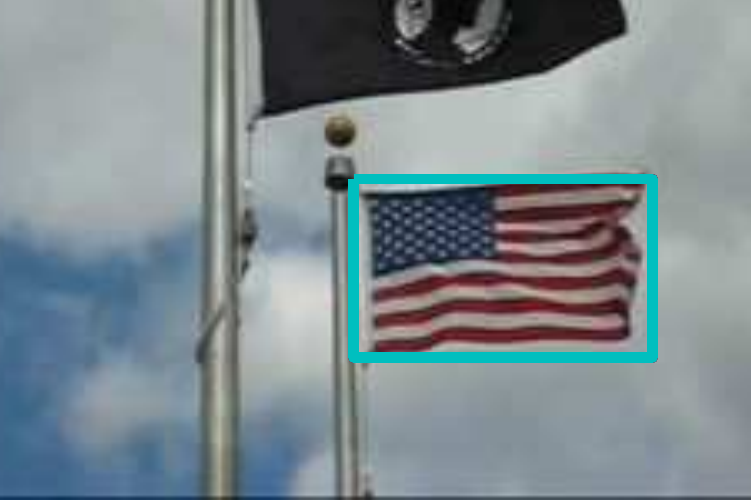}
\includegraphics*[width=0.25\linewidth, viewport=60 0 276 144]{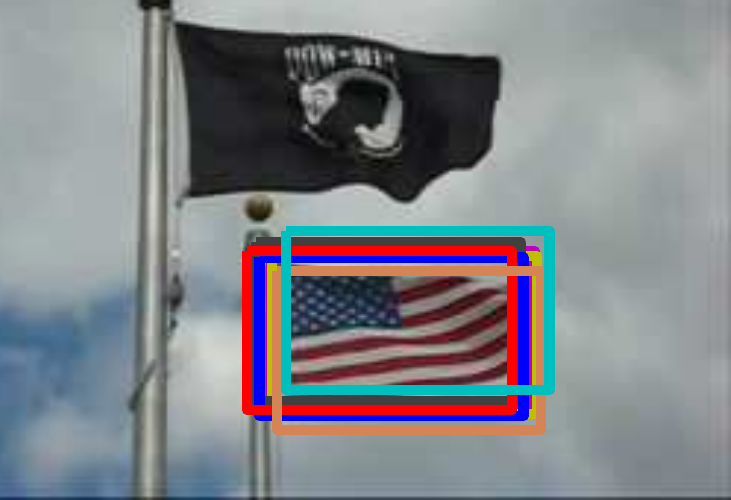}
\includegraphics*[width=0.25\linewidth, viewport=60 0 276 144]{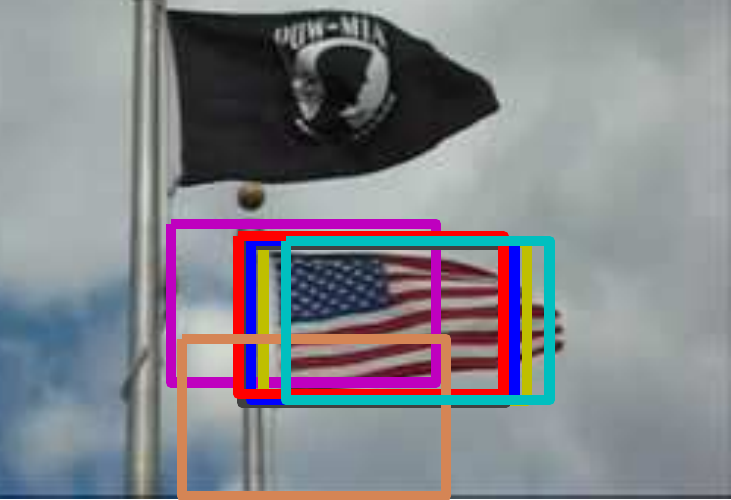}
\includegraphics*[width=0.25\linewidth, viewport=60 0 276 144]{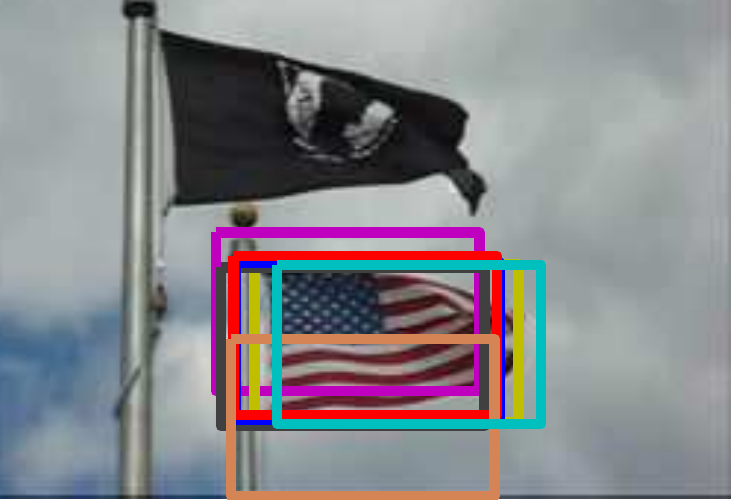}
}
%
\subfigure[Training video with labeled stationary patch for learning \emph{fire} dynamic texture system parameters.]{
\includegraphics*[width=0.25\linewidth, viewport=108 72 684 534]{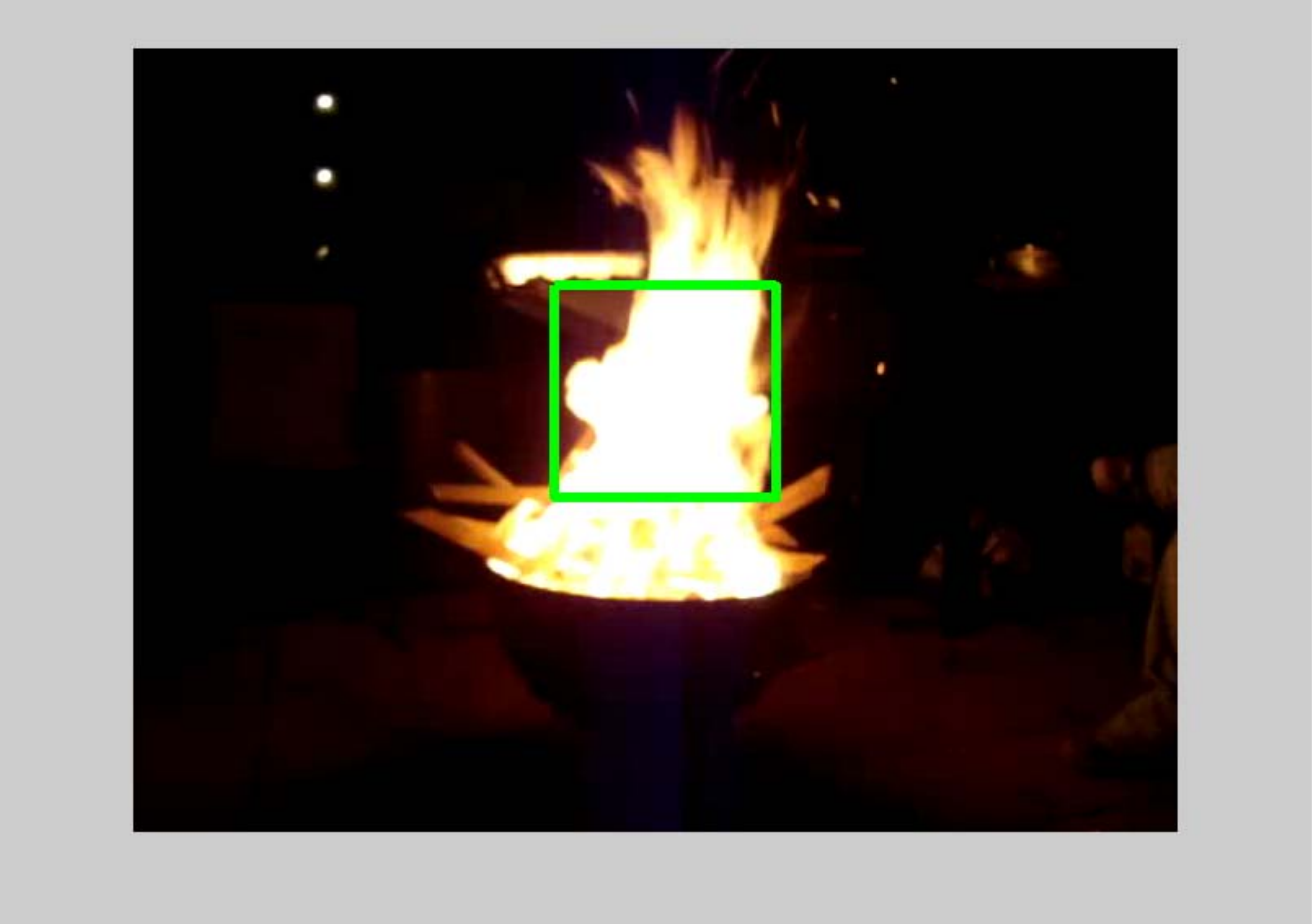}
\includegraphics*[width=0.25\linewidth, viewport=108 72 684 534]{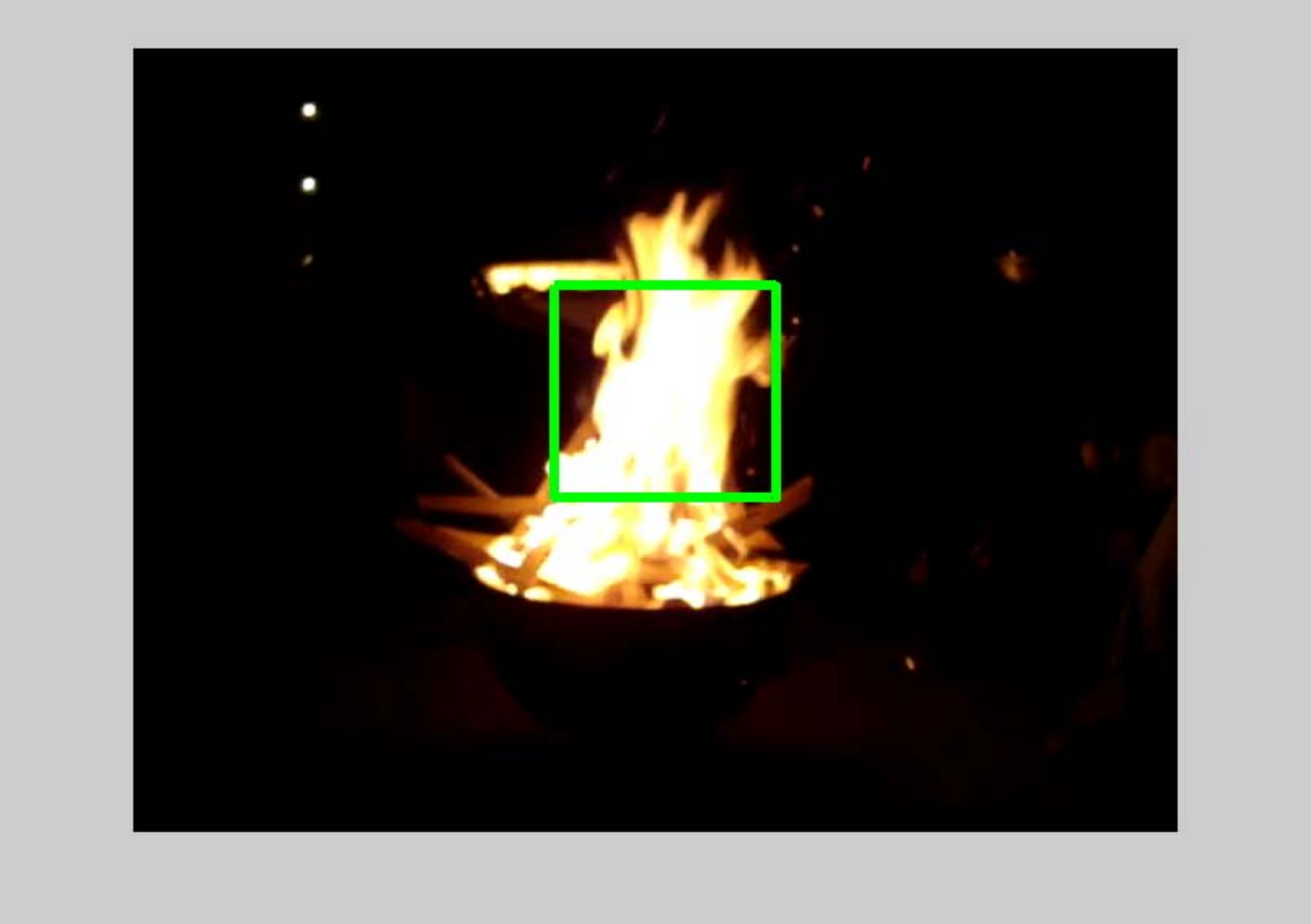}
\includegraphics*[width=0.25\linewidth, viewport=108 72 684 534]{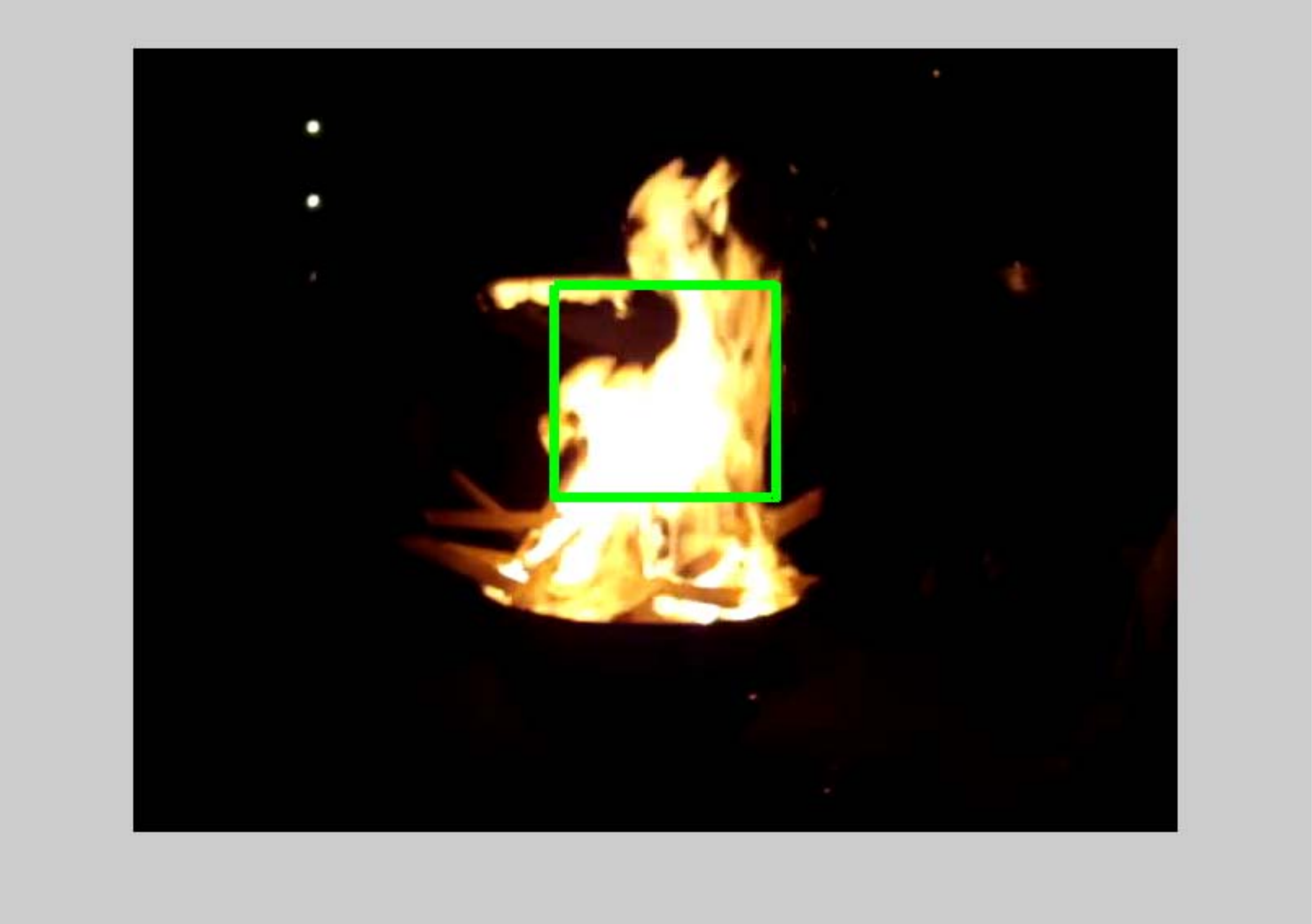}
\includegraphics*[width=0.25\linewidth, viewport=108 72 684 534]{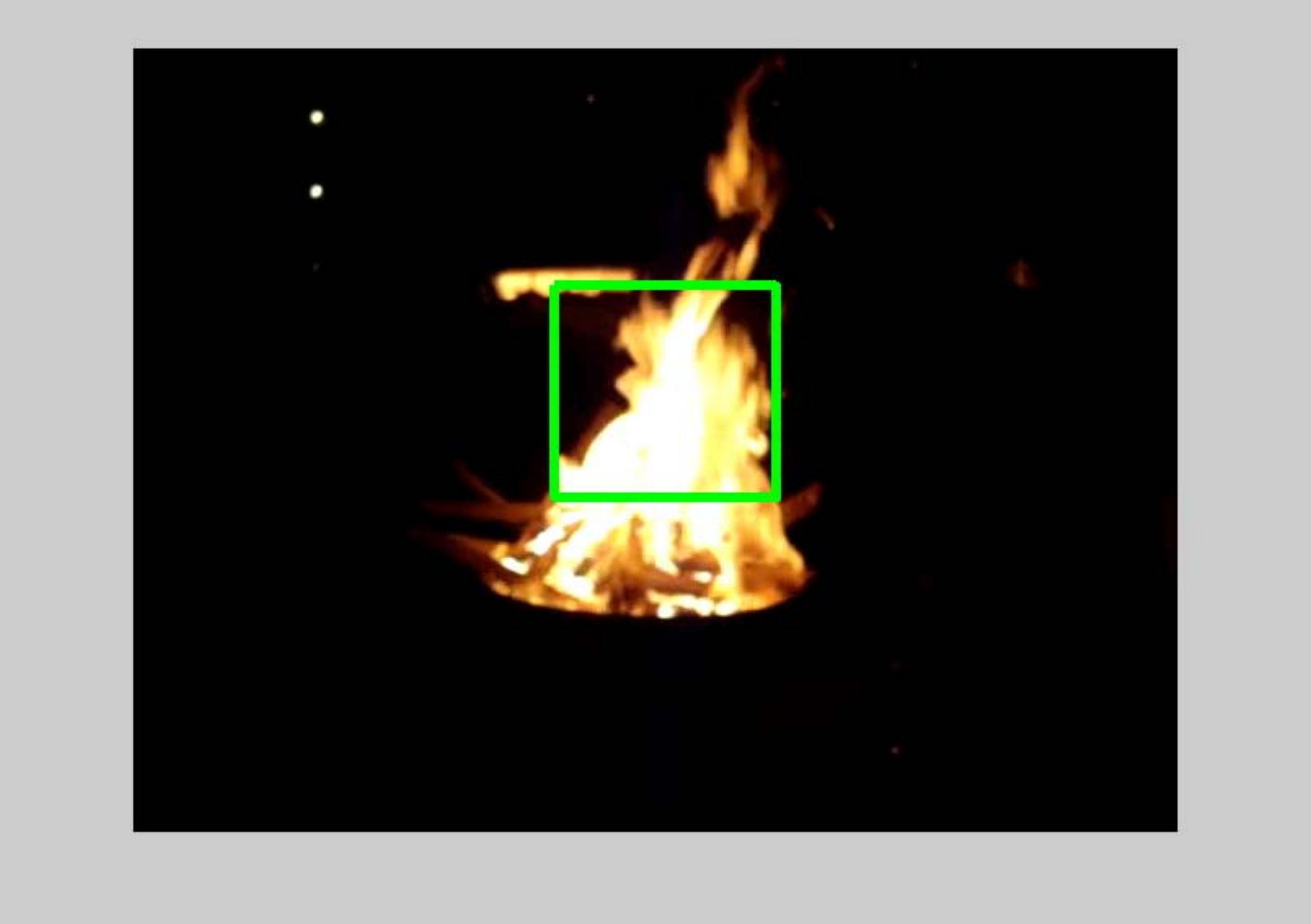}
}
\subfigure[Test video with tracked locations of \emph{fire}.]{
\includegraphics*[width=0.25\linewidth, viewport=200 120 400 288]{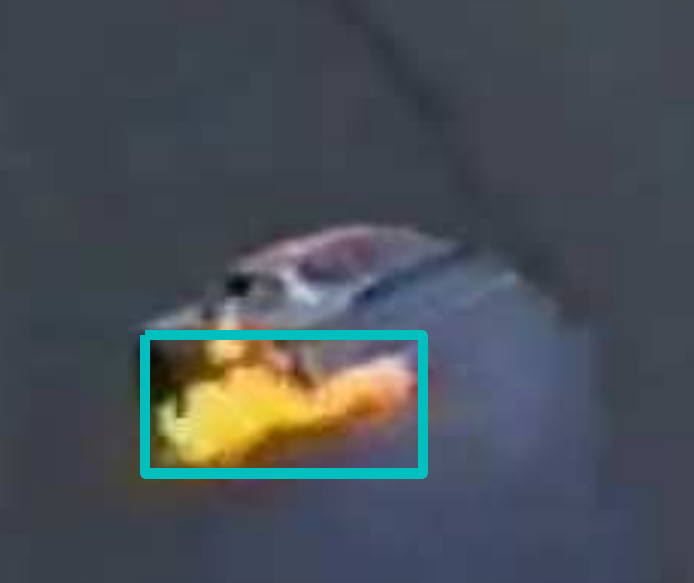}
\label{fig:nascar-2-1}
\includegraphics*[width=0.25\linewidth, viewport=200 120 400 288]{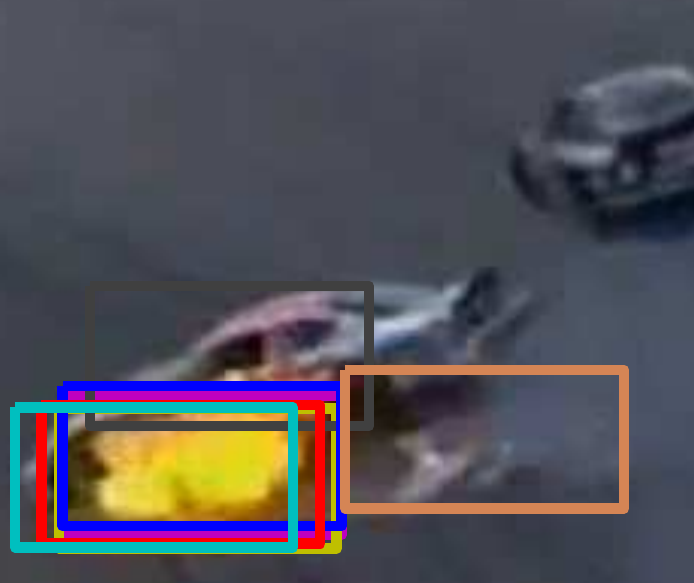}
\label{fig:nascar-2-2}
\includegraphics*[width=0.25\linewidth, viewport=200 120 400 288]{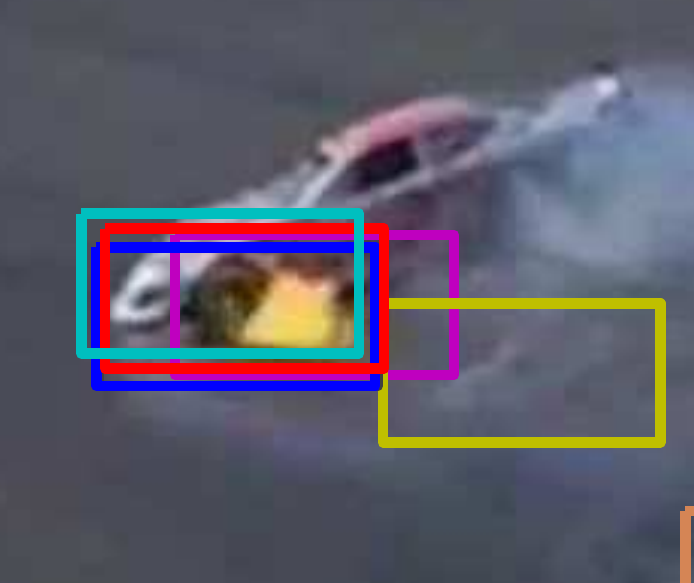}
\label{fig:nascar-2-3}
\includegraphics*[width=0.25\linewidth, viewport=200 120 400 288]{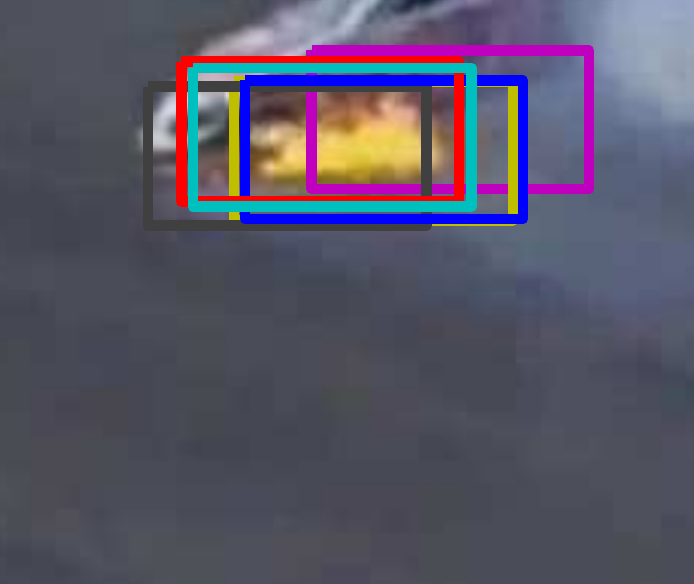}
\label{fig:nascar-2-4}
}
\caption{Training and Testing results for dynamic texture tracking. Boost (magenta), TM (yellow), MS (black), MS-VR (blue) and MS-HR (red), DT-PF (light brown), DK-SSD-T (cyan). Training (green).}
\label{fig:dyntex-results}
\end{figure*}

\subsection{Tracking Real Dynamic Textures}

To test our algorithm on real videos with dynamic textures, we provide results on three different scenarios in real videos. We learn the system parameters of the dynamic texture by marking a bounding box around the texture in a video where the texture is stationary. We then use these parameters to track the dynamic texture in a different video with camera motion causing the texture to move around in the scene. All the trackers are initialized at the same location in the test video. Fig. \ref{fig:dyntex-results} provides the results for the three examples by alternatively showing the training video followed by the tracker results in the test video. We will use (row, column) notation to refer to specific images in Fig. \ref{fig:dyntex-results}.

\myparagraph{Candle Flame} This is a particularly challenging scene as there are multiple candles with similar dynamics and appearance and it is easy for a tracker to switch between candles. As we can see from (2,2), MS-HR and MS-VR seem to jump around between candles. Our method also jumps to another candle in (2,3) but recovers in (2,4), whereas MS is unable to recover. DT-PF quickly loses track and diverges. Overall, all trackers, except DT-PF seem to perform equally well.

\myparagraph{Flags} Even though the flag has a distinct appearance compared to the background, the movement of the flag fluttering in the air changes the appearance in a dynamic fashion. Since our tracker has learnt an explicit model of the dynamics of these appearance changes, it stays closest to the correct location while testing. Boost, DT-PF, and the other trackers deviate from the true location as can be seen in (4,3) and (4,4). 

\myparagraph{Fire} Here we show another practical application of our proposed method: tracking fire in videos. We learn a dynamical model for fire from a training video which is taken in a completely different domain, \eg a campfire. We then use these learnt parameters to track fire in a NASCAR video as shown in the last row of Fig. \ref{fig:dyntex-results}. Our foreground-only dynamic tracker performs better than MS and TM, the other foreground-only methods in comparison. Boost, MS-VR and MS-HR use background information, which is discriminative enough in this particular video and achieve similar results. DT-PF diverges from the true location and performs the worst.



\begin{figure*}[tb]
\centering
\subfigure{
\includegraphics[width=0.23\linewidth]{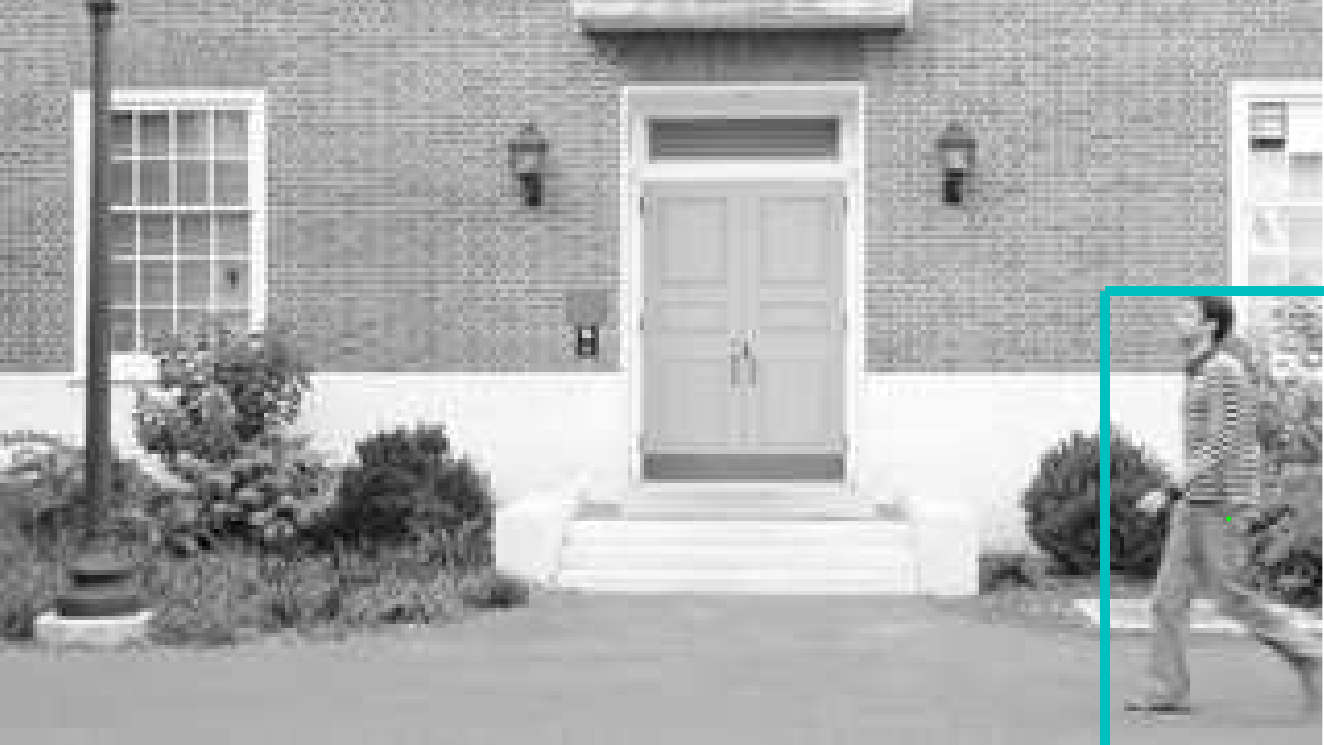}
\label{fig:walk-1}
}
\subfigure{
\includegraphics[width=0.23\linewidth]{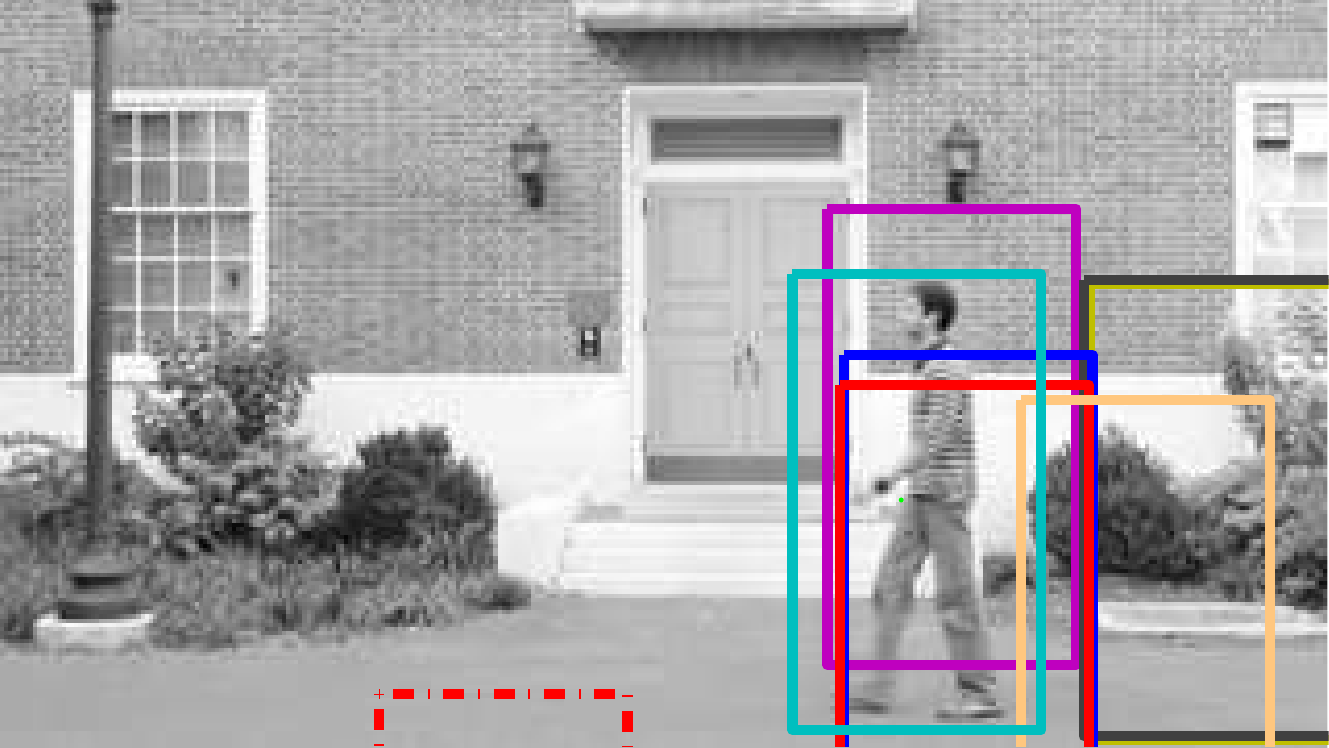}
\label{fig:walk-2}
}
\subfigure{
\includegraphics[width=0.23\linewidth]{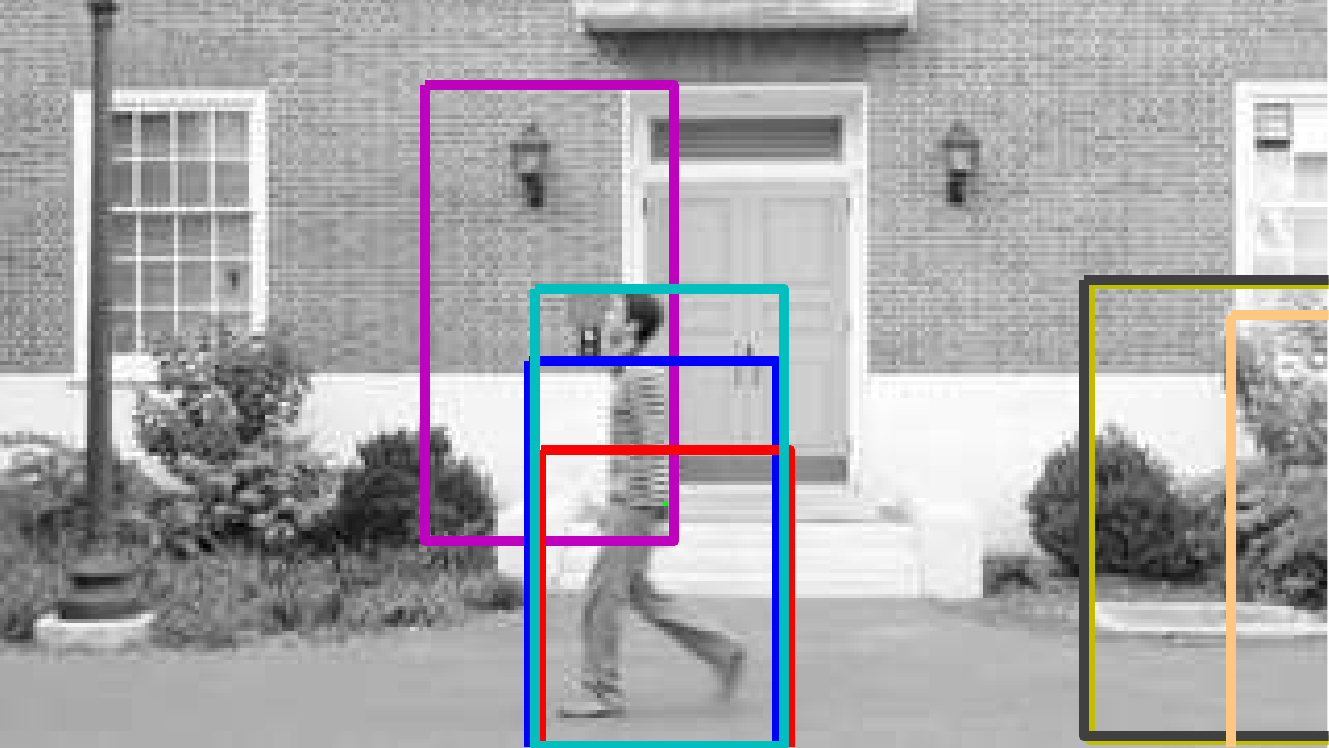}
\label{fig:walk-3}
}
\subfigure{
\includegraphics[width=0.23\linewidth]{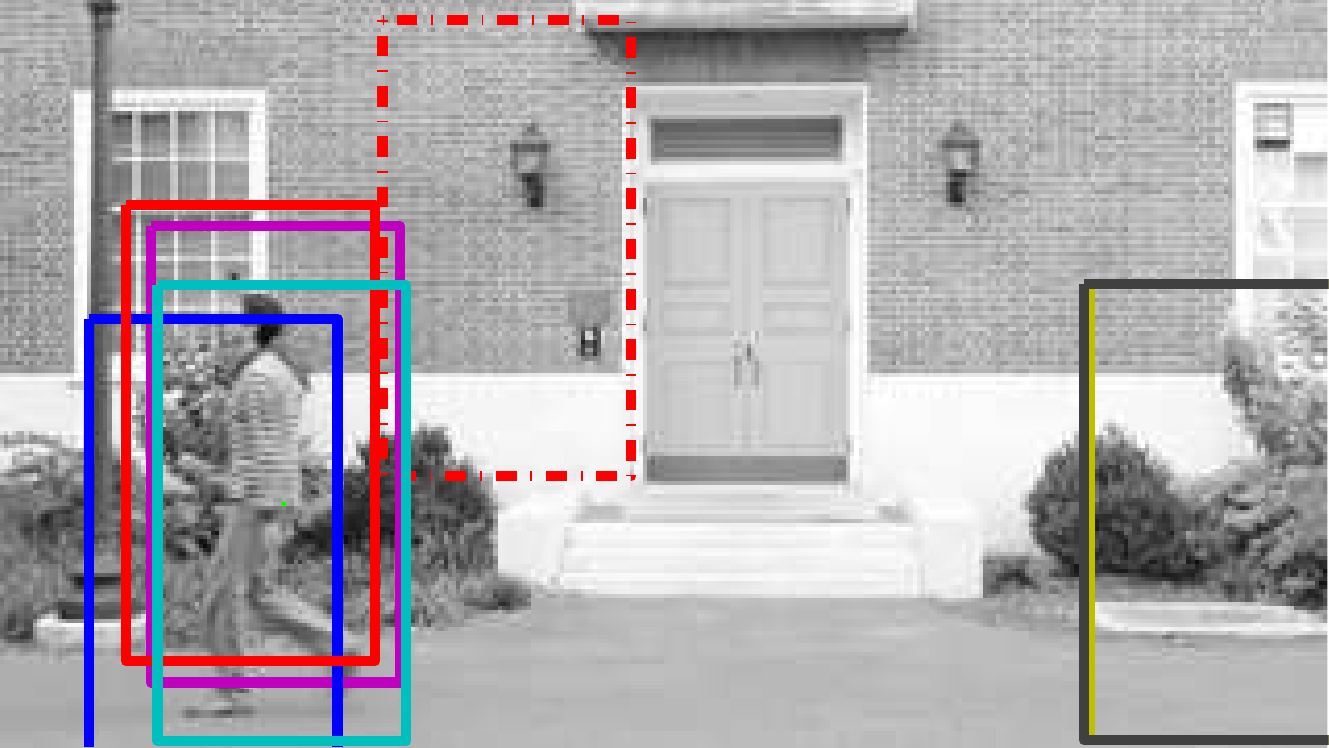}
\label{fig:walk-4}
}
\subfigure{
\includegraphics[width=0.23\linewidth]{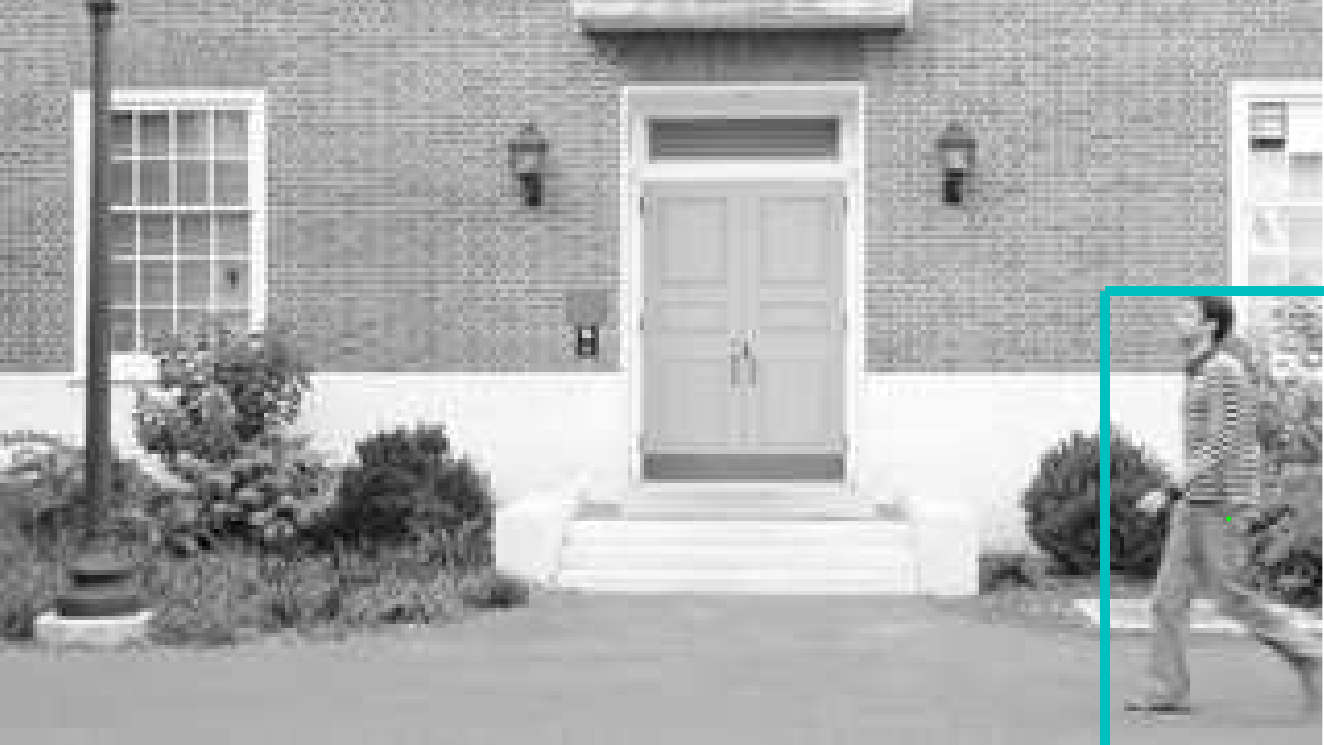}
\label{fig:walk-1-backSub}
}
\subfigure{
\includegraphics[width=0.23\linewidth]{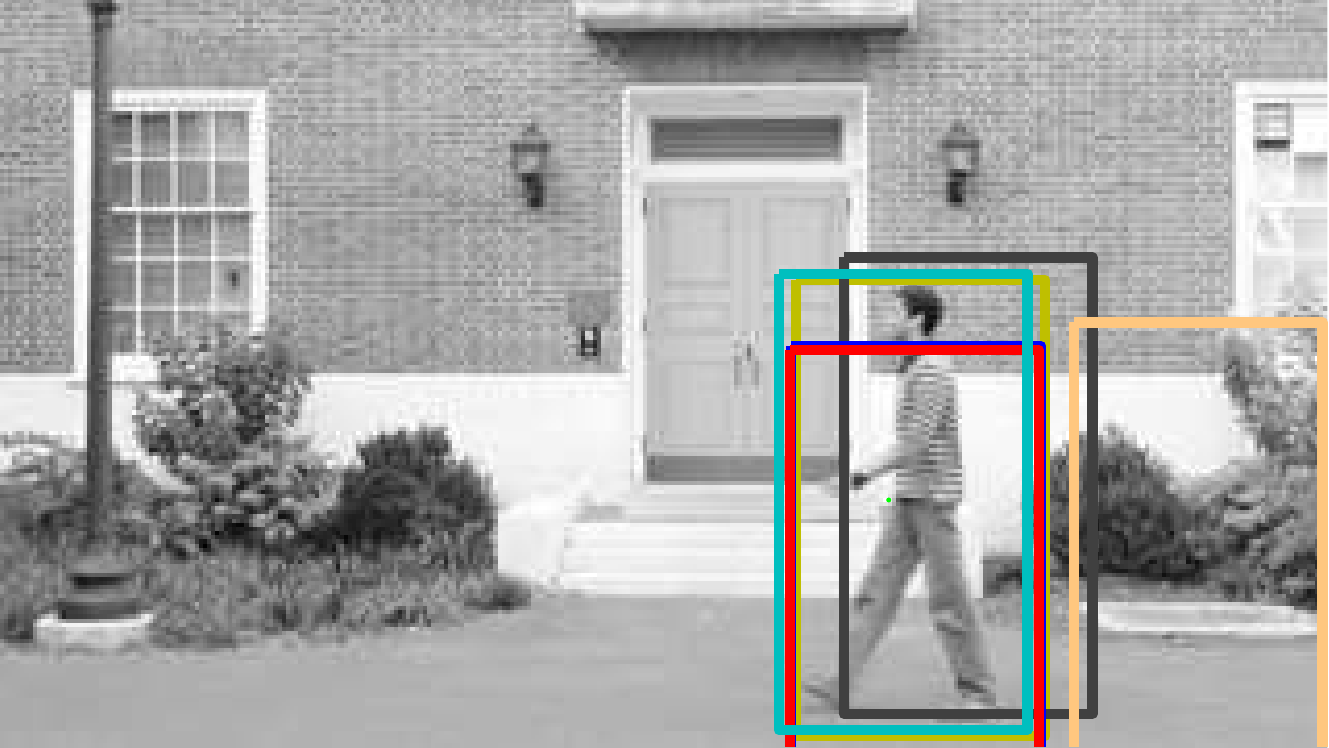}
\label{fig:walk-2-backSub}
}
\subfigure{
\includegraphics[width=0.23\linewidth]{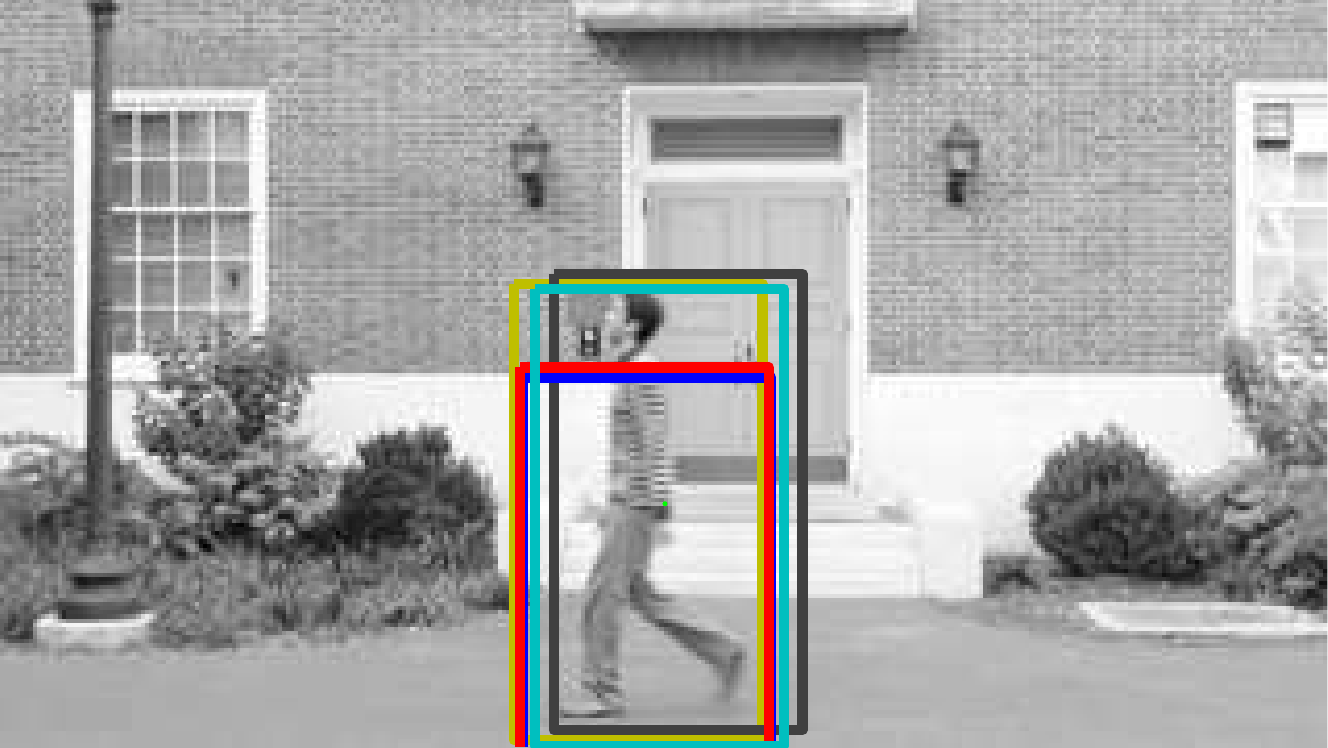}
\label{fig:walk-3-backSub}
}
\subfigure{
\includegraphics[width=0.23\linewidth]{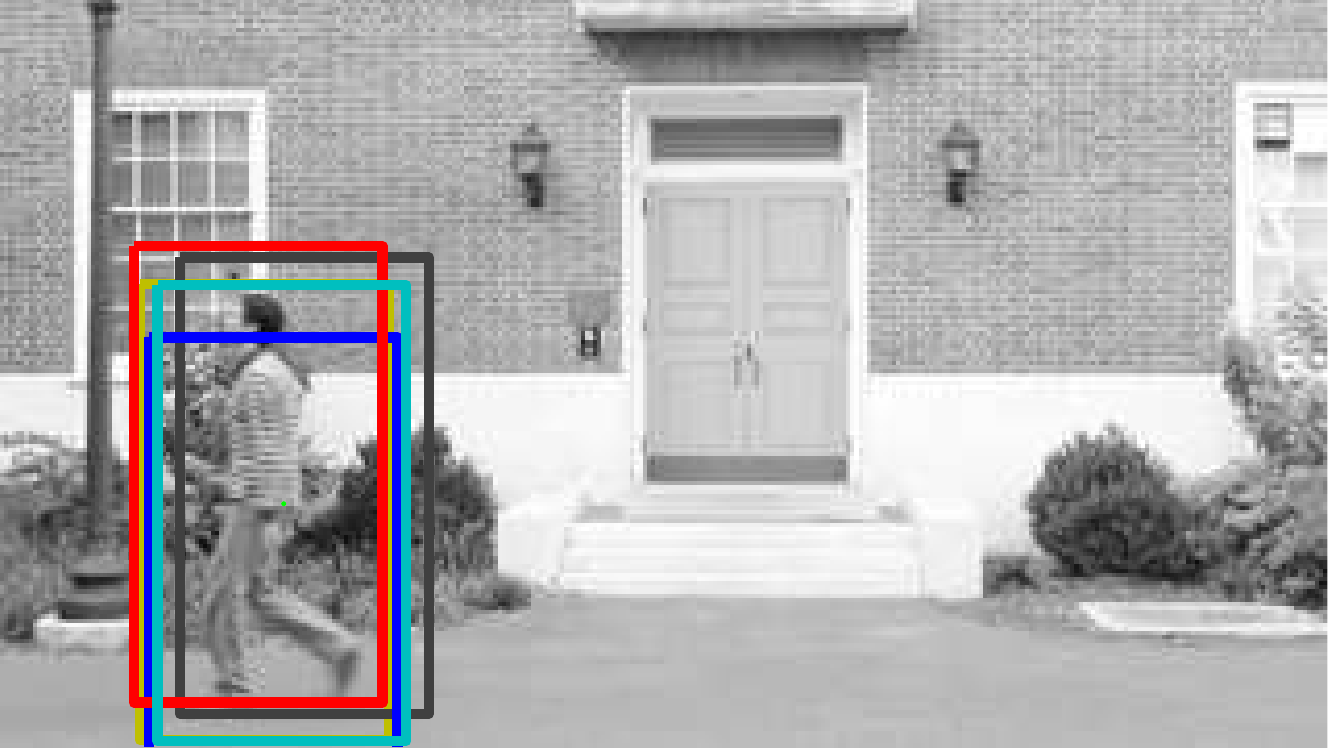}
\label{fig:walk-4-backSub}
}
\subfigure{
\includegraphics[width=0.23\linewidth]{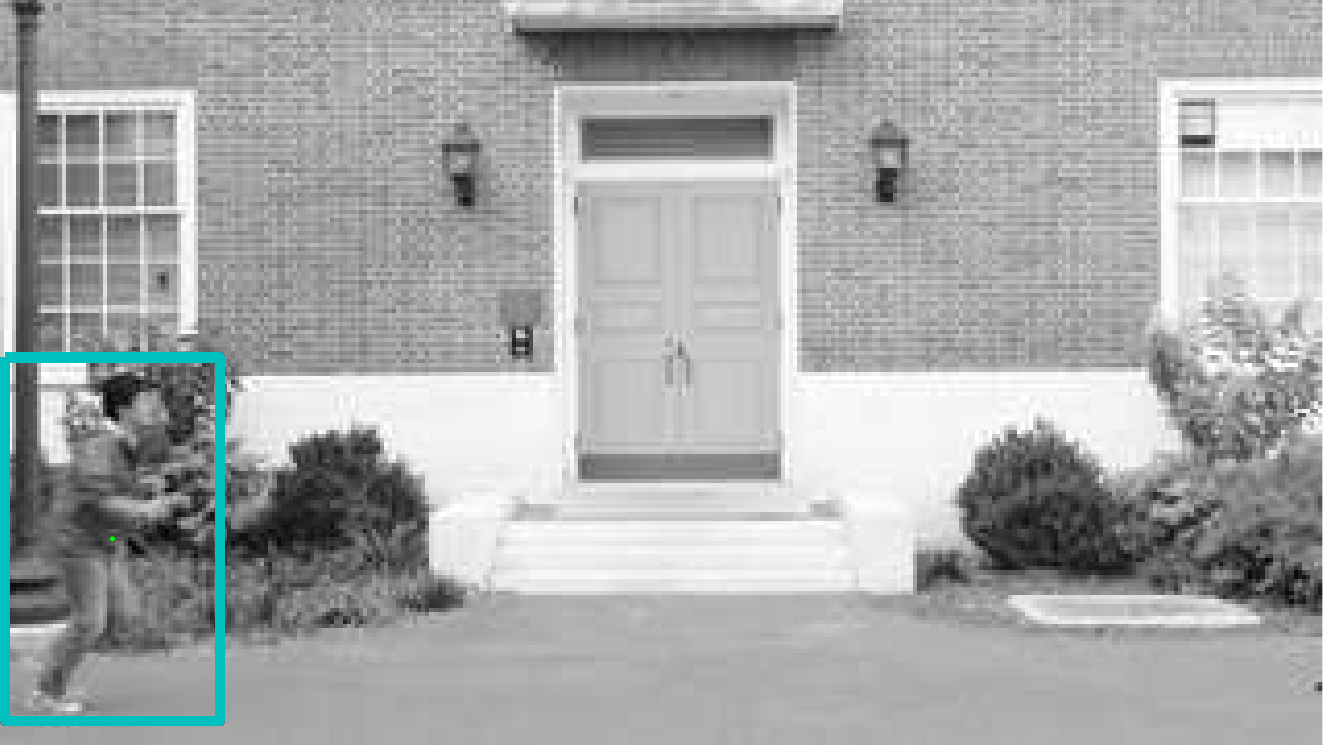}
\label{fig:run-1-backSub}
}
\subfigure{
\includegraphics[width=0.23\linewidth]{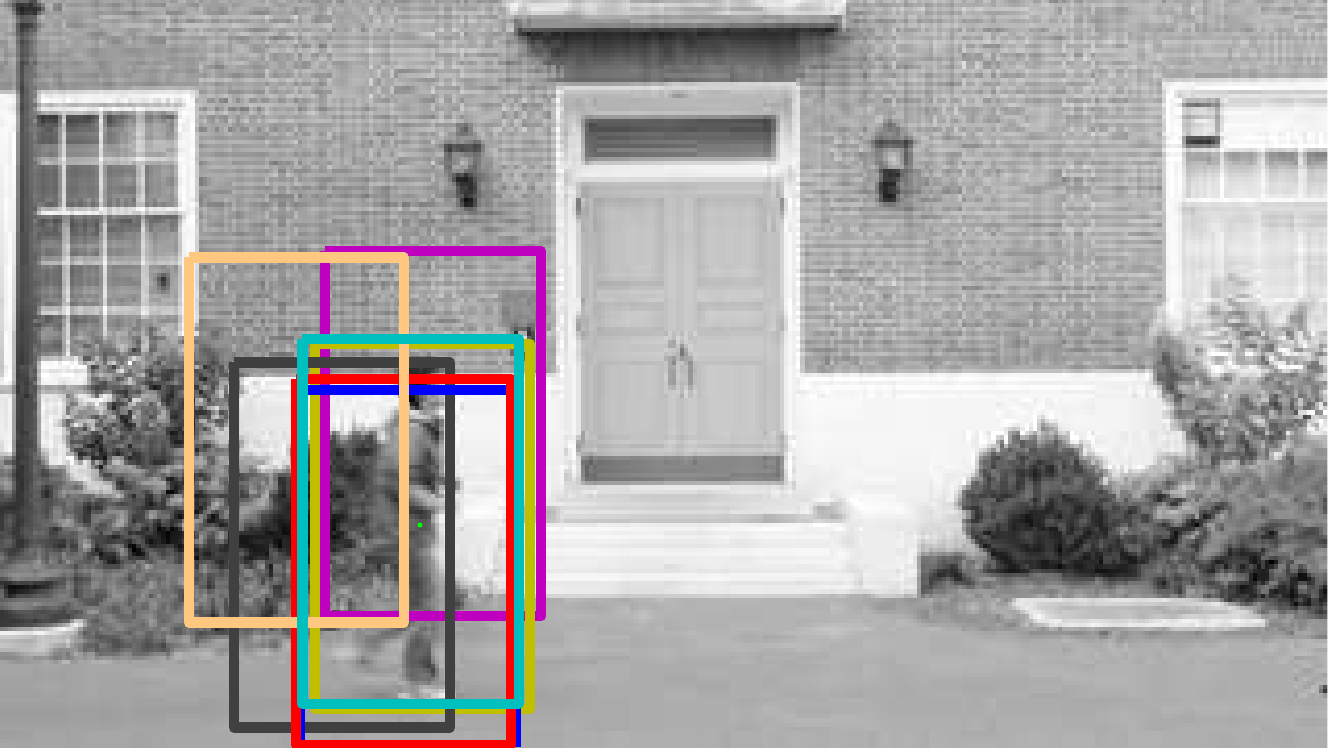}
\label{fig:run-2-backSub}
}
\subfigure{
\includegraphics[width=0.23\linewidth]{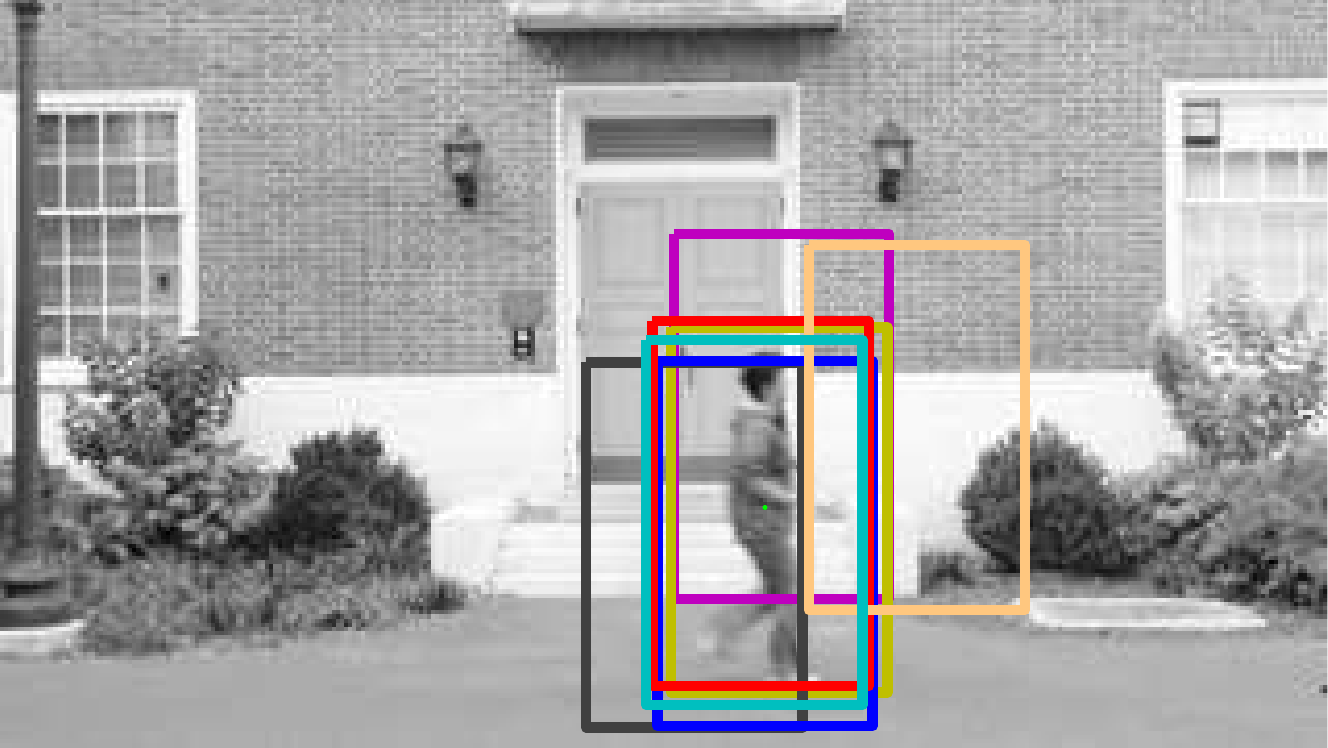}
\label{fig:run-3-backSub}
}
\subfigure{
\includegraphics[width=0.23\linewidth]{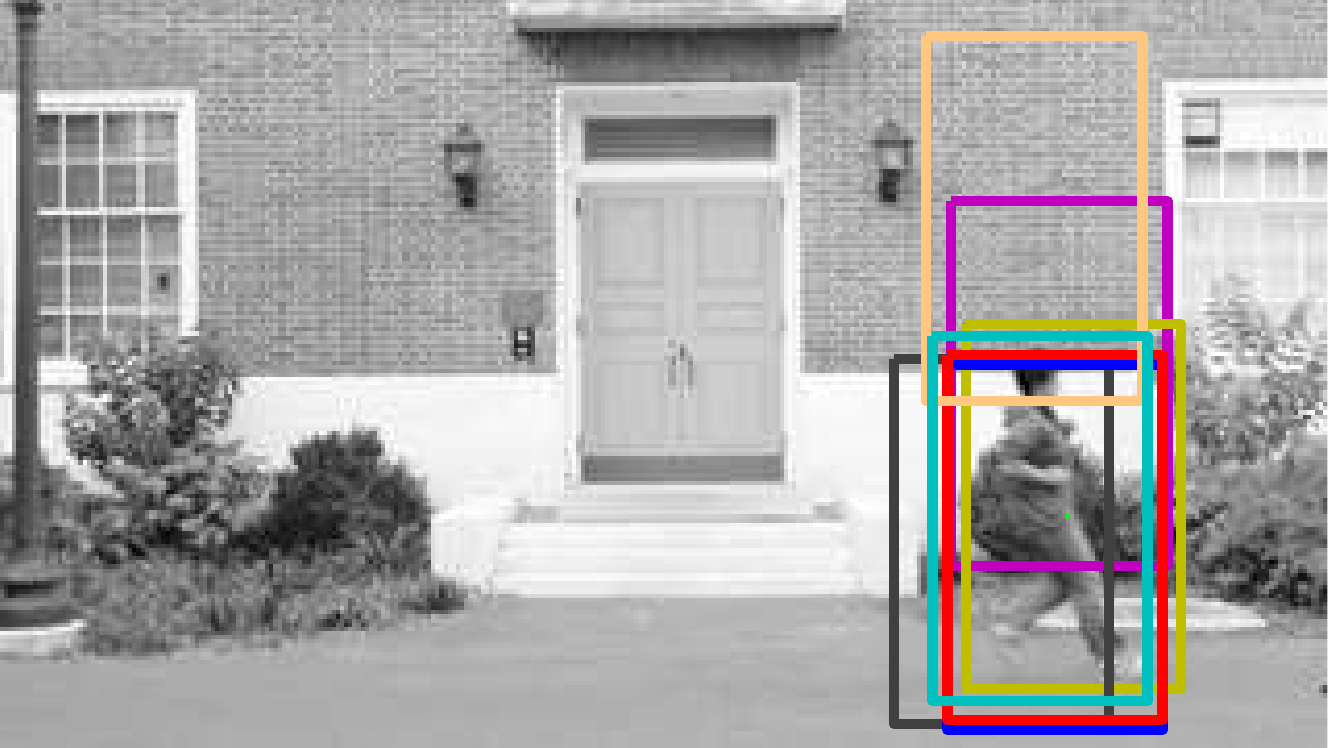}
\label{fig:run-4-backSub}
}
\caption{Using the dynamics of optical flow to track a human action. Parts-based human detector (red broken), Boost (magenta), TM (yellow), MS (black), MS-VR (blue) and MS-HR (red), DK-SSD-T (cyan). First row: Tracking a walking person, without any pre-processing for state-of-the-art methods. Second row: Tracking a walking person, with pre-processing for state-of-the-art methods. Third row: Tracking a running person, with pre-processing for state-of-the-art methods.}
\label{fig:action-tracking}
\end{figure*}

\section{Experiments on Tracking Human Actions} \label{sec:ActivityExperiments}

To demonstrate that our framework is general and can be applied to track dynamic visual phenomenon in any domain, we consider the problem of tracking humans while performing specific actions. This is different from the general problem of tracking humans, as we want to track humans performing specific actions such as walking or running. It has been shown, \eg~in \cite{Efros03, Chaudhry:CVPR09, Lin:ICCV09} that the optical flow generated by the motion of a person in the scene is characteristic of the action being performed by the human. In general global features extracted from optical flow perform better than intensity-based global features for action recognition tasks. The variation in the optical flow signature as a person performs an action displays very characteristic dynamics. We therefore model the variation in optical flow generated by the motion of a person in a scene as the output of a linear dynamical system and pose the action tracking problem in terms of matching the observed optical flow with a dynamic template of flow fields for that particular action.

\begin{figure*}[htb]
\centering
\subfigure{
\includegraphics*[width=0.47\linewidth]{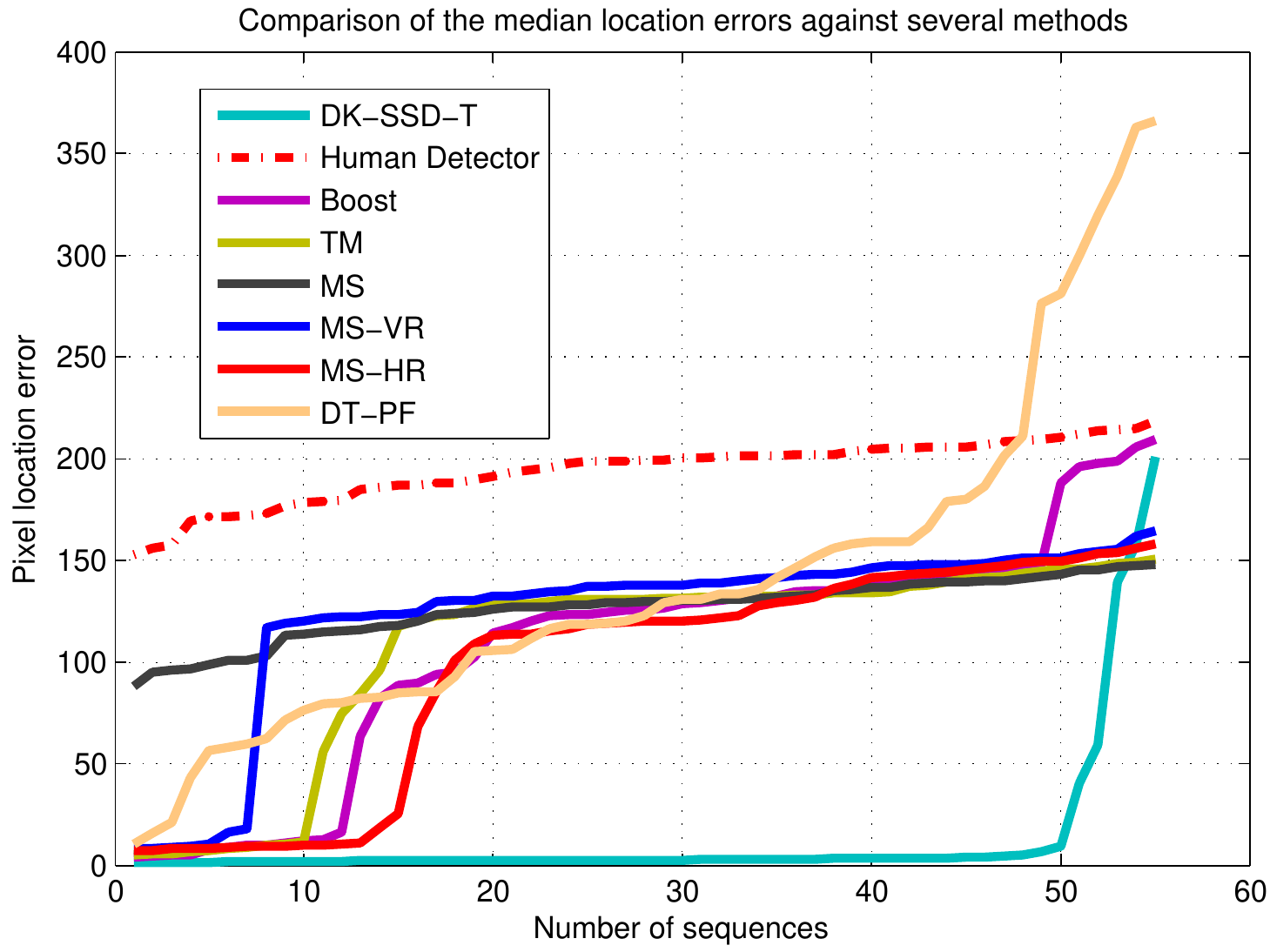}
\label{fig:trackerComp}
}
\subfigure{
\includegraphics*[width=0.47\linewidth]{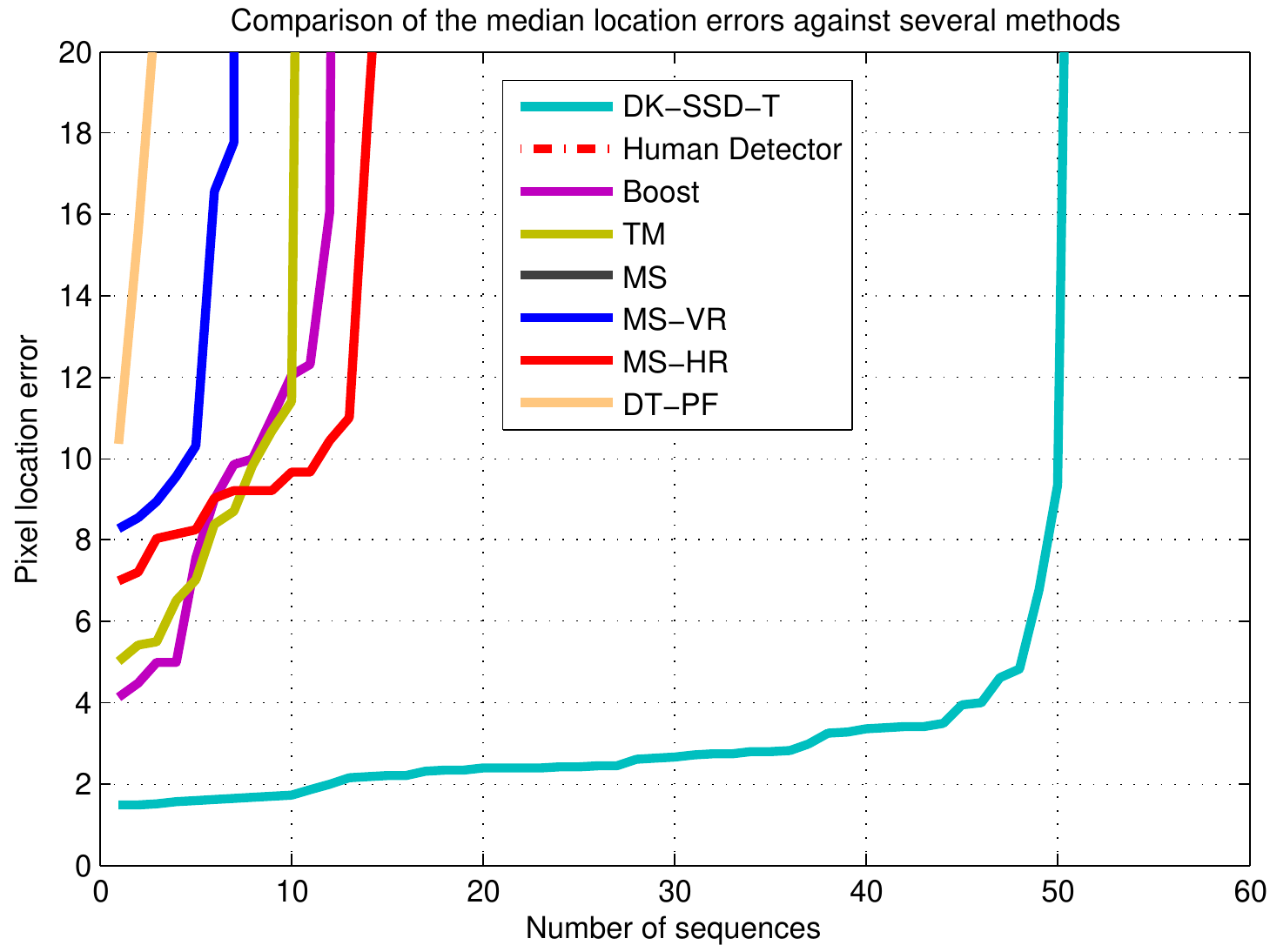}
\label{fig:trackerComp-20pix}
}
\\
\subfigure{
\includegraphics*[width=0.47\linewidth]{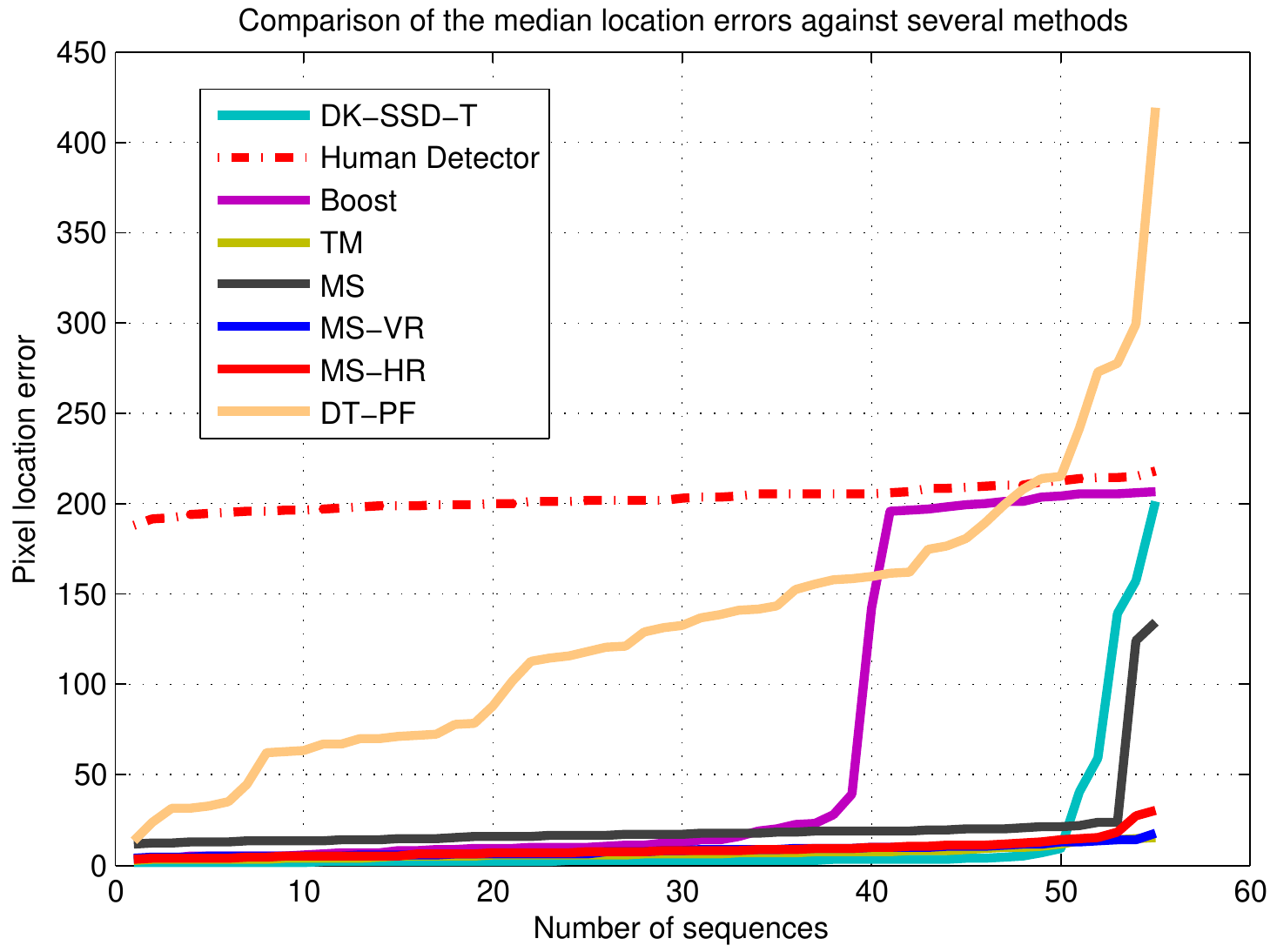}
\label{fig:trackerComp-backSub}
}
\subfigure{
\includegraphics*[width=0.47\linewidth]{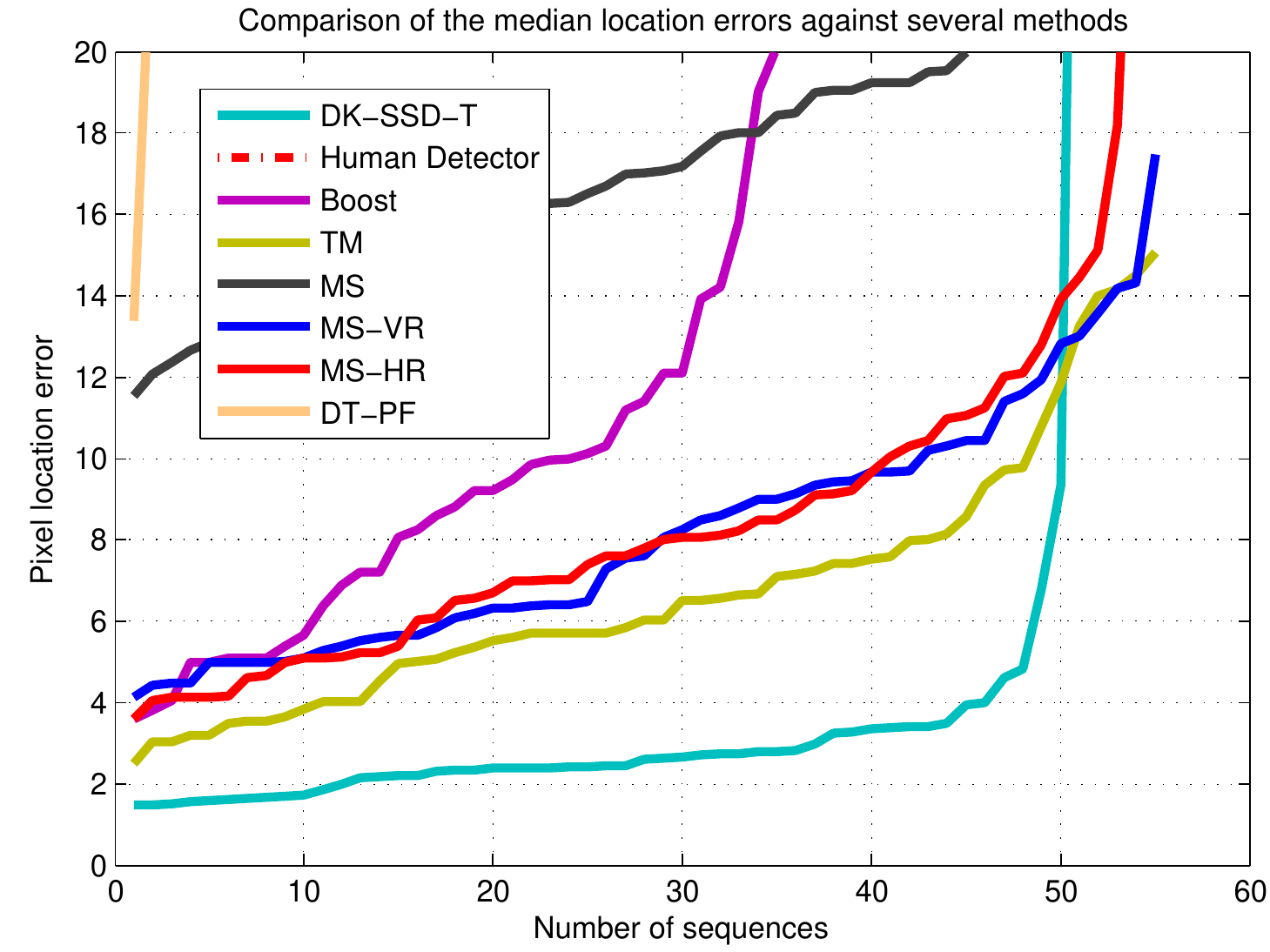}
\label{fig:trackerComp-20pix-backSub}
}
\caption{Comparison of various tracking methods without (top row) and with (bottom row) background subtraction. Figures on right are zoomed-in versions of figures on left.}
\label{fig:median-tracker-error}
\end{figure*}

We collect a dataset of 55 videos, each containing a single human performing either of two actions (walking and running). The videos are taken with a stationary camera, however the background has a small amount of dynamic content in the form of waving bushes. Simple background subtraction would therefore lead to erroneous bounding boxes. We manually extract bounding boxes and the corresponding centroids to mark the location of the person in each frame of the video. These are then used as ground-truth for later comparisons as well as to learn the dynamics of the optical flow generated by the person as they move in the scene.

For each bounding box centered at $\ll_t$, we extract the corresponding optical flow $\mathcal{F}(\ll_t) = [F_x(\ll_t), F_y(\ll_t)]$, and model the optical flow time-series, $\{\mathcal{F}(\ll_t)\}_{t=1}^T$ as a Linear Dynamical System. We extract the system parameters, $(\mu,A,C,Q,R)$ for each optical flow time-series using the system identification method in \S\ref{subsec:sysID}. This gives us the system parameters and ground-truth tracks for each of the 55 human action samples.

Given a test video, computing the tracks and internal state of the optical-flow dynamical system at each time instant amounts to minimizing the function,
\begin{align}
O(\ll_t,\x_t) = & \dfrac{1}{2R}\| \mathcal{F}(\ll_t) - (\mu+C\x_t)\|^2 + \nonumber \\
& \dfrac{1}{2}(\x_t - A\x_{t-1})^\top Q^{-1} (\x_t - A\x_{t-1}). \label{eq:opticalFlowJointOptimization}
\end{align}
To find the optimal $\ll_t$, and $\x_t$, \eqref{eq:opticalFlowJointOptimization} is optimized in the same gradient descent fashion as \eqref{eq:jointOptimization}.

We use the learnt system parameters in a leave-one-out fashion to track the activity in each test video sequence. Taking one sequence as a test sequence, we use all the remaining sequences as training sequences. The flow-dynamics system parameters extracted from each training sequence are used to track the action in the test sequence. Therefore, for each test sequence $j \in \{1,\ldots,N\}$, we get $N-1$ tracked locations and state estimate time-series by using all the remaining $N-1$ extracted system parameters. As described in \S\ref{sec:jointTrackingRecognition}, we choose the tracks that give the minimum objective function value in \eqref{eq:jointOptimizationAllFramesGivenTrain}.

Fig. \ref{fig:action-tracking} shows the tracking results against the state-of-the-art algorithms. Since this is a human activity tracking problem, we also compare our method against the parts-based human detector of \citet{Felzenszwalb:PAMI10} when trained on the PASCAL human dataset. We used the publicly available\footnote{\url{http://people.cs.uchicago.edu/~pff/latent/}} code for this comparison with default parameters and thresholds\footnote{It might be possible to achieve better detections on this dataset by tweaking parameters/thresholds. However we did not attempt this as it is not the focus of our paper.}. The detection with the highest probability is used as the location of the human in each frame. 

The first row in Fig. \ref{fig:action-tracking} shows the results for all the trackers when applied to the test video. The state-of-the-art trackers do not perform very well. In fact foreground-only trackers, MS, TM and DT-PF, lose tracks altogether. Our proposed method (cyan) gives the best tracking results and the best bounding box covering the person across all frames. The parts-based detector at times does not give any responses or spurious detections altogether, whereas Boost, MS-VR and MS-HR do not properly align with the human. Since we use optical flow as a feature, it might seem that there is an implicit form of background subtraction in our method. As a more fair comparison, we performed background subtraction as a pre-processing step on all the test videos before using the state-of-the-art trackers. The second row in Fig. \ref{fig:action-tracking} shows the tracking results for the same walking video as in row 1. We see that the tracking has improved but our tracker still gives the best results. The third row in Fig. \ref{fig:action-tracking} shows tracking results for a running person. Here again, our tracker performs the best whereas Boost and DT-PF perform the worst.

\begin{figure*}
\centering
\subfigure[DK-SSD-TR-R]{
\includegraphics[width=0.45\linewidth]{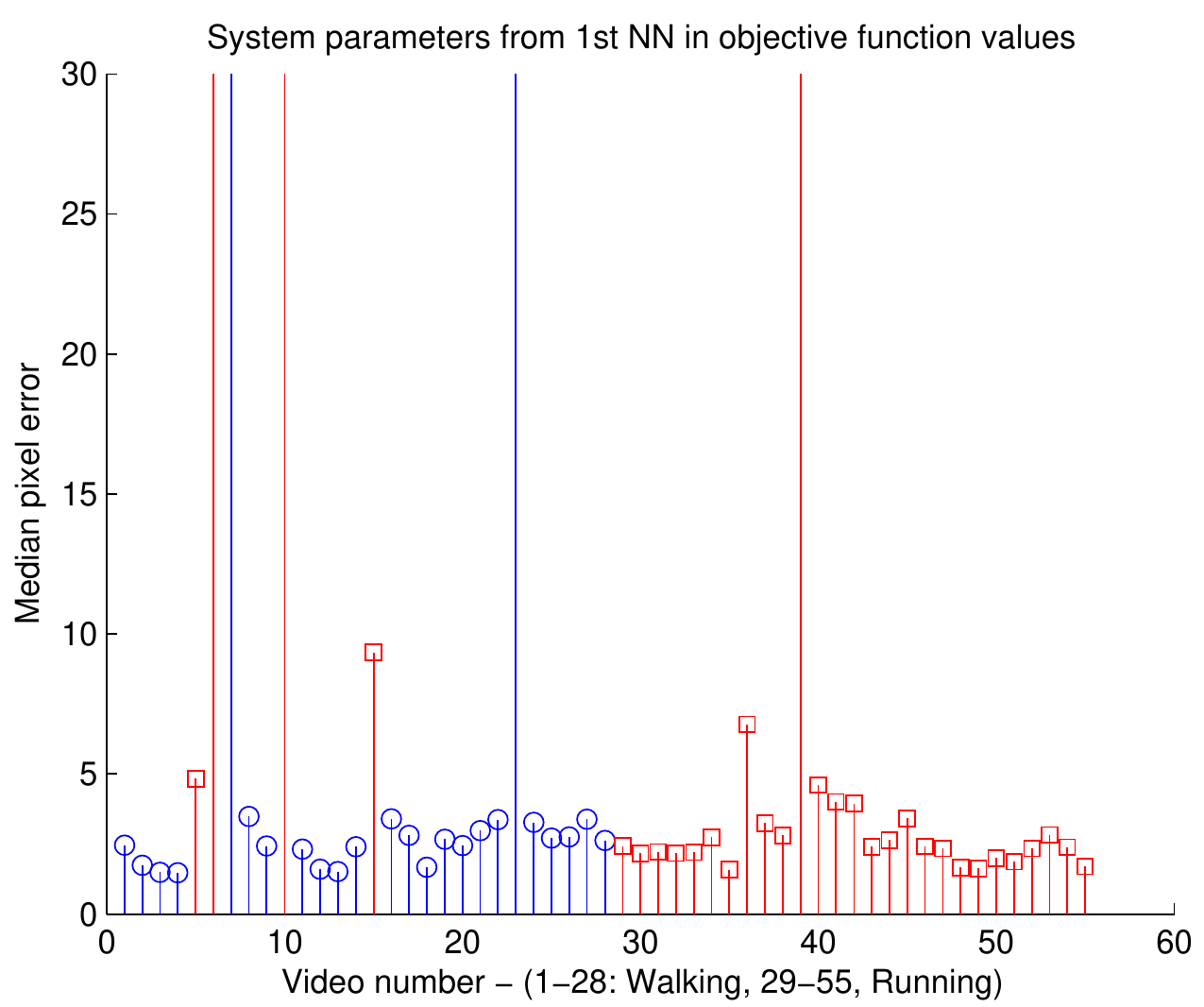}
\label{fig:tracking-error-against-class-objFunVal}
}
%
\subfigure[DK-SSD-TR-C]{
\includegraphics[width=0.45\linewidth]{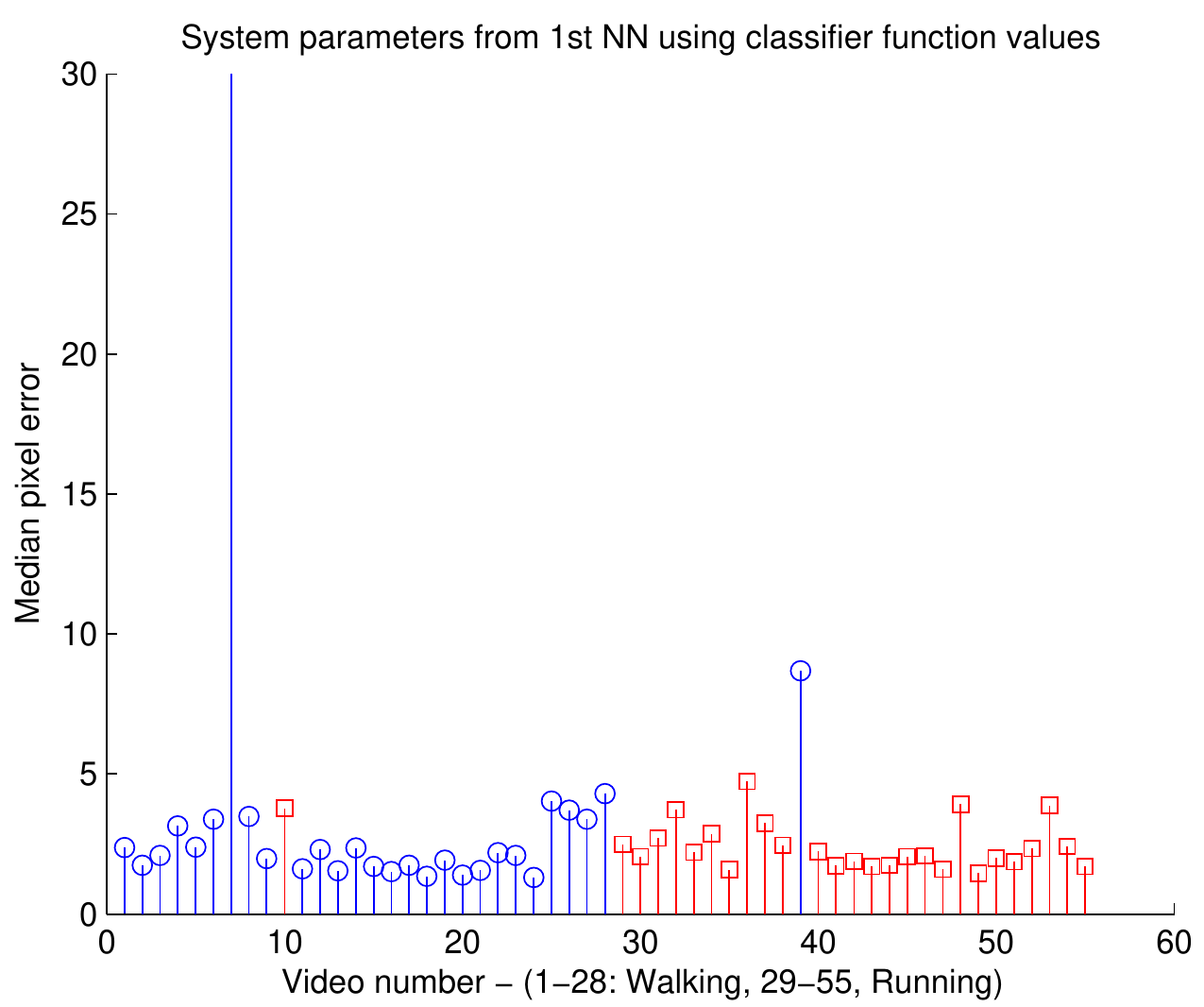}
\label{fig:tracking-error-against-class-classifier}
}
\caption{Simultaneous Tracking and Recognition results showing median tracking error and classification results, when using (a) DK-SSD-TR-R: the objective function in \eqref{eq:jointOptimizationAllFramesGivenTrain}, and (b) DK-SSD-TR-C: the Martin distance between dynamical systems in \eqref{eq:classificationCost} with a 1-NN classifier. (walking (blue), running (red)).}
\label{fig:tracking-error-against-class}
\end{figure*}

To derive quantitative conclusions about tracker performance, Fig. \ref{fig:median-tracker-error} shows the median tracker location error for each video sorted in ascending order. The best method should have the smallest error for most of the videos. As we can see, both without (1st row) and with background subtraction (2nd row), our method provides the smallest median location error against all state of the art methods for all except 5 sequences. Moreover, as a black box and without background subtraction as a pre-processing step, all state-of-the-art methods perform extremely poorly.


\section{Experiments on Simultaneous Action Tracking and Recognition} \label{sec:jointTrackingRecognitionExperiments}

In the previous section, we have shown that given training examples for actions, our tracking framework, DK-SSD-T, can be used to perform human action tracking using system parameters learnt from training data with correct tracks. We used the value of the objective function in \eqref{eq:opticalFlowJointOptimization} to select the tracking result. In this section, we will extensively test our simultaneous tracking and classification approach presented in \S\ref{sec:jointTrackingRecognition} on the two-action database introduced in the previous section as well as the commonly used Weizmann human action dataset \citep{Gorelick:PAMI07}. We will also show that we can learn the system parameters for a class of dynamic templates from one database and use it to simultaneously track and recognize the template in novel videos from other databases.

\begin{figure*}
\subfigure[Intensity and optical flow frame with ground-truth tracks (green) and tracked outputs of our algorithm (blue) on test video with walking person tracked using system parameters learnt from the walking person video in (b).]{
\includegraphics[width=0.48\linewidth]{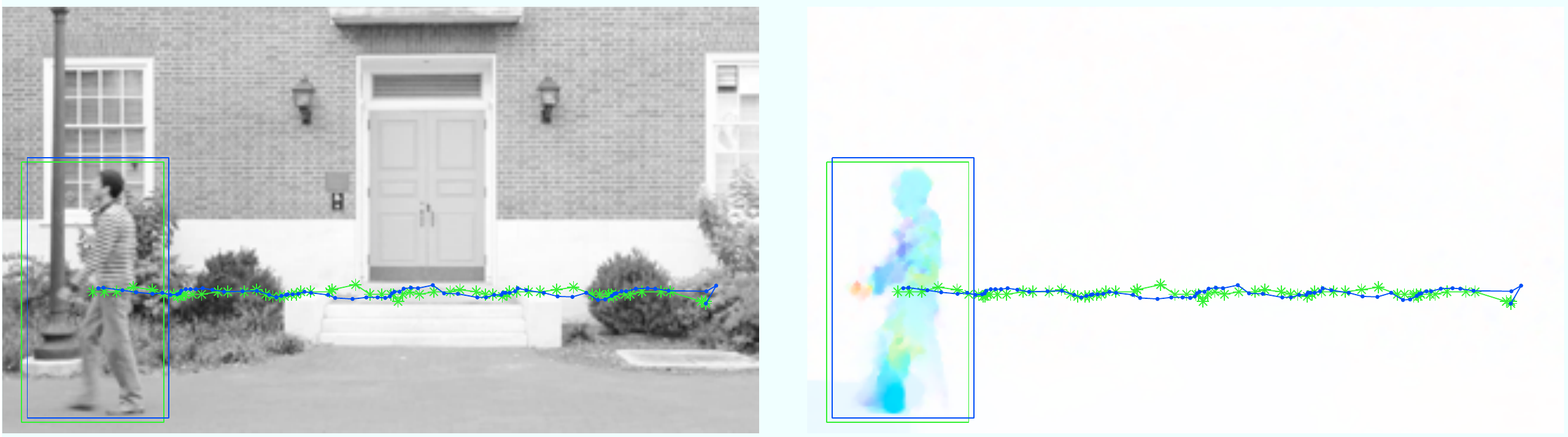}
\label{fig:trainWalkTestWalkTestVidFlowTracks}
}
\hfill
\subfigure[Intensity and optical flow frame with ground-truth tracks (green) used to train system parameters for a walking model.]{
\includegraphics[width=0.48\linewidth]{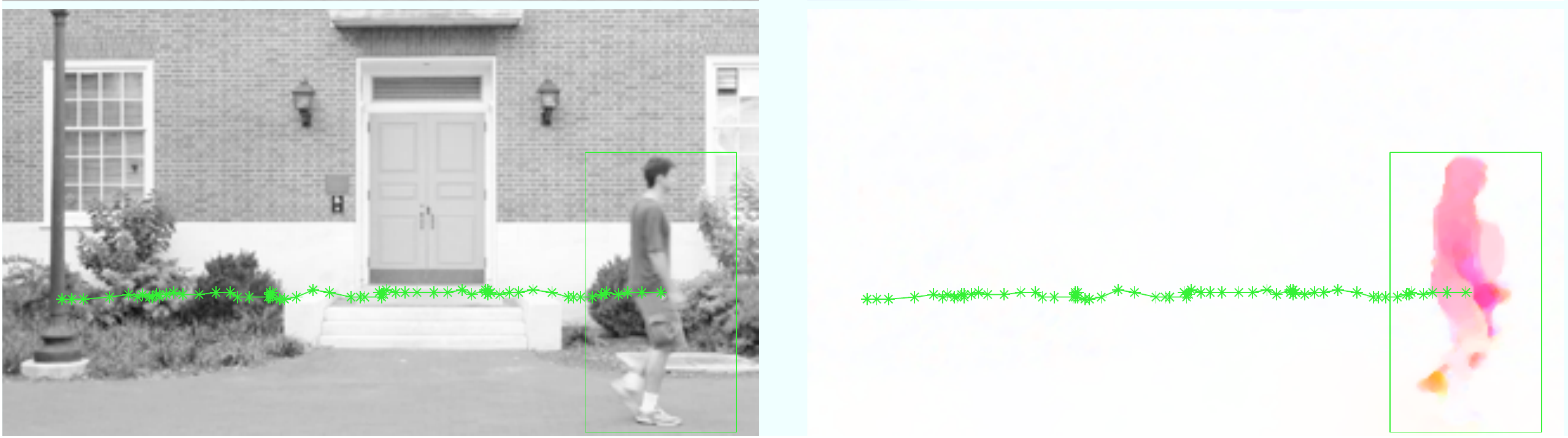}
\label{fig:trainWalkTestWalkTrainVidFlowTracks}
}

\subfigure[Optical flow bounding boxes at ground truth locations.]{
\centering
\includegraphics[width=0.06\linewidth]{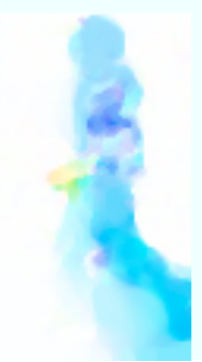}
\includegraphics[width=0.06\linewidth]{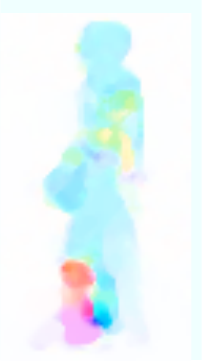}
\includegraphics[width=0.06\linewidth]{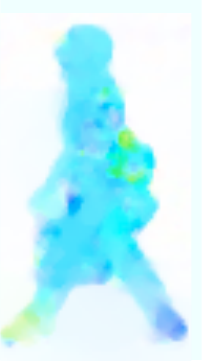}
\includegraphics[width=0.06\linewidth]{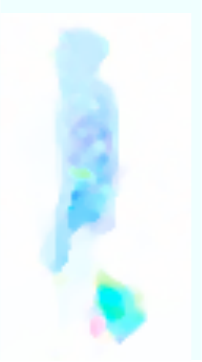}
\includegraphics[width=0.06\linewidth]{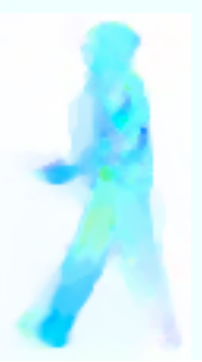}
\includegraphics[width=0.06\linewidth]{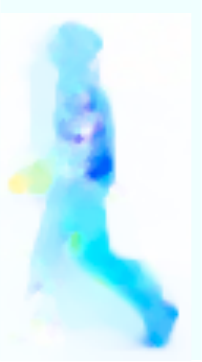}
\includegraphics[width=0.06\linewidth]{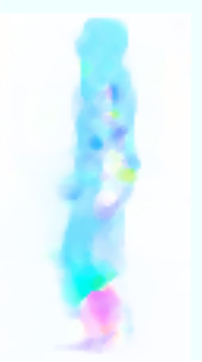}
\includegraphics[width=0.06\linewidth]{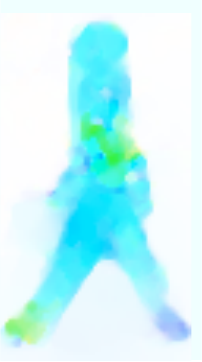}
\includegraphics[width=0.06\linewidth]{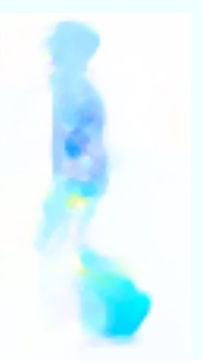}
\includegraphics[width=0.06\linewidth]{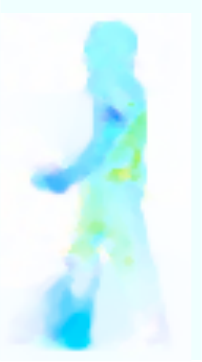}
\includegraphics[width=0.06\linewidth]{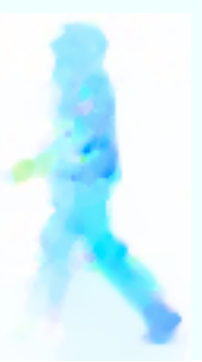}
\includegraphics[width=0.06\linewidth]{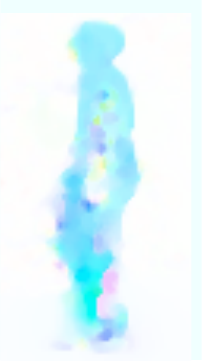}
\includegraphics[width=0.06\linewidth]{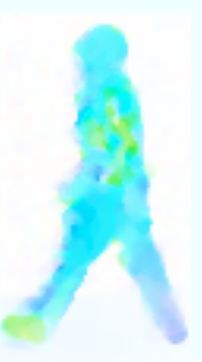}
\includegraphics[width=0.06\linewidth]{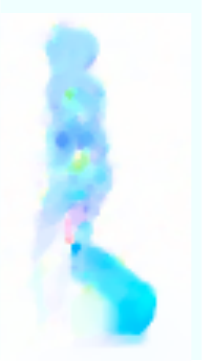}
\includegraphics[width=0.06\linewidth]{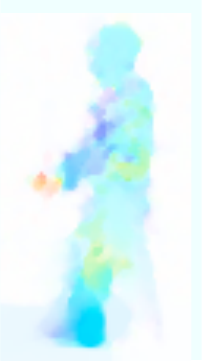}
\label{fig:trainWalkTestWalkFlowAtGT}
} 

\subfigure[Optical flow bounding boxes at tracked locations.]{
\centering
\includegraphics[width=0.06\linewidth]{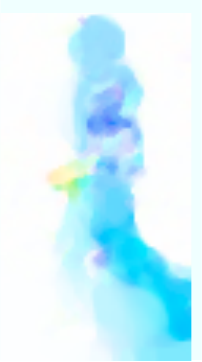}
\includegraphics[width=0.06\linewidth]{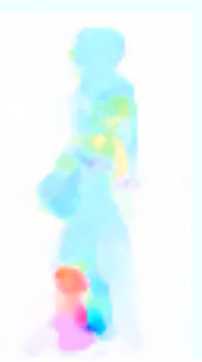}
\includegraphics[width=0.06\linewidth]{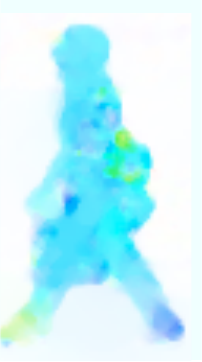}
\includegraphics[width=0.06\linewidth]{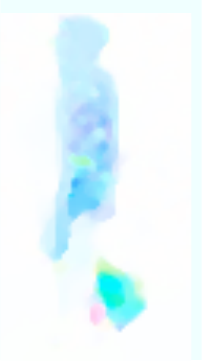}
\includegraphics[width=0.06\linewidth]{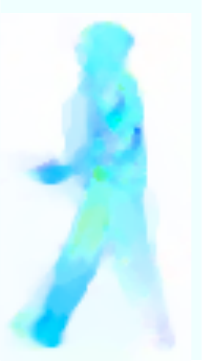}
\includegraphics[width=0.06\linewidth]{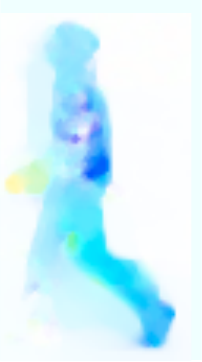}
\includegraphics[width=0.06\linewidth]{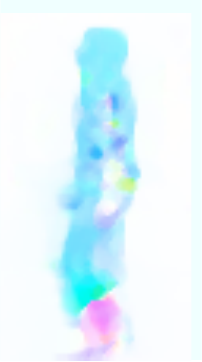}
\includegraphics[width=0.06\linewidth]{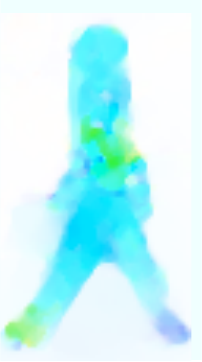}
\includegraphics[width=0.06\linewidth]{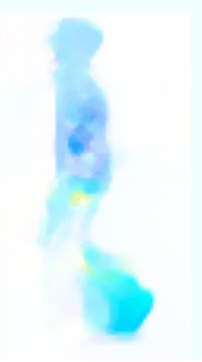}
\includegraphics[width=0.06\linewidth]{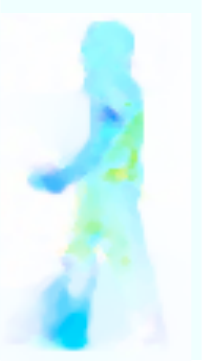}
\includegraphics[width=0.06\linewidth]{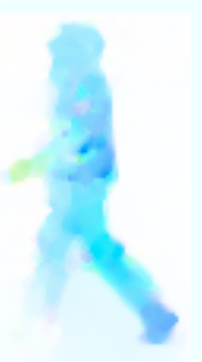}
\includegraphics[width=0.06\linewidth]{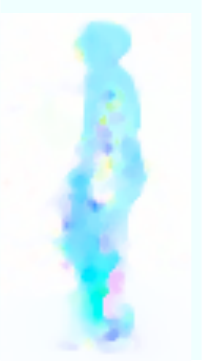}
\includegraphics[width=0.06\linewidth]{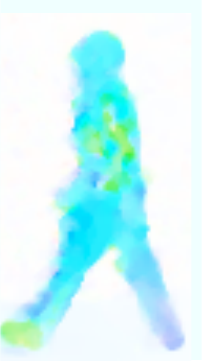}
\includegraphics[width=0.06\linewidth]{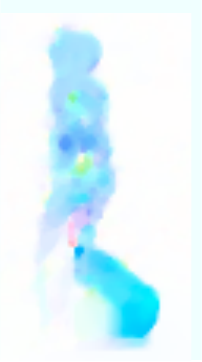}
\includegraphics[width=0.06\linewidth]{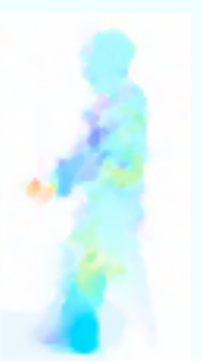}
\label{fig:trainWalkTestWalkFlowAtTrackedLoc}
}

\subfigure[Optical flow as generated by the computed states at corresponding time instants at the tracked locations.]{
\centering
\includegraphics[width=0.06\linewidth]{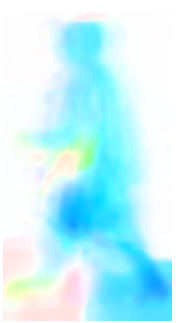}
\includegraphics[width=0.06\linewidth]{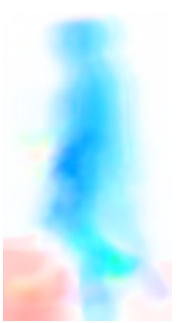}
\includegraphics[width=0.06\linewidth]{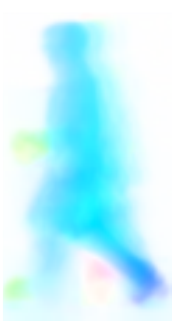}
\includegraphics[width=0.06\linewidth]{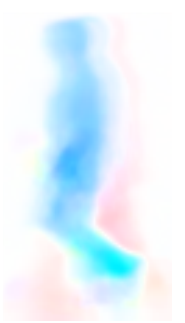}
\includegraphics[width=0.06\linewidth]{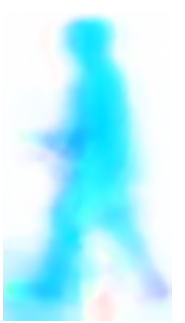}
\includegraphics[width=0.06\linewidth]{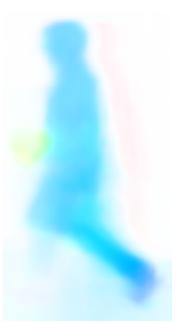}
\includegraphics[width=0.06\linewidth]{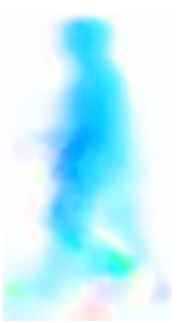}
\includegraphics[width=0.06\linewidth]{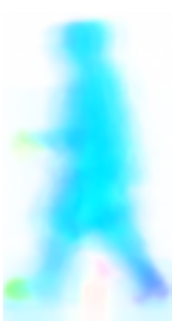}
\includegraphics[width=0.06\linewidth]{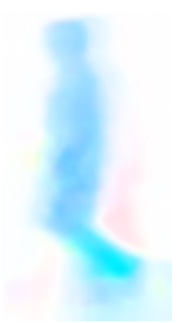}
\includegraphics[width=0.06\linewidth]{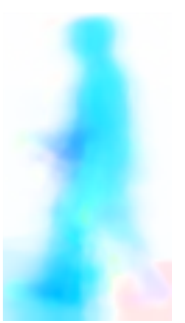}
\includegraphics[width=0.06\linewidth]{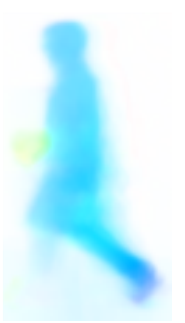}
\includegraphics[width=0.06\linewidth]{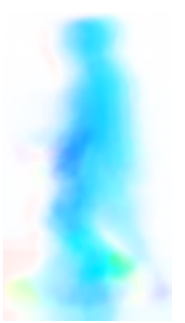}
\includegraphics[width=0.06\linewidth]{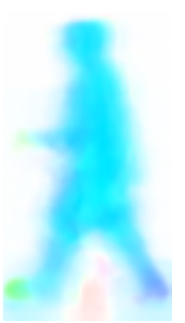}
\includegraphics[width=0.06\linewidth]{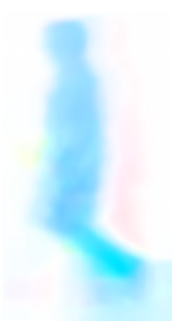}
\includegraphics[width=0.06\linewidth]{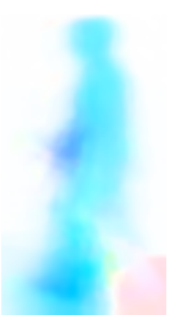}
\label{fig:trainWalkTestWalkGeneratedFlow}
}

\subfigure[Optical flow bounding boxes from the training video used to learn system parameters.]{
\centering
\includegraphics[width=0.06\linewidth]{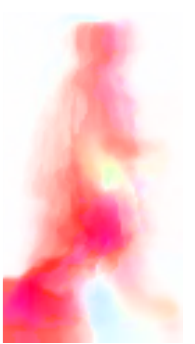}
\includegraphics[width=0.06\linewidth]{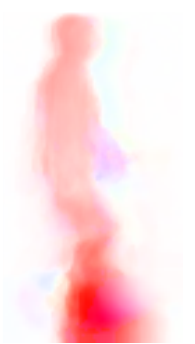}
\includegraphics[width=0.06\linewidth]{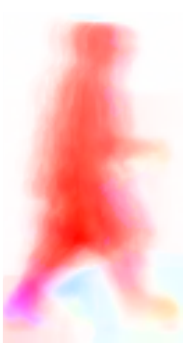}
\includegraphics[width=0.06\linewidth]{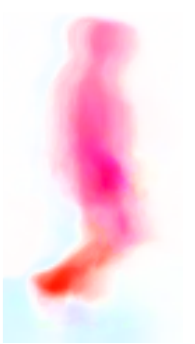}
\includegraphics[width=0.06\linewidth]{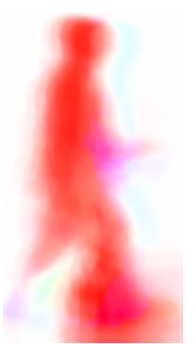}
\includegraphics[width=0.06\linewidth]{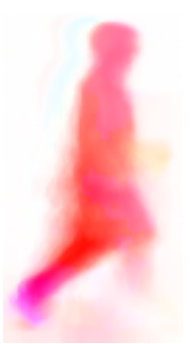}
\includegraphics[width=0.06\linewidth]{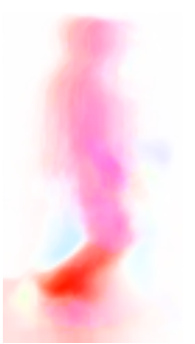}
\includegraphics[width=0.06\linewidth]{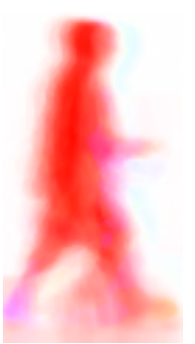}
\includegraphics[width=0.06\linewidth]{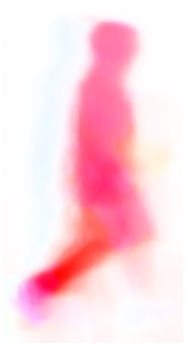}
\includegraphics[width=0.06\linewidth]{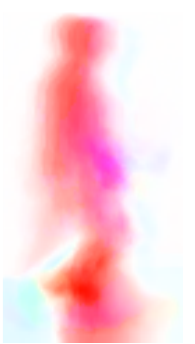}
\includegraphics[width=0.06\linewidth]{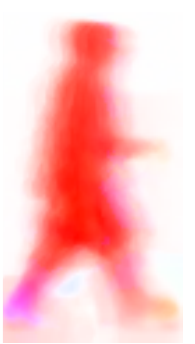}
\includegraphics[width=0.06\linewidth]{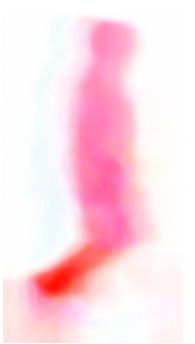}
\includegraphics[width=0.06\linewidth]{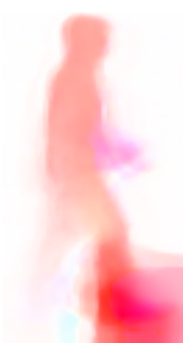}
\includegraphics[width=0.06\linewidth]{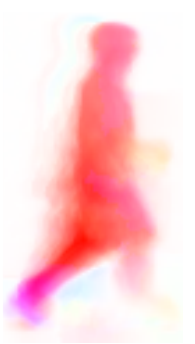}
\includegraphics[width=0.06\linewidth]{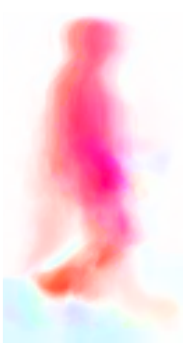}
\label{fig:trainWalkTestWalkTrainFlow}
}
\caption{Tracking a walking person using dynamical system parameters learnt from another walking person with opposite walking direction. The color of the optical flow diagrams represents the direction (\eg right to left is cyan, right to left is red) and the intensity represents the magnitude of the optical flow vector.}
\label{fig:trainWalkTestWalkFlowGeneration}
\end{figure*}

\begin{figure*}
\centering
\subfigure[Intensity and optical flow frame with ground-truth tracks (green) and tracked outputs of our algorithm (red) on test video with walking person tracked using system parameters learnt from the running person video in (b).]{
\includegraphics[width=0.48\linewidth]{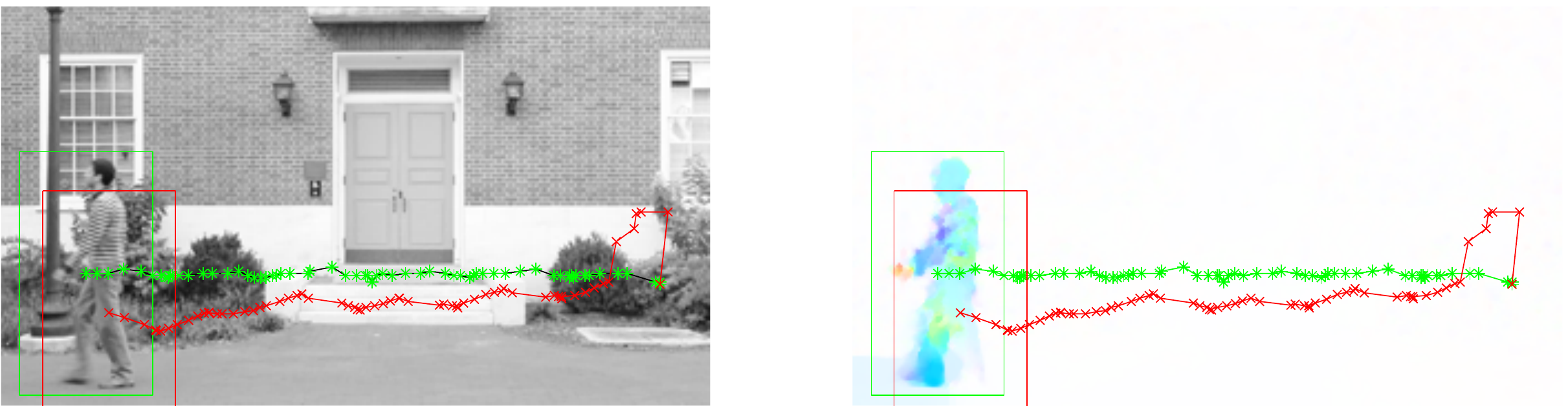}
\label{fig:trainRunTestWalkTestVidFlowTracks}
}
\hfill
\subfigure[Intensity and optical flow frame with ground-truth tracks (green) used to train system parameters for a running model.]{
\includegraphics[width=0.48\linewidth]{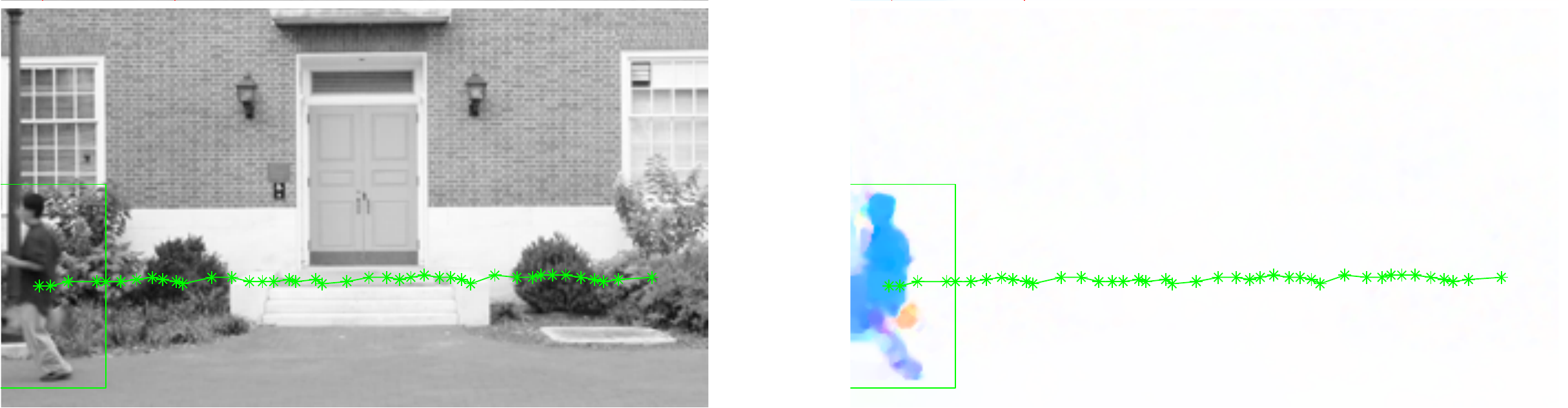}
\label{fig:trainRunTestWalkTrainVidFlowTracks}
}

\subfigure[Optical flow bounding boxes at ground truth locations.]{
\centering
\includegraphics[width=0.06\linewidth]{walking1-walking-8-flowFrame-00001.pdf}
\includegraphics[width=0.06\linewidth]{walking1-walking-8-flowFrame-00006.pdf}
\includegraphics[width=0.06\linewidth]{walking1-walking-8-flowFrame-00011.pdf}
\includegraphics[width=0.06\linewidth]{walking1-walking-8-flowFrame-00016.pdf}
\includegraphics[width=0.06\linewidth]{walking1-walking-8-flowFrame-00021.pdf}
\includegraphics[width=0.06\linewidth]{walking1-walking-8-flowFrame-00026.pdf}
\includegraphics[width=0.06\linewidth]{walking1-walking-8-flowFrame-00031.pdf}
\includegraphics[width=0.06\linewidth]{walking1-walking-8-flowFrame-00036.pdf}
\includegraphics[width=0.06\linewidth]{walking1-walking-8-flowFrame-00041.pdf}
\includegraphics[width=0.06\linewidth]{walking1-walking-8-flowFrame-00046.pdf}
\includegraphics[width=0.06\linewidth]{walking1-walking-8-flowFrame-00051.pdf}
\includegraphics[width=0.06\linewidth]{walking1-walking-8-flowFrame-00056.pdf}
\includegraphics[width=0.06\linewidth]{walking1-walking-8-flowFrame-00061.pdf}
\includegraphics[width=0.06\linewidth]{walking1-walking-8-flowFrame-00066.pdf}
\includegraphics[width=0.06\linewidth]{walking1-walking-8-flowFrame-00071.pdf}
\label{fig:trainRunTestWalkFlowAtGT}
}

\subfigure[Optical flow bounding boxes at tracked locations.]{
\centering
\includegraphics[width=0.06\linewidth]{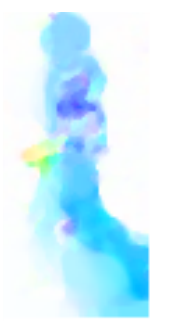}
\includegraphics[width=0.06\linewidth]{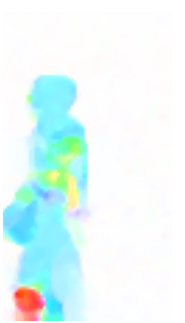}
\includegraphics[width=0.06\linewidth]{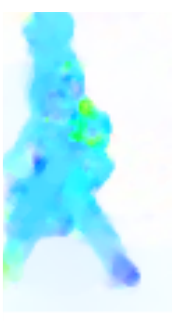}
\includegraphics[width=0.06\linewidth]{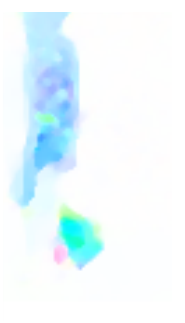}
\includegraphics[width=0.06\linewidth]{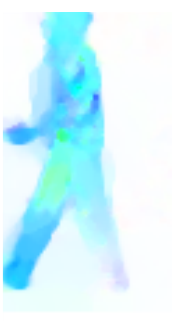}
\includegraphics[width=0.06\linewidth]{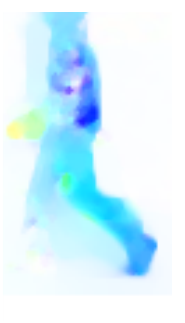}
\includegraphics[width=0.06\linewidth]{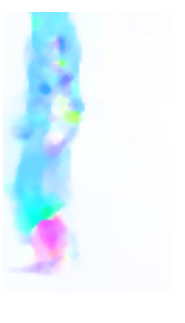}
\includegraphics[width=0.06\linewidth]{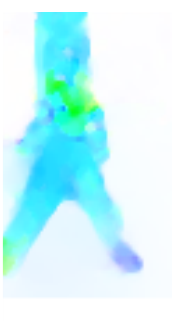}
\includegraphics[width=0.06\linewidth]{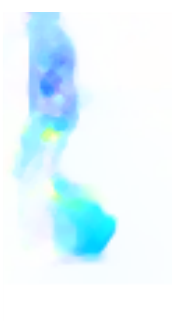}
\includegraphics[width=0.06\linewidth]{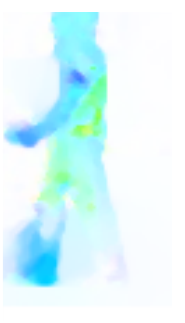}
\includegraphics[width=0.06\linewidth]{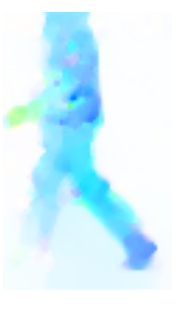}
\includegraphics[width=0.06\linewidth]{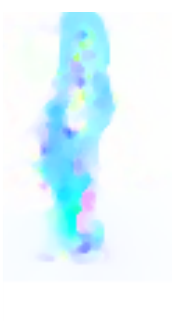}
\includegraphics[width=0.06\linewidth]{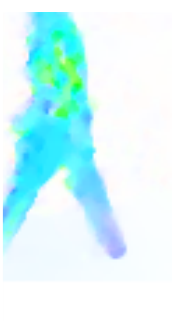}
\includegraphics[width=0.06\linewidth]{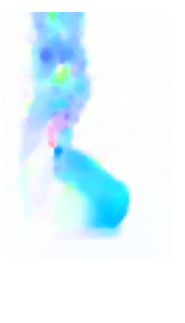}
\includegraphics[width=0.06\linewidth]{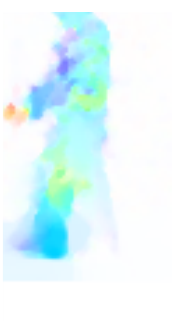}
\label{fig:trainRunTestWalkFlowAtTrackedLoc}
}

\subfigure[Optical flow as generated by the computed states at corresponding time instants at the tracked locations.]{
\centering
\includegraphics[width=0.06\linewidth]{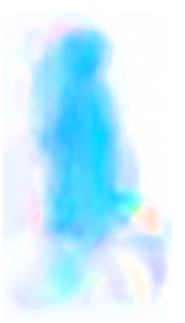}
\includegraphics[width=0.06\linewidth]{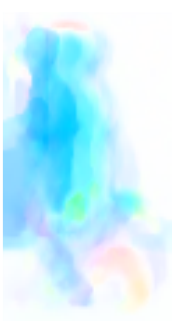}
\includegraphics[width=0.06\linewidth]{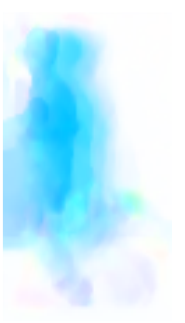}
\includegraphics[width=0.06\linewidth]{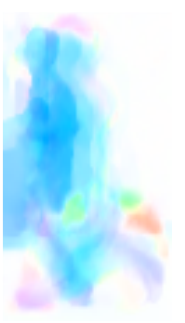}
\includegraphics[width=0.06\linewidth]{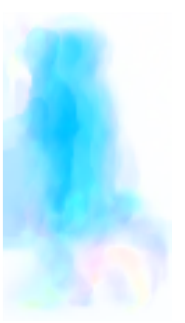}
\includegraphics[width=0.06\linewidth]{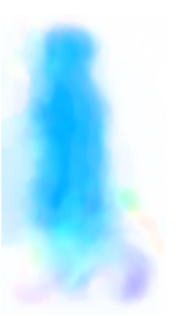}
\includegraphics[width=0.06\linewidth]{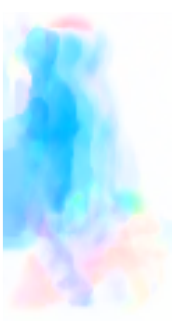}
\includegraphics[width=0.06\linewidth]{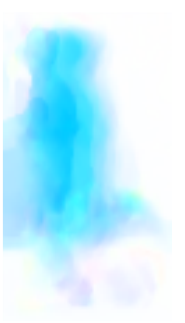}
\includegraphics[width=0.06\linewidth]{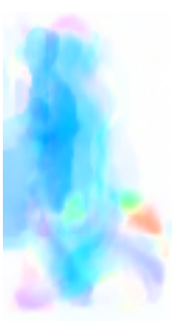}
\includegraphics[width=0.06\linewidth]{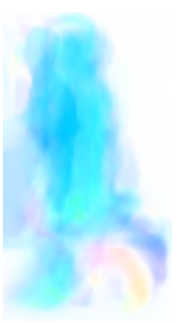}
\includegraphics[width=0.06\linewidth]{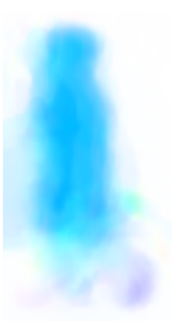}
\includegraphics[width=0.06\linewidth]{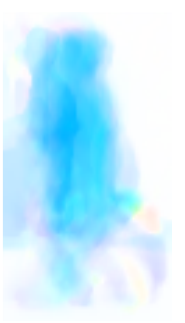}
\includegraphics[width=0.06\linewidth]{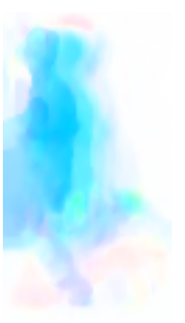}
\includegraphics[width=0.06\linewidth]{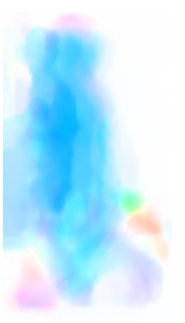}
\includegraphics[width=0.06\linewidth]{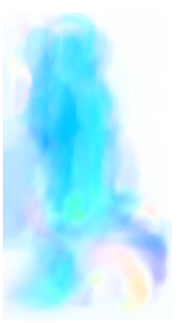}
\label{fig:trainRunTestWalkGeneratedFlow}
}

\subfigure[Optical flow bounding boxes from the training video used to learn system parameters.]{
\centering
\qquad\qquad
\includegraphics[width=0.06\linewidth]{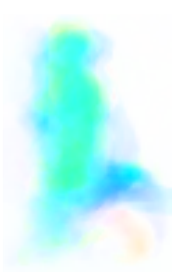}
\includegraphics[width=0.06\linewidth]{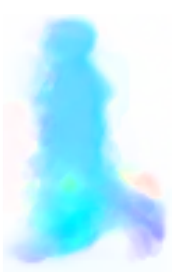}
\includegraphics[width=0.06\linewidth]{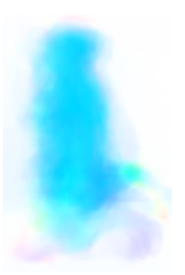}
\includegraphics[width=0.06\linewidth]{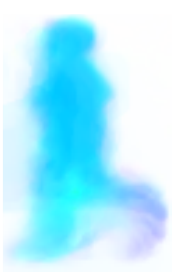}
\includegraphics[width=0.06\linewidth]{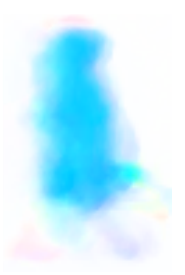}
\includegraphics[width=0.06\linewidth]{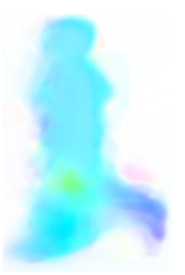}
\includegraphics[width=0.06\linewidth]{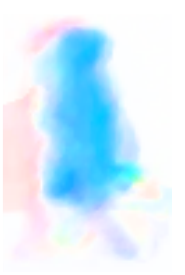}
\includegraphics[width=0.06\linewidth]{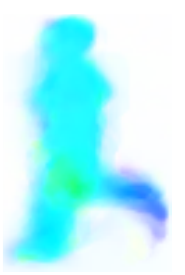}
\includegraphics[width=0.06\linewidth]{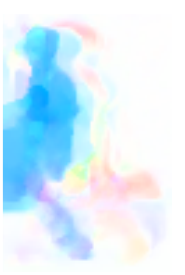}
\qquad\qquad
\label{fig:trainRunTestWalkTrainFlow}
}
\caption{Tracking a walking person using dynamical system parameters learnt from a running person. The color of the optical flow diagrams represents the direction (\eg right to left is cyan, right to left is red) and the intensity represents the magnitude of the optical flow vector.}
\label{fig:trainRunTestWalkFlowGeneration}
\end{figure*}

\subsection{Walking/Running Database} \label{subsec:HopkinsActionsExperiments}

\figref{fig:tracking-error-against-class} shows the median pixel tracking error of each test sequence using the leave-one-out validation described in the previous section for selecting the tracks, sorted by true action class. The first 28 sequences belong to the class \emph{walking}, while the remaining 27 sequences belong to the \emph{running} class. Sequences identified by our proposed approach as walking are colored blue, whereas sequences identified as running are colored red. \figref{fig:tracking-error-against-class-objFunVal} shows the tracking error and classification result when using DK-SSD-TR-R, \ie the objective function value in \eqref{eq:jointOptimizationAllFramesGivenTrain} with a 1-NN classifier to simultaneously track and recognize the action. The tracking results shown also correspond to the ones shown in the previous section. As we can see, for almost all sequences, the tracking error is within 5 pixels from the ground truth tracks. Moreover, for all but 4 sequences, the action is classified correctly, leading to an overall action classification rate of 93\%. \figref{fig:tracking-error-against-class-classifier} shows the tracking error and class labels when we using DK-SSD-TR-C, \ie the Martin distance based classifier term proposed in \eqref{eq:classificationCost} to simultaneously track and recognize the action. As we can see, DK-SSD-TR-C results in even better tracking and classification results. Only two sequences are mis-classified for an overall recognition rate of 96\%.

To show that our simultaneous tracking and classification framework computes the correct state of the dynamical system for the test video given that the class of the training system is the same, we illustrate several components of the action tracking framework for two cases: a right to left walking person tracked using dynamical system parameters learnt from another walking person moving left to right, in \figref{fig:trainWalkTestWalkFlowGeneration}, and the same person tracked using dynamical system parameters learnt from a running person in \figref{fig:trainRunTestWalkFlowGeneration}.

\figref{fig:trainWalkTestWalkTestVidFlowTracks} shows a frame with its corresponding optical flow, ground truth tracks (green) as well as the tracks computed using our algorithm (blue). As we can see the computed tracks accurately line-up with the ground-truth tracks. \figref{fig:trainWalkTestWalkTrainVidFlowTracks} shows a frame and corresponding optical flow along with the ground-truth tracks used to extract optical flow bounding boxes to learn the dynamical system parameters. \figref{fig:trainWalkTestWalkFlowAtGT} shows the optical flow extracted from the bounding boxes at the ground-truth locations in the test video at intervals of 5 frames and \figref{fig:trainWalkTestWalkFlowAtTrackedLoc} shows the optical flow extracted from the bounding boxes at the tracked locations at the same frame numbers. As the extracted tracks are very accurate, the flow-bounding boxes line up very accurately. Since our dynamical system model is generative, at each time-instant, we can use the computed state, $\hat{\x}_t$, to generate the corresponding output $\hat{\y}_t = \mu + C \hat{\x}_t$. \figref{fig:trainWalkTestWalkGeneratedFlow} displays the optical flow computed in this manner at the corresponding frames in \figref{fig:trainWalkTestWalkFlowAtTrackedLoc}. We can see that the generated flow appears like a smoothed version of the observed flow at the correct location. This shows that the internal state of the system was correctly computed according to the training system parameters, which leads to accurate dynamic template generation and tracking.
\figref{fig:trainWalkTestWalkTrainFlow} shows the optical flow at the bounding boxes extracted from the ground-truth locations in the training video. The direction of motion is the opposite as in the test video, however using the mean optical flow direction, the system parameters can be appropriately transformed at test time to account for this change in direction as discussed in \S\ref{subsec:scaleDirInv}.

\figref{fig:trainRunTestWalkTestVidFlowTracks} repeats the above experiment when tracking the same walking person with a running model. As we can see in \figref{fig:trainRunTestWalkTestVidFlowTracks}, the computed tracks are not very accurate when using the wrong action class for tracking. This is also evident in the extracted flow bounding boxes at the tracked locations as the head of the person is missing from almost all of the boxes. \figref{fig:trainRunTestWalkTrainFlow} shows several bounding box optical flow from the training video of the running person. The states computed using the learnt dynamical system parameters from these bounding boxes leads to the generated flows in \figref{fig:trainRunTestWalkGeneratedFlow} at the frames corresponding to those in \figref{fig:trainRunTestWalkFlowAtTrackedLoc}. The generated flow does not match the ground-truth flow and leads to a high objective function value.

\subsection{Comparison to Tracking then Recognizing}

We will now compare our simultaneous tracking and recognition approaches, DK-SSD-TR-R and DK-SSD-TR-C to the tracking \emph{then} recognizing approach where we first track the action using standard tracking algorithms described in \S\ref{sec:DTExperiments} as well as our proposed dynamic tracker DK-SSD-T, and then use the Martin distance for dynamical systems with a 1-NN classifier to classify the tracked action.

Given a test video, we compute the location of the person using all the trackers described in \S\ref{sec:DTExperiments} and then extract the bounding box around the tracked locations. For bounding boxes that do not cover the image area, we zero-pad the optical flow. This gives us an optical-flow time-series corresponding to the extracted tracks. We then use the approach described in \S\ref{subsec:sysID} to learn the dynamical system parameters of this optical flow time-series. To classify the tracked action, we compute the Martin distance of the tracked system to all the training systems in the database and use 1-NN to classify the action. We then average the results over all sequences in a leave-one-out fashion. 

Table \ref{tab:recAfterTrack} shows the recognition results for the 2-class Walking/Running database by using this classification scheme after performing tracking as well as our proposed simultaneous tracking and recognition algorithms. We have provided results for both the original sequences as well as when background subtraction was performed prior to tracking. We showed in \figref{fig:trackerComp} and \figref{fig:trackerComp-backSub} that all the standard trackers performed poorly without using background subtraction. Our tracking \emph{then} recognizing method, DK-SSD-T+R, provides accurate tracks and therefore gives a recognition rate of 96.36\%. The parts-based human detector of \citet{Felzenszwalb:PAMI10} tracker fails to detect the person in some frames and therefore the recognition rate is the worst. When using background subtraction to pre-process the videos, the best recognition rate is provided by TM at 98.18\% whereas MS-HR performs at the same rate as our proposed method. The recognition rate of the other methods, except the human detector, also increase due to this pre-processing step. For comparison, if we use the ground-truth tracks to learn the dynamical system parameters and classify the action using the Martin distance and 1-NN classification, we get 100\% recognition.

Our joint-optimization scheme using the objective function value as the classifier, DK-SSD-TR-R, performs slightly worse at 92.73\%, than the best tracking \emph{then} recognizing approach. However, when we use the Martin-distance based classification cost in DK-SSD-TR-C, we get the best action classification performance, 96.36\%. Needless to say the tracking \emph{then} recognizing scheme is more computationally intensive and is a 2-step procedure. Moreover, our joint optimization scheme based only on foreground dynamics performs better than tracking \emph{then} recognizing using all other trackers without pre-processing. At this point, we would also like to note that even though the best recognition percentage achieved by the tracking-then-recognize method was 98.18\% using TM, it performed worse in tracking as shown in \figref{fig:trackerComp-20pix-backSub}. Overall, our method simultaneously gives the best tracking and recognition performance.

\begin{table}[t]
\centering
\begin{tabular}{|l||p{2cm}|p{2cm}|}
\hline
Method & Recognition \% without background subtraction & Recognition \% with background subtraction \\
\hline
\hline
Track \emph{then} recognize & & \\
 \quad Boost & 81.82 & 81.82 \\
 \quad TM & 78.18 & 98.18 \\
 \quad MS & 67.27 & 89.09 \\
 \quad MS-VR & 74.55 & 94.55 \\
 \quad MS-HR & 78.18 & 96.36 \\
 \quad Human-detector & 45.45 & 50.91 \\
 \quad DT-PF & 69.09 & 72.73 \\
\hline
 \quad DK-SSD-T+R & 96.36 & - \\
\hline
 \quad Ground-truth tracks & 100 & - \\
\hline
\hline
Joint-optimization & & \\
 \quad DK-SSD-TR-R & 92.73 & - \\
 \quad DK-SSD-TR-C &  96.36 & - \\
\hline
\end{tabular}
\caption{Recognition rates for the 2-class Walking/Running database using 1-NN with Martin distance for dynamical systems after computing tracks from different algorithms}
\label{tab:recAfterTrack}
\end{table}

\subsection{Weizmann Action Database}

We also tested our simultaneous tracking and testing framework on the Weizmann Action database \citep{Gorelick:PAMI07}. This database consists of 10 actions with a total of 93 sequencess and contains both stationary actions such as jumping in place, bending \etc, and non-stationary actions such as running, walking, \etc~We used the provided backgrounds to extract bounding boxes and ground-truth tracks from all the sequences and learnt the parameters of the optical-flow dynamical systems using the same approach as outlined earlier. A commonly used evaluation scheme for the Weizmann database is leave-one-out classification. Therefore, we also used our proposed framework to track the action in a test video given the system parameters of all the remaining actions.

\begin{table}[t]
\centering
\begin{tabular}{|l|c||l|c|}
\hline
Action 	& Median $\pm$ RSE 	& Action 	& Median $\pm$ RSE \\
\hline
\hline
Bend 		& 5.7 $\pm$ 1.9 		& Side 	& 2.1 $\pm$ 0.9	\\
Jack 		& 4.2 $\pm$ 0.6		& Skip 	& 2.8 $\pm$ 1.1	\\
Jump 		& 1.5 $\pm$ 0.2 		& Walk 	& 3.3 $\pm$ 2.4	\\
PJump 	& 3.7 $\pm$ 1.3 		& Wave1 	& 14.8 $\pm$ 7.4 	\\
Run 		& 4.2 $\pm$ 2.9		& Wave2 	& 2.3 $\pm$ 0.7	\\
\hline
\end{tabular}
\caption{Median and robust standard error (RSE) (see text) of the tracking error for the Weizmann Action database, using the proposed simultaneous tracking and recognition approach with the classification cost. We get an overall action recognition rate of 92.47\%.}
\label{tab:weizmannTrackAndRec}
\end{table}

\begin{figure}
\centering
\includegraphics[width=\linewidth]{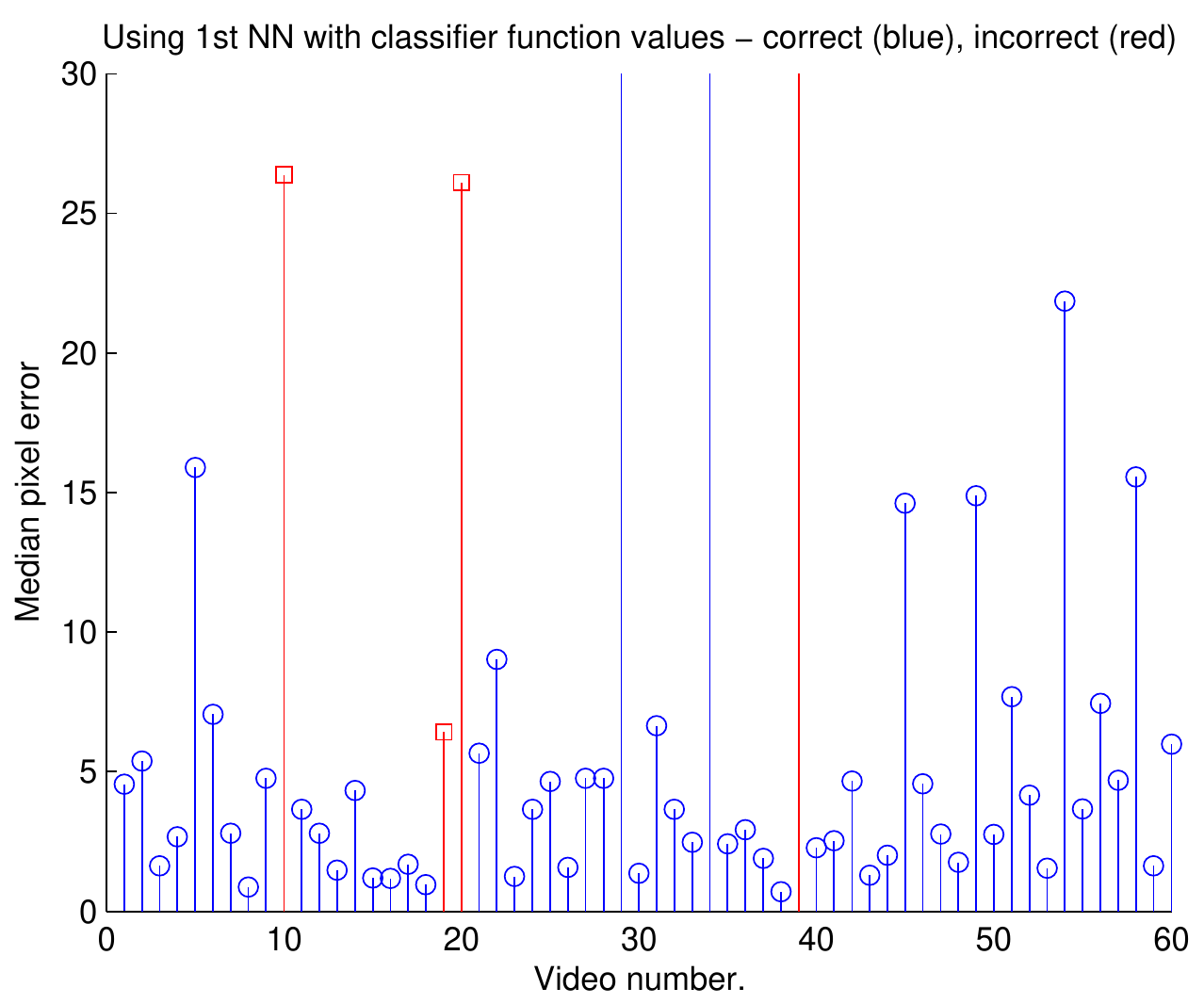}
\caption{Simultaneous Tracking and Recognition results showing median tracking error and classification results, when using the objective function, and when using the Martin distance between dynamical systems with a 1-NN classifier. (walking (blue), running (red)).}
\label{fig:tracking-error-against-class-weizmann}
\end{figure}

\begin{figure}
\centering
\includegraphics[width=0.8\linewidth]{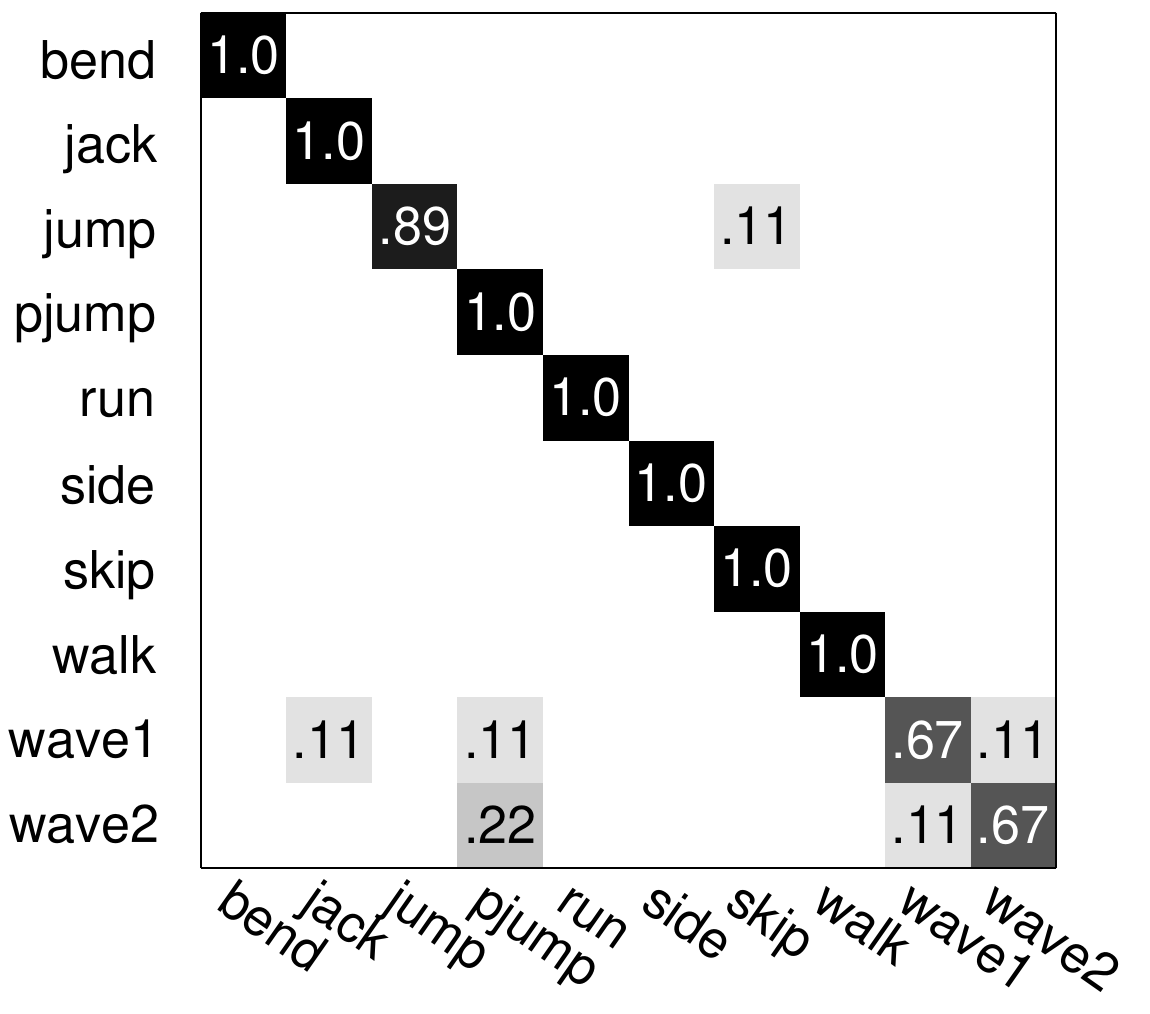}
\caption{Confusion matrix for leave-one-out classification on the Weizmann database using our simultaneous tracking and recognition approach. Overall recognition rate of 92.47\%.}
\label{fig:weizmannConfMat}
\end{figure}

Table \ref{tab:weizmannTrackAndRec} shows the median and robust standard error (RSE), \ie the square-root of the median of $\{(x-\text{median}(x))^2\}$, of the tracking error for each class. Other than Wave1, all classes have a median tracking error under 6 pixels as well as very small deviations from the median error. 
Furthermore, we get a simultaneous recognition rate of 92.47\% which corresponds to only 7 mis-classified sequences. \figref{fig:weizmannConfMat} shows the corresponding confusion matrix. \figref{fig:tracking-error-against-class-weizmann} shows the median tracking error per frame for each of the 93 sequences in the Weizmann database. The color of the stem-plot indicates whether the sequence was classified correctly (blue) or incorrectly (red). Table \ref{tab:weizmannComparison} shows the recognition rate of some state-of-the-art methods on the Weizmann database. However notice that, all these methods are geared towards recognition and either assume that tracking has been accurately done before the recognition is performed, or use spatio-temporal features for recognition that can not be used for accurate tracking. Furthermore, if the ground-truth tracks were provided, using the Martin distance between dynamical systems with a 1-NN classifier gives a recognition rate of 96.77\%. Our simultaneous tracking and recognition approach is very close to this performance. The method by \citet{Xie:CVPR11} seems to be the only attempt to simultaneously locate and recognize human actions in videos. However their method does not perform tracking, instead it extends the parts-based detector by \citet{Felzenszwalb:PAMI10} to explicitly consider temporal variations caused by various actions. Moreover, they do not have any tracking results in their paper other than a few qualitative detection results. Our approach is the first to explicitly enable simultaneous tracking and recognition of dynamic templates that is generalizable to any dynamic visual phenomenon and not just human actions.

\begin{table}
\centering
\begin{tabular}{|l||c|}
\hline
Method 				& Recognition (\%) \\
\hline
\hline
\cite{Xie:CVPR11} 			& 95.60 \\
\cite{Thurau:CVPR08}		& 94.40 \\
\cite{Ikizler:IVC09} 			& 100.00 \\
\cite{Gorelick:PAMI07}		& 99.60 \\
\cite{Niebles:IJCV08}		& 90.00 \\
\cite{Ali:PAMI10}			& 95.75 \\
\hline
Ground-truth tracks (1-NN Martin)	& 96.77 \\
\hline
Our method, DK-SSD-TR-C		& 92.47 \\
\hline
\end{tabular}
\caption{Comparison of different approaches for action recognition on the Weizmann database against our simultaneous tracking and recognition approach.}
\label{tab:weizmannComparison}
\end{table}

\begin{figure}
\centering
\includegraphics[width=\linewidth]{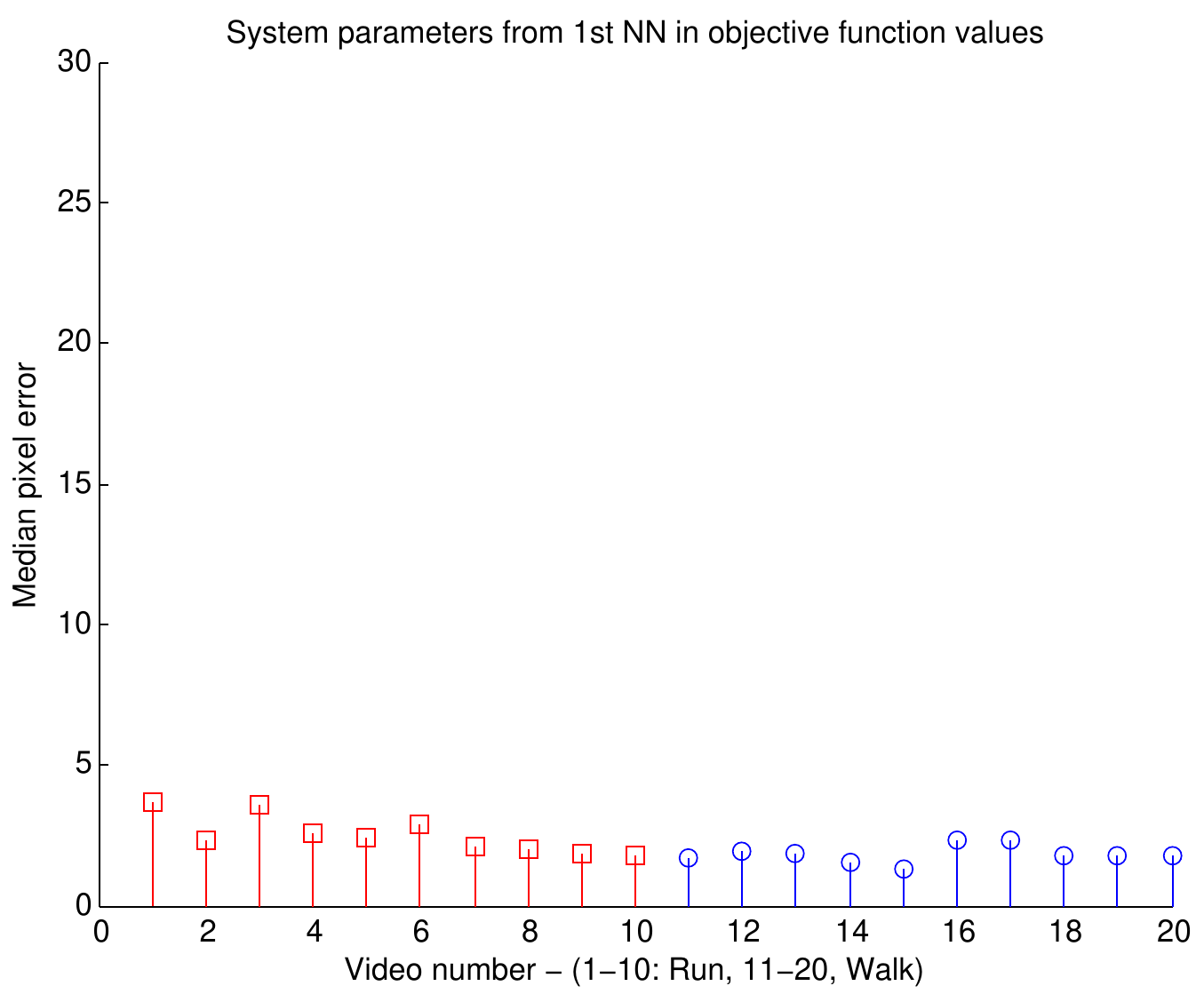}
\caption{Tracking walking and running sequences in the Weizmann database using trained system parameters from the database introduced in \S\ref{subsec:HopkinsActionsExperiments}. The ground-truth label of the first 10 sequences is running, while the rest are walking. The result of the joint tracking and recognition scheme are labeled as running (red) and walking (blue). We get 100\% recognition and under 5 pixel median location error.}
\label{fig:trackRecWeizmann}
\end{figure}

We will now demonstrate that our simultaneous tracking and recognition framework is fairly general and we can train for a dynamic template on one database and use the models to test on a totally different database. We used our trained walking and running action models (\ie the corresponding system parameters) from the 2-class walking/running database in \S\ref{subsec:HopkinsActionsExperiments} and applied our proposed algorithm for joint tracking and recognizing the running and walking videos in the Weizmann human action database \citep{Gorelick:PAMI07}, without any a-priori training or adapting to the Weizmann database. Of the 93 videos in the Weizmann database, there are a total of 10 walking and 10 running sequences. \figref{fig:trackRecWeizmann} shows the median tracking error for each sequence and the color codes its action, running (red) and walking (blue). As we can see, all the sequences have under 5 pixel median pixel error and all the 20 sequences are classified correctly. This demonstrates the fact that our scheme is general and that the requirement of learning the system parameters of the dynamic template is not necessarily a bottleneck as the parameters need not be learnt again for every scenario. We can use the learnt system parameters from one dataset to perform tracking and recognition in a different dataset.

\section{Conclusions, Limitations and Future Work}\label{sec:conclusions}

In this paper, we have proposed a novel framework for tracking dynamic templates such as dynamic textures and human actions that are modeled by Linear Dynamical Systems. We posed the tracking problem as a maximum a-posteriori estimation problem for the current location and the LDS state, given the current image features and the previous state. By explicitly considering the dynamics of only the foreground, we are able to get state-of-the-art tracking results on both synthetic and real video sequences against methods that use additional information about the background. Moreover we have shown that our approach is general and can be applied to any dynamic feature such as optical flow. Our method performs at par with state-of-the-art methods when tracking human actions. We have shown excellent results for simultaneous tracking and recognition of human actions and demonstrated that our method performs better than simply tracking then recognizing human actions when no pre-processing is performed on the test sequences. However our approach is computationally more efficient as it provides both tracks and template recognition at the same cost. Finally, we showed that the requirement of having a training set of system parameters for the dynamic templates is not restrictive as we can train on one dataset and then use the learnt parameters at test time on any sequence where the desired action needs to be found.

Although our simultaneous tracking and recognition approach has shown promising results, there are certain limitations. Firstly, as mentioned earlier, since our method uses gradient descent, it is amenable to converge to non-optimal local minima. 
Having a highly non-linear feature function or non-linear dynamics could potentially result in sub-optimal state estimation which could lead to high objective function values even when the correct class is chosen for tracking. Therefore, if possible, it is better to choose linear dynamics and model system parameter changes under different transformations instead of modeling dynamics of transformation-invariant but highly non-linear features. However it must be noted that for robustness, some non-linearities in features such as kernel-weighted histograms are necessary and need to be modeled. Secondly since our approach requires performing tracking and recognition using all the training data, it could be computationally expensive as the number of training sequences and the number of classes increases. This can be alleviated by using a smart classifier or more generic one-model-per-class based methods. We leave this as future work.

In other future work, we are looking at online learning of the system parameters of the dynamic template, so that for any new dynamic texture in the scene, the system parameters are also learnt simultaneously as the tracking proceeds. This way, our approach will be applicable to dynamic templates for which system parameters are not readily available.

\begin{acknowledgements}
The authors would like to thank Diego Rother for useful discussions. This work was supported by the grants, NSF 0941463, NSF 0941362 and ONR N00014-09-10084.
\end{acknowledgements}

\appendix{
\section{Derivation of the Gradient Descent Scheme} \label{sec:app-gradDescentDerivation}
In this appendix, we will show the detailed derivation of the iterations \eqref{eq:gradDescent} to minimimize the objective function in  \eqref{eq:jointOptimization}, with $\rho$, the non-differentiable kernel weighted histogram replaced by $\zeta$, our proposed differentiable kernel weighted histogram:
\begin{align}\label{eq:jointOptimization-contKWHist}
O(\ll_t,\x_t) = & \dfrac{1}{2\sigma_H^2}\|\sqrt{\zeta(\y_t(\ll_t))}-\sqrt{\zeta(\mu+C\x_t)}\|^2+ \nonumber \\
& \quad \dfrac{1}{2}(\x_t-A\hat{\x}_{t-1})^\top Q^{-1}(\x_t-A\hat{\x}_{t-1}).
\end{align}
Using the change in variable, $\z^\prime = \z+\ll_t$, 
the proposed kernel weighted histogram for bin $u$, \eqref{eq:contHistFunction},
\begin{align}
\zeta_u(\y_t(\ll_t)) = & \dfrac{1}{\kappa}\dsum_{\z \in \Omega} K(\z).\nonumber \\
& \quad  \left(\phi_{u-1}(\y_t(\z+\ll_t))-\phi_{u}(\y_t(\z+\ll_t))\right),  \nonumber
\end{align}
can be written as,
\begin{align} \label{eq:contHistFunction-1}
\zeta_u(\y_t(\ll_t)) = & \dfrac{1}{\kappa}\dsum_{\z^\prime \in \{\Omega-\ll_t\}} K(\z^\prime-\ll_t).\nonumber \\
& \quad  \left(\phi_{u-1}(\y_t(\z^\prime)) - \phi_{u}(\y_t(\z^\prime))\right),
\end{align}
Following the formulation in \cite{Hager:CVPR04}, we define the \emph{sifting} vector, 
\begin{align}
\tilde{\u}_j = \begin{bmatrix}
\phi_{j-1}(\y_t(\z^\prime_1))-\phi_{j}(\y_t(\z^\prime_1)) \\
\phi_{j-1}(\y_t(\z^\prime_2))-\phi_{j}(\y_t(\z^\prime_2)) \\
\vdots \\
\phi_{j-1}(\y_t(\z^\prime_N))-\phi_{j}(\y_t(\z^\prime_N))
\end{bmatrix}.
\end{align}
We can then combine these sifting vectors into the \emph{sifting matrix}, $\tilde{\mathbf{U}} = [\tilde{\u}_1, \ldots, \tilde{\u}_B]$. Similarly, we can define the kernel vector, $\mathbf{K}(\z) = \frac{1}{\kappa}[K(\z_1), K(\z_2), \ldots, K(\z_N)]^\top$, where $N = |\Omega|$ and the indexing of the pixels is performed column wise. Therefore we can write the full kernel-weighted histogram, $\zeta$ as,
\begin{align}
\zeta(\y_t(\ll_t)) = \tilde{\mathbf{U}}^\top \mathbf{K}(\z^\prime-\ll_t)
\end{align}
Since $\tilde{\mathbf{U}}$ is not a function of $\ll_t$,
\begin{align}\label{eq:appeanceTermDerivative}
\nabla_{\ll_t}(\zeta(\y_t(\ll_t))) = \tilde{\mathbf{U}}^\top \mathbf{J}_K,
\end{align}
where,
\begin{align}
\mathbf{J}_K & = \dfrac{1}{\kappa}\left[ \dfrac{\partial \mathbf{K}(\z^\prime - \ll_t)}{\partial \ll_{t;1}}, \dfrac{\partial \mathbf{K}(\z^\prime - \ll_t)}{\partial \ll_{t;2}}\right] \nonumber \\
 & = [\nabla K(\z_1^\prime-\ll_t), \nabla K(\z_2^\prime-\ll_t), \ldots, \nabla K(\z_N^\prime-\ll_t)]^\top.
\end{align}
where $\nabla K(\z^\prime - \ll_t)$ is the derivative of the kernel function, \eg the Epanechnikov kernel in \eqref{eq:epanKernel}. Therefore the derivative of the first kernel-weighted histogram, $\sqrt{\zeta(\y_t(\ll_t))}$ \wrt $\ll_t$ is,
\begin{align}\label{eq:gradDescentLTerm}
\mathbf{L} \doteq \dfrac{1}{2}\text{diag}(\zeta(\y_t(\ll_t)))^{-\frac{1}{2}}\tilde{\mathbf{U}}^\top \mathbf{J}_K
\end{align}
where $\text{diag}(\vv)$ creates a diagonal matrix with $\vv$ on its diagonal and $0$ in its off-diagonal entries. Since $\y_t(\ll_t)$ does not depend on the state of the dynamic template, $\x_t$, the derivative of the first kernel-weighted histogram \wrt $\x_t$ is $0$.

In the same manner, the expression for the second kernel weighted histogram, for bin $u$,
\begin{align}
\zeta_u(\mu+C\x_t) & = \frac{1}{\kappa} \dsum_{\z \in \Omega} K(\z) .\nonumber \\
& \quad \bigg(\phi_{u-1}(\mu(\z)+C(\z)^\top \x_t) - \nonumber \\
& \quad \quad \phi_{u}(\mu(\z)+X(\z)^\top \x_t)\bigg)
\end{align}
By using a similar sifting vector for the predicted dynamic template,
\begin{align}
\Phi_j = \begin{bmatrix}
(\phi_{j-1}-\phi_{j})(\mu(\z_1)+C(\z_1)^\top\x_t)\\
(\phi_{j-1}-\phi_{j})(\mu(\z_2)+C(\z_2)^\top\x_t)\\
\vdots \\
(\phi_{j-1}-\phi_{j})(\mu(\z_N)+C(\z_N)^\top\x_t)
\end{bmatrix},
\end{align}
where for brevity, we use
\begin{align}
& (\phi_{j-1}-\phi_{j})(\mu(\z)+C(\z)^\top\x_t) = \nonumber \\
& \quad \phi_{j-1}(\mu(\z)+C(\z)^\top\x_t)-\phi_{j}(\mu(\z)+C(\z)^\top\x_t). \nonumber
\end{align}
and the pixel indices $\z_j$ are used in a column-wise fashion as discussed in the main text. Using the corresponding sifting matrix, $\mathbf{\Phi}=[\Phi_1,\Phi_2,\ldots,\Phi_B] \in \Re^{N\times B}$, we can write, 
\begin{align}
\zeta(\mu+C\x_t) = \mathbf{\Phi}^\top\text{diag}(\mathbf{K}(\z))
\end{align}
Since $\zeta(\mu+C\x_t)$ is only a function of $\x_t$,
\begin{align}
\nabla_{\x_t}(\zeta(\mu+C\x_t)) = (\mathbf{\Phi}^\prime)^\top \text{diag}(\mathbf{K}(\z)) C,
\end{align}
where $\mathbf{\Phi}^\prime = [\Phi_1^\prime, \Phi_2^\prime, \ldots, \Phi_B^\prime]$, is the derivative of the sifting matrix with, 
\begin{align} \label{eq:gradDescentMTerm}
\Phi_j^\prime = 
\begin{bmatrix}
(\phi_{j-1}^\prime - \phi_j^\prime)(\mu(\z_1)+C(\z_1)^\top \x_t) \\
(\phi_{j-1}^\prime - \phi_j^\prime)(\mu(\z_2)+C(\z_2)^\top \x_t) \\
\vdots \\
(\phi_{j-1}^\prime - \phi_j^\prime)(\mu(\z_N)+C(\z_N)^\top \x_t)
\end{bmatrix},
\end{align}
and $\Phi^\prime$ is the derivative of the sigmoid function. Therefore, the derivative of the second kernel-weighted histogram, $\sqrt{\zeta(\mu+C\x_t)}$ \wrt $\x_t$ is,
\begin{align}
\mathbf{M}\doteq \dfrac{1}{2}\text{diag}(\zeta(\mu+C\x_t))^{-\frac{1}{2}} (\mathbf{\Phi}^\prime)^\top \text{diag}(\mathbf{K}(\z)) C.
\end{align}

The second part of the cost function in \eqref{eq:jointOptimization-contKWHist} depends only on the current state, and the derivative \wrt $\x_t$ can be simply computed as,
\begin{align} \label{eq:gradDescentLTerm}
\mathbf{d} \doteq Q^{-1}(\x_t-A\hat{\x}_{t-1}).
\end{align}
When computing the derivatives of the squared difference function in first term in \eqref{eq:jointOptimization-contKWHist}, the difference,
\begin{align}
\mathbf{a} \doteq \sqrt{\zeta(\y_t(\ll_t))} - \sqrt{\zeta(\mu+C\x_t)},
\end{align}
will be multiplied with the derivatives of the individual square root kernel-weighted histograms. 

Finally, the derivative of the cost function in \eqref{eq:jointOptimization-contKWHist} \wrt $\ll_t$ is computed as,
\begin{align}\label{eq:derivativeL}
\nabla_{\ll_t} O(\ll_t,\x_t) = \dfrac{1}{\sigma_H^2} \mathbf{L}^\top \mathbf{a},
\end{align}
and the derivative of the cost function \eqref{eq:jointOptimization-contKWHist} \wrt $\x_t$ is computed as,
\begin{align}\label{eq:derivativeX}
\nabla_{\x_t} O(\ll_t,\x_t) = \dfrac{1}{\sigma_H^2} (-\mathbf{M})^\top \mathbf{a} + \mathbf{d}.
\end{align}
We can incorporate the $\sigma_H^{-2}$ term in $\mathbf{L}$ and $\mathbf{M}$ to get the gradient descent optimization scheme in \eqref{eq:gradDescent}.

\bibliographystyle{spbasic}      
\bibliography{dynamicTrackingIJCV11.bbl}

%
%

\end{document}